\documentclass[12pt]{article}
\usepackage{times}
\usepackage{graphicx}%
\usepackage{multirow}%
\usepackage{amsmath,amssymb,amsfonts}%
\usepackage{amsthm}%
\usepackage[title]{appendix}%
\usepackage{xcolor}%
\usepackage{textcomp}%
\usepackage{manyfoot}%
\usepackage{booktabs}%
\usepackage{algorithm}%
\usepackage{algorithmicx}%
\usepackage{algpseudocode}%

\usepackage[numbers]{natbib}
\usepackage{url}

\usepackage{geometry}
\newgeometry{vmargin={0.5in}, hmargin={0.5in,0.5in}}

\usepackage{setspace}
\usepackage{caption}
\usepackage{subcaption}
\usepackage{diagbox}

\newcommand*\rotB[1]{\rotatebox[origin = c]{90}{\parbox{2.15cm}{#1}}}

\usepackage{listings}
\usepackage{xspace}
\def\CGPT{ChatGPT\xspace}
\def\lama{Llama\xspace}
\def\lamatri{Llama-3\xspace}
\def\rone{Rater 1\xspace}
\def\rtwo{Rater 2\xspace}
\def\ltool{LanguageTool\xspace}
\def\Tone{Task 1\xspace}
\def\Tseven{Task 7\xspace}

\theoremstyle{thmstyleone}

\theoremstyle{thmstyletwo}%

\theoremstyle{thmstylethree}%

\begin{document}

\author{
  Kundu, Anindita\\
  \texttt{kundu2@ualberta.ca}
  \and
  Barbosa, Denilson\\
  \texttt{denilson@ualberta.ca}
}
\title{Are Large Language Models Good Essay Graders?}

\maketitle




\abstract{We evaluate the effectiveness of Large Language Models (LLMs) in assessing essay quality, focusing on their alignment with human grading. 
More precisely, we evaluate \CGPT and \lama in the Automated Essay Scoring (AES) task, a crucial natural language processing (NLP) application in Education. 
We consider both zero-shot and few-shot learning and different prompting approaches.
We compare the numeric grade provided by the LLMs to human rater-provided scores utilizing the ASAP dataset, a well-known benchmark for the AES task. 
Our research reveals that both LLMs generally assign lower scores compared to those provided by the human raters; moreover, those scores do not correlate well with those provided by the humans. 
In particular, \CGPT tends to be harsher and further misaligned with human evaluations than \lama. 
We also experiment with a number of essay features commonly used by previous AES methods, related to length, usage of connectives and transition words, and readability metrics, including the number of spelling and grammar mistakes.
We find that, generally, none of these features correlates strongly with human or LLM scores.
We note that some of these features are intricately related to grading rubrics, including those in the ASAP dataset, suggesting that human graders often overlook some aspects of the rubric while grading. 
Indeed, we observe that, generally speaking, longer essays that use more connectives tend to get higher grades regardless of spelling and grammar mistakes. 
On the other hand, both LLMs not only can reliably detect spelling and grammar mistakes but also seem to take those mistakes into account when computing their score, further distancing themselves from human grading. 
Finally, we report results on \lamatri, which are generally better across the board, as expected.
Overall, while LLMs do not seem an adequate replacement for human grading, our results are somewhat encouraging for their use as a tool to assist humans in the grading of written essays in the future.}


\section{Introduction} \label{introduction}

Essay writing is a common component of student assessment, playing a pivotal role in education by providing insights into the text comprehension, critical thinking, and communication skills of the students~\cite{gibbs2005conditions}. Writing good essays requires students to articulate their thoughts clearly and coherently, demonstrating their understanding of a subject matter and their ability to construct logical arguments \cite{Joughin2009}. Moreover, effective essay assessment not only measures the knowledge and the skills students have acquired but also encourages deeper learning and engagement with the material \cite{scouller1998influence}.


Traditionally, essay grading has been mostly done by human graders. 
However, this presents significant challenges in modern education settings, especially when it comes to distance education, which contributes to the democratization of learning \cite{fomba2023institutional}.
Moreover, the global teacher shortage is a real and growing crisis. Countries struggle to find enough qualified educators to meet the needs of their student populations, resulting in ever-increasing workloads for teachers \cite{koc2015impact}. When teachers are responsible for large numbers of students it becomes difficult to provide individualized attention and detailed feedback for each essay. High student/teacher ratios can lead to burnout and decreased quality of evaluations, as teachers are pressed for time and may not be able to devote the necessary effort to each student's work.

Another significant challenge is the time-consuming nature of human evaluation. Reading, understanding, and providing constructive feedback on essays is a labor-intensive process that requires considerable time and effort \cite{ramesh2022automated}. This is especially problematic during peak periods, such as midterms or finals when teachers are inundated with grading responsibilities. The extensive time required for thorough evaluation can delay feedback, hindering students' ability to learn from their mistakes and improve their writing skills in a timely manner \cite{azaiz2024feedback}.

Finally, even setting aside the time pressure involved in essay scoring, one cannot forget that grades assessed by humans are subject to cognitive biases and stereotypical beliefs (intentionally or unintentionally), which can lead to inconsistent evaluations across different essays \cite{williamson1999mental}. Evaluators may have unconscious biases based on their personal preferences, cultural background, or prior experiences, which can influence their judgment \cite{benson2020unconscious}. Factors such as handwriting or familiarity with the student can also skew the assessment. These biases can lead to significant variability in scores, making it difficult to ensure that all students are fairly evaluated on the same level playing field.


In response to the limitations and scarcity of human graders, Automated Essay Scoring (AES) has emerged aimed at automatically grading and assessing the quality of written essays, thereby streamlining the assessment process and offering consistent scores to students. 
Automated essay scoring (AES) is the task of using computer-based assessment systems to automatically scores student-generated content (i.e., with no manual intervention). 
The development of AES systems has coincided with pivotal advancements in computational linguistics and Natural Language Processing (NLP) technology. AES relies heavily on NLP techniques at its core because it helps to extract deep meaningful language features that are indicative of writing quality. AES is an essential educational application of NLP which consists of having an AI agent assign a score to an essay based on its content, organization, and overall quality.

\begin{figure}[t]
    \centering
    \captionsetup{justification=centering}
    \includegraphics[width=0.85\textwidth]{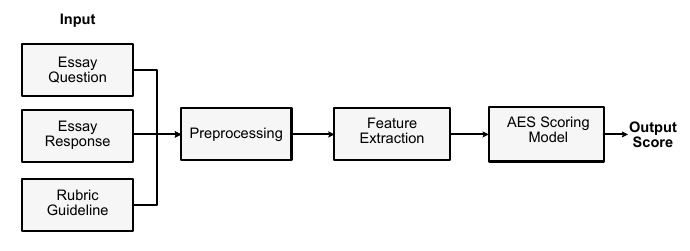}
    \caption{Automated Essay Scoring Pipeline}
    \label{aes-pipeline}
\end{figure}

AES system generally go through the followoing phases: data preprocessing, feature extraction, model training, and essay scoring, as illustrated in Figure \ref{aes-pipeline}. During data preprocessing, the essay is cleaned and standardized to ensure consistency. Feature extraction involves identifying relevant attributes, such as vocabulary usage, sentence structure, and coherence, which are then used to train the scoring models. Model training involves using labelled datasets of scored essays to teach the AES system how to grade new essays. Once the AES system is trained, it can be used to score unseen essays.

To date, AES has been approached primarily as a supervised learning task in which models are learned from sample essays graded by humans. 
Common techniques to solving the AES task include rule-based classification, feature-based statistical machine learning or deep neural network model training, and fine-tuning pre-trained models \cite{ramesh2022automated}. 
There is an underlying assumption in this approach, which is that the human graders correctly graded the essays according to the rubric. Under this assumption, a reasonable AES method learns how to emulate the scoring as done by the human graders and captured in the training data.

Although these developments are encouraging, there are several key technical challenges in applying these methods. 
The efficacy of machine learning AES tools depends heavily on the features used, be them selected by humans or via feature engineering. 
Deep learning AES systems, like all other tools based on deep learning, can generalize well but require large numbers of training samples (essays, in our task) which can be problematic. 
Fine-tuning a pre-trained model often requires a lot of resources. Also, for different domain/ subject applications, we need to train different models which involves additional expense. Additionally, AES systems must be robust enough to handle diverse writing styles and topics. 
The expectations of the AES system are frequently out of sync with the actual material due to the vast diversity of student-generated content \cite{horbach2019influence}.


Pre-trained Large Language Models (LLMs) are built using highly parallel architectures in an unsupervised fashion using massive amounts of data~\cite{zhao2023survey}. LLMs are capable of processing and understanding human language better than any kind of model before. Advancements in computing, the availability of larger datasets, and the improvement of machine learning algorithms have brought these language models closer to human-level performance on a multitude of tasks. LLMs learned from massive corpora have been shown to perform remarkably well on various language tasks, several of which they were not explicitly trained to do. For this reason, LLMs are described and evaluated as zero-shot (or few-shot) learners in the recent literature \cite{zero-shot-reasoners}. Recent results provide compelling arguments demonstrating that these models can generalize to unseen tasks and perform astonishingly well on complex tests such as the bar exam \cite{aydin2023chatgpt}. 

Different prompt engineering tactics can be utilized to bring their maximum potential to any low-resource learning tasks. Prompt engineering is the art of communicating with a generative model by writing precise and task-specific instruction prompts for improving the performance of LLMs \cite{giray2023prompt}. 
LLMs with proper prompts have been shown to adapt to various tasks without fine-tuning. Moreover, LLM performance can be enhanced by showing several examples of input and expected outcomes in the prompt which is known as a few-shot learning capability \cite{jiang2021learning}. Additionally, because of their ability to comprehend and produce human-like language outputs, LLMs can provide explanations of how they arrived at a grade which can be used to help students understand their grade. For this reason, LLMs are robust candidates for the development of AES systems.

While it is not entirely clear how or under which conditions language models like \CGPT and \lama acquire the ability to successfully complete NLP tasks without training~\cite{zhao2023survey}, the advent of LLMs has opened up the door for a different kind of AES system that does not rely on expert training data: providing a prompt requesting an essay to be graded given the rubric. 
In this study, we examine the performance of two popular LLMs: OpenAI's \CGPT (ChatGPT-3.5-Turbo) and Meta's \lama \cite{touvron2023llama} as automated essay scoring tools. 
We used version 3.5 of ChatGPT, also known as InstructGPT~\cite{ouyang2022training}, which was obtained by refining the original GPT-3 \cite{gpt-3-paper} through reinforcement learning guided by actual human feedback by OpenAI engineers, resulting in a model that was far superior in the understanding of instructions. Additionally, we extended our analysis to include the most recent release, \lamatri as shown in Appendix \ref{llama3-results}.

\subsection{Research Questions}
We investigate this in-context learning approach, guided by a fundamental question: Can AES based on LLMs match human graders in judging the quality of essay writing? Related to that question, we revisit the role of annotated training data for the AES task and consider whether LLMs can act as zero/few-shot classifiers to assess essay quality and explain their understanding reasonably behind the score they provide. In particular, our study aims to address the following research questions:

\begin{itemize}
    \item \textbf{RQ1:} Do human scores align with the LLM scores? 
    \item \textbf{RQ2:} What are the possible reasons behind the similarity/difference in scores?
    \item \textbf{RQ3:} Do LLMs offer explanations in a tone that reflects their scores?
    \item \textbf{RQ4:} Can LLMs correctly identify spelling and grammatical errors and reflect those into their scoring? 
\end{itemize}

\subsection{Outline}

The rest of the paper is structured as follows. Section \ref{background} provides a concise review of related research. In Section \ref{methods}, we present detailed methodology about the dataset in subsection \ref{dataset}, along with the experimental setup in subsection \ref{settings}. Section \ref{results} is dedicated to the presentation and analysis of the results.
Chapter \ref{sec-prompt-eng} is an extension and additional experiment results with different prompts. 
Finally, in Section \ref{conclusion}, we offer our conclusions and discuss potential avenues for future enhancements.

\section{Related Work} \label{background}

Human raters possess diverse skills in evaluating assignments, including understanding the meaning of the text, assessing critical thinking, creativity, and content relevance \cite{cheville}. They excel in evaluating logic, argument quality, and factual correctness, and can judge audience awareness \cite{lewis_sevcikova_human_2018}. However, maintaining consistency and eliminating subjectivity can be challenging \cite{lewis_sevcikova_human_2018}. Additionally, large-scale essay scoring can be labor-intensive and time-consuming \cite{zhang_contrasting_2013}, which in turn can lead to fatigue and inconsistencies by human scorers. With the increasing demand for personalized education and the growing shortage of teachers, Automated Essay Scoring (AES) systems are increasingly needed. AES can assist in managing large classes by providing consistent and efficient essay assessments. This capability can alleviate some of the workload from educators, allowing them to focus on more interactive and engaging aspects of teaching.



The first automated essay scoring system was developed more than 50 years ago, in 1966 \cite{suzen2020automatic}. Since then, AES systems have become more advanced, offering more features than the early versions. 
There are many good surveys of AES systems, such as \citet{ramesh2022automated} and \citet{ke2019automatedsurvey}. 
Most AES research focuses on supervised learning of holistic scoring due to the availability of annotated corpora and their commercial value in automating standardized test grading.
Following a similar evolution as other fields of AI, AES systems evolved from using hand-crafted rules in PEG (Project Essay Grade) \cite{page-1966} to features-based statistical machine learning such as E-raters \cite{attali2006automated}, Intelligent Essay Assessor (IEA) \cite {foltz1999intelligent} to using deep neural models such as Recurrent Neural Network (RNN), Convolutional Neural Network (CNN), Long Short-Term Memory (LSTM) that learn own representations of the data \cite{taghipour-ng-2016-neural, dong2017attention}.
We also see approaches which learn from transfer learning for domain adaptation ~\cite{phandi2015flexible}, and a combination of hand-picked features and deep learning approaches ~\cite{uto2020neural_handcraft}.

As argued by \citet{ke2019automatedsurvey}, the most important features for evaluating essay quality are grammaticality, organization, persuasiveness, coherence, mechanics, and relevance. 
It should be noted that extracting these features from text is difficult as even state-of-the-art NLP tools to assess these aspects of running text cannot achieve human-level performance. 
Therefore, it becomes challenging to train models to holistically evaluate an essay according to all of these dimensions together. 
It is worth mentioning that the pre-LLM state-of-the-art in NLP was far from capable of addressing most of these criteria, especially those related to higher-level notions of writing quality. 


The advent of pre-trained large language models represents a major step forward in NLP allowing more generalized language representations from vast corpora that can then be fine-tuned for a specific task. 
Of course, LLMs have been used for the AES task.
\citet{wang-bert-use} introduce a joint multi-scale essay representation using BERT employing multiple losses and transfer learning to demonstrate effective generalization to long-text AES tasks. \citet{yang-etal-2020-enhancing} focus on fine-tuning LLMs such as BERT utilizing regression and ranking.
\citet{khademi6can} evaluated \CGPT and Google Bard against human scores on IELTS academic writing tasks and found that these models are far from achieving human performance. However, their study did not focus in detail on why there are vast differences between the scores. \citet{mizumoto2023exploring} leveraged GPT-3 version \textit{text-davinci-003} in a zero-shot setting using non-native English corpus (TOFEL11) highlighting the potential use of ChatGPT incorporated with linguist features can enhance accuracy in AES. \citet{latif2024fine} fine-tuned \CGPT to confirm the effectiveness in education-specific tasks by significantly improving score prediction over BERT (on average $10.6\%$). They neither fully exploit few-shot learning approaches nor utilize any prompt clarity-enhancing techniques with zero-shot.


While fine-tuning has been shown to greatly improved the performance of LLMs across tasks, including AES, this approach require considerable amounts of labelled data, which is a problem. 
Prompt engineering approaches become useful in this context as they require less data and allow for enhancing in-context learning and maximizing efficient outcomes. 
One recent work by \citet{mansour2024can} investigated \CGPT and \lama with four different prompts and found that LLMs are highly sensitive to the prompt. This study showed with the right prompt engineering, LLMs can offer competitive albeit not yet superior performance compared to state-of-the-art models. However, their study did not report any single prompt type using which LLMs can consistently perform close to human scoring across all essays. Another recent work by \citet{Helmeczi2023Few} explores few-shot learning using BERT for AES tasks. 


Offering feedback is crucial for students to learn from their mistakes and develop their writing skills. However, delivering timely and valuable feedback, particularly to middle and high school students, can pose a significant challenge for educators. 
Some recent works discuss training models to provide scores for individual essay traits as feedback to improve students' writing performance \cite{Hussein2020, kumar2021many}. 
However, it is unclear whether and how these methods impacted students as measured by the overall score. 
Other researchers have looked into whether generative models can be used to assist students in learning. \citet{LLMonassignments} conducted an experimental study integrating GPT-3.5 into an automated assessment platform to generate personalized hints for programming assignments, observing that while students rated GPT-generated hints positively, these hints did not significantly improve performance. 
Another study by \citet{MEYER2024100199} examines the impact of large language model-generated feedback on upper secondary students' writing tasks, revealing significant improvements in revision performance, task motivation, and positive emotions compared to the baseline of receiving no feedback.
While both \CGPT and \lama did provide output that could potentially be used as feedback to improve the essays, we did not investigate this question, as we do not have access to students. Instead, we looked at that output to test whether its general tone (measured through sentiment analysis) was consistent with the numeric scores provided by the LLMs (i.e., whether harsher language was correlated with lower scores).


In this study, we aim to investigate the alignment between human grades and grades assigned by LLMs in AES tasks, assessing the extent to which LLM-based AES can potentially substitute human raters in evaluating essay quality. 
Generally speaking, we find a very weak correlation between human-assigned and LLM-generated scores.
We explore the possible reasons underlying this discrepancy by looking at a range of essay features, most of which are explicitly mentioned in the grading rubric and widely used in previous work for the AES task. 
Incidentally, we also find that human grades do not correlate very strongly with these features, suggesting that AES systems that rely on these features, even those that exhibit low error, might not be learning the ``right'' scoring function.
Additionally, this research aims to investigate whether the explanations provided by large language models (LLMs) are consistent with the scores they assign to essays. we explore if the tone and content of the explanations align with the grading criteria and reflect the same level of evaluation as indicated by the scores. We seek to understand whether LLMs not only generate numerical scores but also provide coherent and relevant explanations that justify these scores, offering insights into their grading logic and approach. Furthermore, we seek to determine whether LLMs possess the ability to accurately identify and assess spelling and grammatical errors, and grade essays accordingly. 

\section{Methodology}\label{methods}

\subsection{Dataset}\label{dataset}
We used the Automated Student Assessment Prize (ASAP) benchmark  \cite{asap-aes}, well-known for the Automated Essay Scoring (AES) task. It consists of around 13000 essays written by students of grade levels from 7 to 10. There are eight sets of tasks in ASAP, each linked to different prompts and scoring ranges. Detailed statistical information about this dataset is shown in Table \ref{dataset-desc}. Students wrote essays in response to three types of prompts. An argumentative essay is when students write to convince others about their opinion on a topic. They need to research the topic, find evidence, and explain their ideas clearly. A narrative essay is when students write a story. Here they can make up characters or use personal experience, and events to create an interesting tale. A source-dependent essay is when students need to read an article and then write an essay using the information from that article. In this dataset, two to three human graders annotate each essay. The final score of an essay is the sum of scores given by the individual human graders. 

\begin{table}[!ht]
    \centering
    \caption{Detail information about ASAP dataset}
    \begin{tabular}{ccccc}
        \textbf{Task}  & \textbf{Essay Type} & \textbf{Number of Essays} & \textbf{Score Range} & \textbf{Grade Level } \\ \hline
        1 & Argumentative & 1783 & 1-6 & 8 \\ 
        2 & Argumentative & 1800 & 1-6 or 1-4 & 10 \\ 
        3 & Source dependent & 1726 & 0-3 & 10 \\ 
        4 & Source dependent & 1772 & 0-3 & 10 \\ 
        5 & Source dependent & 1805 & 0-4 & 8 \\ 
        6 & Source dependent & 1800 & 0-4 & 10 \\ 
        7 & Narrative & 1569 & 0-15 & 7 \\ 
        8 & Narrative & 723  & 0-30 & 10 \\ \hline
    \end{tabular}
    \label{dataset-desc}
\end{table}

In the ASAP dataset, only two of the eight prompts (sets 7 and 8) provide scores for specific essay traits like content, organization, and style, while the remaining six prompts offer only overall scores. For our experiment, we selected \Tone and \Tseven, with holistic and trait scores respectively, to avoid repetition of the same kind of analysis. Two human raters graded both of these tasks. \Tone asked for students' opinions on the usefulness of computers and was evaluated holistically starting from 1 to 6 scores. The rubric for this task describes six different score specifications. Conversely, \Tseven focused on students' experiences with patience and was assessed based on four trait scores: ideas, organization, style, and convention. 
The rubric for this task is more intricate, with requirements describing each trait score ranging from 0 to 3, except for the ``ideas'' trait, which ranges between 0 and 6, resulting in a final score range of 0-15 for each essay. 
Appendix \ref{secA1} includes a detailed rubric for \Tone and \Tseven.

In 2018, \citet{mathias-bhattacharyya-2018-asap} introduced ASAP++, a dataset with manual annotations for the other six prompts with respect to the particular traits. One proficient English annotator graded each essay, scoring from 1 to 6 for independent attributes. 
We used this extended version of the ASAP dataset mentioned previously\footnote{Data can be found at \url{https://lwsam.github.io/ASAP++/lrec2018.html}}. The traits for \Tone are ideas and content, organization, word choice, sentence fluency, and conventions which are also annotated by the human graders. 
Table~\ref{Asap++criteria} shows the details of the trait score rubric used for grading the essays in \Tone of ASAP++, which we used in our study. It is worth mentioning that ASAP (and ASAP++) are \emph{de facto} the best and largest AES datasets widely available today. 
\begin{table}[!t]
    \centering
    \caption{Traits of ASAP++ dataset for \Tone }
    \begin{tabular}{|p{1.5in}|p{3.5in}|}
    \hline
        Traits & Scoring Requirements \\ \hline
        Ideas \& Content & Assesses the clarity, depth, and engagement of the essay's content, focusing on the presence of clear main ideas, well-supported arguments, and relevance to the audience and purpose. \\ \hline
        Organization & Evaluates the structure and flow of the essay, looking for a logical sequence of ideas, well-defined paragraphs, and overall organization suitable for the given task. \\ \hline
        Word Choice & Examines the use of vocabulary to convey the message effectively, emphasizing the selection of precise, impactful words, and their appropriateness for the intended audience and purpose. \\ \hline
        Sentence Fluency & Considers the writing's flow and rhythm, assessing the variety and structure of sentences, and their ability to engage the reader while maintaining clarity and coherence. \\ \hline
        Conventions & Focuses on the writer's command of standard writing conventions, including punctuation, spelling, capitalization, grammar, and usage, with an emphasis on minimizing errors that may disrupt readability and communication. \\ \hline
    \end{tabular}
    \label{Asap++criteria}
\end{table}

\subsection{Experimental Setup}\label{settings}
\subsubsection{Prompt design and response generation}
We consider two popular LLMs for response generation: \CGPT and \lama. Specifically, we consider the versions \textit{gpt-3.5-turbo}, and \textit{\lama-2-70B-chat} for our experiment. From the ASAP dataset, we input the prompt given to students, their corresponding essays, rubric guidelines, score range, and any additional instructions into the \CGPT and \lama. The rubric and guidelines are presented in the same format and detail as they would be for human raters, mirroring how they were described in the dataset. Appendix \ref{secA1} includes examples of prompt and ChatGPT or \lama-generated response. Next, we ask LLMs for a numeric score in the appropriate range and an explanation for \Tone. \Tseven has four trait scores and a total score along with an explanation from \CGPT, \lama. We make sure that the entire input (prompt + actual essay + output) never exceeds its token limit (4096 token as of December 2023). 
While generating responses, we employed default parameter settings to maintain balance and maximize the creativity and diversity of LLMs when solving a complex task like AES.

\subsubsection{Extracting features from essays}
\label{sec-essay-feature-extraction}
We extract a wide range of features from essays to further analyze different aspects of how the grading process of LLMs coaligns with humans. 

\begin{itemize}

\item \textbf{{Essay Statistics}:}
We have extracted several basic statistical features for each essay such as the number of sentences and the number of tokens found in the essay. \\

\item \textbf{{Readability}: }
This assessment in our analysis aimed at understanding whether human graders implicitly consider readability and if LLMs are pre-trained to evaluate text difficulty. Readability refers to the ease with which a reader can understand a written text. It is often assessed using metrics that consider factors such as sentence length, word complexity, and sentence structure. To achieve this, we employed several well-known readability formulas, including the Flesch Reading Ease \cite{kincaid}, Automated Readability Index \cite{automated}, Coleman-Liau Index \cite{coleman}, Dale–Chall readability score \cite{dale}, Flesch–Kincaid Grade Level \cite{kincaid}, Gunning Fog Index \cite{gunning}, Linsear Write Formula, and SMOG index \cite{smog}. These formulas help determine how challenging an essay is to read and comprehend. We have followed the implementation of text readability formulas from the paper by \citet{martinc-etal-2021-supervised}. This approach allows us to investigate how readability influences the grading process for both human graders and language models.\\

\item \textbf{{Linking Words}:}
In our analysis, we extracted features from essays that include the use of the total and unique number of transition words\footnote{\url{https://www.grammarly.com/blog/connecting-sentences/}} and ``FANBOYS''\footnote{For, and, but, or, yet, and so.} words. Transition words serve as a proxy for the number of arguments in the essay. Transition words and Fanboys (coordinating conjunctions) serve similar purposes in English writing by connecting ideas and improving the flow of text. Transition words, such as ``however" or ``therefore," indicate relationships between sentences or paragraphs, aiding in smooth transitions and clarifying the logical progression of ideas. Fanboys, on the other hand, specifically join clauses within a sentence, emphasizing relationships like addition (``and"), contrast (``but"), choice (``or"), or conclusion (``so"). Both types of words enhance coherence and cohesion in writing, helping readers navigate complex texts more effectively.\\

\item \textbf{{Language Mistakes}:} Our study further analyzes the impact of language mistakes on essay scores given by human raters and LLMs. We extracted various types of language mistakes from the essays, such as spelling, grammar, and punctuation mistakes, to see how these were reflected in the scores assigned. Our aim was to understand the extent to which these language mistakes affect the overall scoring. To achieve this, we utilized two different language-checking tools, which are detailed below, to identify and categorize these errors. By examining the correlation between the presence of language mistakes and the scores given, we aim to gain insights into the evaluation criteria used by both human and LLMs.

\begin{itemize}
    \item \textbf{{Aspell:}} We have utilized \textit{Aspell}\footnote{\url{http://aspell.net/}}, an open-source spell-checker which offers an improved and comprehensive English language dictionary. \textit{Aspell} accommodates American, British, and Canadian spelling preferences encompassing multiple spellings for a single word. Our use of the \textit{Aspell} enabled us to find the number of misspelled words in each essay. We have also kept a record of all the misspelled words reported by \textit{Aspell} for every essay. \\

    \item \textbf{{LanguageTool}:}
Additionally, to cross-check the mention of mistakes in LLM's explanation, we used the open-source language checking tool \textit{LanguageTool}\footnote{\url{https://languagetool.org/}} similar to the work of Zesch et al.~\cite{grammar-checker-paper}. 
It helps to detect issues related to spelling, grammar, punctuation, style, and more. We have utilized specifically version 6.1 which can identify the several language errors as specified below. 

We have further classified these errors into five main categories for simplification. Compounding and spelling errors fall under the spelling category. Collocation, nonstandard phrases, confused words, miscellaneous errors, and grammar issues are grouped together as grammar errors. Typography, redundancy, style, and British English errors form the style category. Casing and punctuation constitute their own group.
\begin{enumerate}
\item \textbf{Spelling Category: }
    \begin{itemize}
        \item \textit{Compounding} errors related to compound words, whether they should be separate words, hyphenated, or combined. Example: \textit{I'm working full time (full-time).} 
        \item \textit{Typos} are simple spelling mistakes or accidental keystrokes that result in incorrect words. Example: \textit{I would like a peice of pie (should be piece).}
    \end{itemize}
\item \textbf{Grammar Category:}
    \begin{itemize}
        \item \textit{Grammar} errors encompass mistakes related to sentence structure, verb tenses, subject-verb agreement, etc. Example: \textit{He go to the store yesterday.}
        
        \item \textit{Confused words} errors where words are confused with others that sound similar or have similar spellings. Example: \textit{Their going to the park later.}
        
        \item \textit{Miscellaneous} errors cover various other language issues not falling into specific categories. Example: \textit{This is is just an example sentence. }
        
        \item \textit{Collocations} errors where words are used together incorrectly or inappropriately. Example: Open your books at (to) page 6. 

        \item \textit{Nonstandard phrases} errors involve using unconventional or informal expressions. Example: \textit{I never have (I have never been) been to London. }
    \end{itemize}
\item \textbf{Style Category:}
    \begin{itemize}
        \item \textit{Style} errors relate to writing style consistency and adherence to specific writing conventions. Example: \textit{I sometimes am happy. }
        
        \item \textit{Typography} errors refer to mistakes related to the visual appearance of text, including font choice, spacing, and formatting. Example: \textit{tHeRe aRe tYPoGraPhIcaL eRRorS iN tHis SeNtEncE. }
        
        \item \textit{British English} errors involve using British spelling or vocabulary inappropriately in American English contexts. Example: \textit{Colour (British) vs. color (American).}
        
        \item \textit{Redundancy} errors occur when unnecessary words or phrases are repeated. Example: \textit{He personally went there himself.}

    \end{itemize}

    \item \textbf{Punctuation Category:} errors involve incorrect placement or misuse of punctuation marks. Example: \textit{Sharon and Sue, went into the office early yesterday, to complete a project }(should be without the unnecessary comma).
    
    \item \textbf{Capitalization Category} errors related to the incorrect use of uppercase or lowercase letters. In LanguageTool it is mentioned as `casing'. Example: \textit{I live in new york.}
\end{enumerate}
\end{itemize}
\end{itemize}

\subsubsection{Extracting features from LLM Explanations}\label{explanation-extraction}
We gathered various fundamental statistics from each LLM explanation, including the count of sentences and tokens. Additionally, we employed sentiment analysis to assess the tone of the explanation and implemented rule-based methods to extract specific information from the LLM explanations, as described below:

\begin{itemize}
\item \textbf{{Length-related Statistics:}} We extracted length-related features from LLM-generated explanations, such as the number of sentences and the number of tokens.

\item \textbf{{Sentiment Score of the Explanation:}} To explore whether the tone and content of the explanations align with the LLM's scores, we calculated the sentiment score of the explanations using VADER (Valence Aware Dictionary and sEntiment Reasoner), a sentiment analysis tool specifically designed for social media content~\cite{Hutto_Gilbert_2014}. 
VADER combines a lexicon-based approach with rule-based techniques to assess sentiment polarity (positive, negative, or neutral) and intensity. VADER provides the following sentiment scores:

\begin{enumerate}
    \item Compound score: This overall sentiment score ranges from -1 (extremely negative) to 1 (extremely positive). It captures the overall sentiment of the text.
    \item Positive score: Indicates the proportion of positive words in the text.
    \item Negative score: Reflects the proportion of negative words.
    \item Neutral score: Represents the neutrality of the text.
\end{enumerate}

We extracted all these four sentiment scores from VADER for each \CGPT and \lama response. Then, we split each response into individual sentences and asked for a polarity score to determine the maximum sentence-wise compound sentiment score. \\

\item {\textbf{Mention of Language Mistakes: }} \label{rule-based-matching}
As mentioned before, we asked the LLMs to provide an explanation to justify the numerical score. We observed that, often, the explanation mentioned grading criteria v explicitly mentioned in the rubric. 
We extracted those mentions to measure the LLMs' self-consistency; more specifically, we sought to investigate whether the LLM's final score was lower when the explanations contained mentions to problems. As will be discussed later, we partitioned the essays into three groups based on the corresponding explanation provided by the LLM when grading them.
The first group consists of essays for which the LLMs did not mention any problem (e.g., no grammar mistakes). The second group consists of essays where the LLMs mentioned problems without a qualifier.
The third group consists of the essays for which the LLM used a qualifier (e.g., ``several'', ``multiple'', etc.) to emphasize the presence of mistakes.

\begin{figure}[!t]
    \centering
    \captionsetup{justification=centering}
    \includegraphics[width=0.8\textwidth]{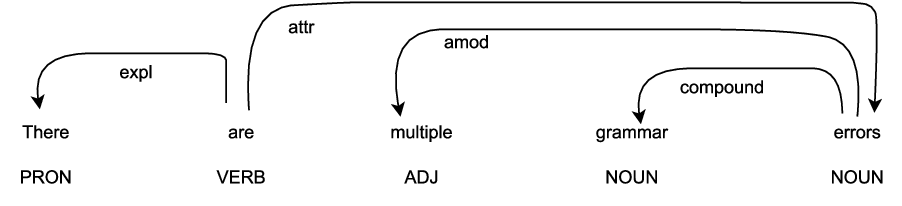}
    \caption{Dependency Tree }
    \label{dependency-tree}
\end{figure}

We used Spacy's \textit{DependencyMatcher}\footnote{\url{https://spacy.io/api/dependencymatcher}} to process the LLM explanations and extract mentions of grammatical, spelling, punctuation, and capitalization mistakes from the LLM's explanations. Dependency matching involves identifying and extracting linguistic patterns based on the syntactic dependencies between words in a sentence and generalizes over lexical methods relying on regular expressions, for example.
We randomly selected 60 samples from the explanations generated by \CGPT to manually hand-craft dependency patterns.  Figure \ref{dependency-tree} shows a partial dependency tree from the actual sentence provided by ChatGPT as part of an explanation. 


Our patterns also took into consideration the case where different kinds of mistakes were present, as in \textit{There are grammatical and spelling mistakes in the essay}. In this sentence, the token \textit{grammatical} is conjoined with another token \textit{spelling}. To represent such a conjunct relationship, we curated patterns to handle up to five different mentions as conjunction. We used the same patterns for both language models (\CGPT and \lama). 

\end{itemize}


\subsection{Evaluation Metric}

We performed a correlation analysis between human scores and LLM-provided scores. Furthermore, we have leveraged the ASAP++ dataset to compare the human provided trait scores of \Tone to LLM's scores. We have also performed correlation analyses between all scoring methods and various features extracted from essays and explanations to examine different aspects of human and LLM essay scoring. Our primary focus centers on convention scores of the ASAP++ dataset, as there are currently no other available NLP tools for validating the remaining four criteria \cite{ke2019automatedsurvey}. Convention scores are derived from assessments related to punctuation, spelling, grammar, and capitalization. Notably, the explanation provided by \CGPT and \lama often mentions spelling and grammar issues. Consequently, we opt for these two metrics to assess the quality of LLM's provided explanation. 

We use Pearson correlation analysis to measure the agreement or similarity between the scores assigned by human raters and those generated by LLMs. The Pearson correlation coefficient is denoted as $r$ ranges from -1 to 1, where:

\begin{itemize}
    \item $r = 1$ indicates a perfect positive correlation, meaning that as one variable increases, the other variable also increases proportionally.
     \item $r = -1$ indicates a perfect negative correlation, meaning that as one variable increases, the other variable decreases proportionally.
    \item $r = 0 $ indicates no linear correlation between the variables.
\end{itemize}
we classified the magnitude of Pearson correlation ($r$) values to assess the strength of relationships between variables as follows: very weak (0.0 to 0.19), weak (0.2 to 0.39), moderate (0.4 to 0.59), strong (0.6 to 0.79), and very strong (0.8 to 1.0).

\section{Results}\label{results}

\subsection{RQ1: Do human scores align with the LLM scores? }

\begin{figure}[!h]%
    \centering
    \captionsetup{justification=centering}
    \includegraphics[width=0.8\textwidth]{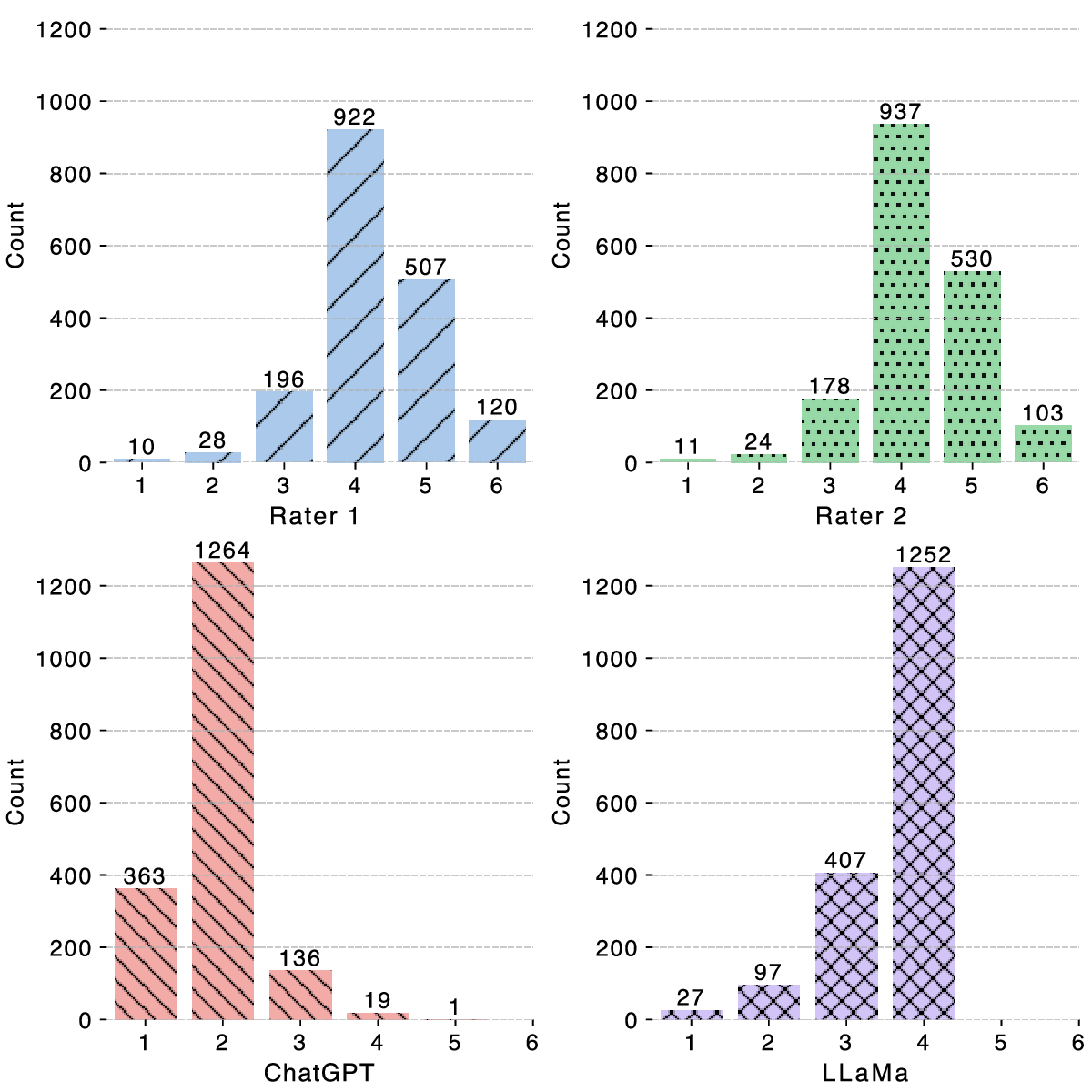}
    \caption{Score distribution of human raters and LLMs for \Tone }
    \label{score-dist-p1}
\end{figure}

For \Tone, both \CGPT and \lama assigned valid overall scores ranging from 1 to 6 to all $1783$ essays. Table \ref{desc-stat-p1} presents the descriptive statistics for the four scoring methods, while Figure \ref{score-dist-p1} illustrates the score distribution. It is notable that LLMs tend to assign lower scores compared to human raters.

\begin{table}[!ht]
    \centering
    \captionsetup{width=0.8\textwidth}
    \caption{Descriptive statistics for each scoring method of \Tone }
    \begin{tabular}{lcccc}
        \textbf{} & \textbf{Rater 1 } & \textbf{Rater 2 } & \textbf{\CGPT } & \textbf{\lama }\\ \hline
        \textbf{Mean} & 4.26 & 4.27 & 1.90 & 3.62 \\ 
        \textbf{SD} & 0.84 & 0.82 & 0.56 & 0.66 \\ 
        \textbf{Min} & 1.00 & 1.00 & 1.00 & 1.00 \\ 
        \textbf{Median} & 4.00 & 4.00 & 2.00 & 4.00 \\ 
        \textbf{Max} & 6.00 & 6.00 & 5.00 & 4.00 \\ 
    \end{tabular}
    \label{desc-stat-p1}
\end{table}

\begin{table}[!h]
    \centering
    \captionsetup{width=0.8\textwidth}
    \caption{Descriptive statistics for each scoring method of \Tseven}
    \begin{tabular}{lcccc}
        \textbf{} & \textbf{Rater 1 } & \textbf{Rater 2 } & \textbf{\CGPT } & \textbf{\lama } \\ \hline
        \textbf{Mean} & 8.02 & 8.04 & 2.07 & 2.66 \\ 
        \textbf{SD} & 2.42 & 2.52 & 1.83 & 2.68 \\ 
        \textbf{Min} & 0.00 & 0.00 & 0.00 & 0.00 \\ 
        \textbf{Median} & 8.00 & 8.00 & 2.00 & 3.00 \\ 
        \textbf{Max} & 12.00 & 12.00 & 9.00 & 26.00 \\ 
    \end{tabular}
    \label{desc-stat-p7}
\end{table}

\begin{figure}[!h]%
    \centering
    \captionsetup{justification=centering}
    \includegraphics[width=1\textwidth]{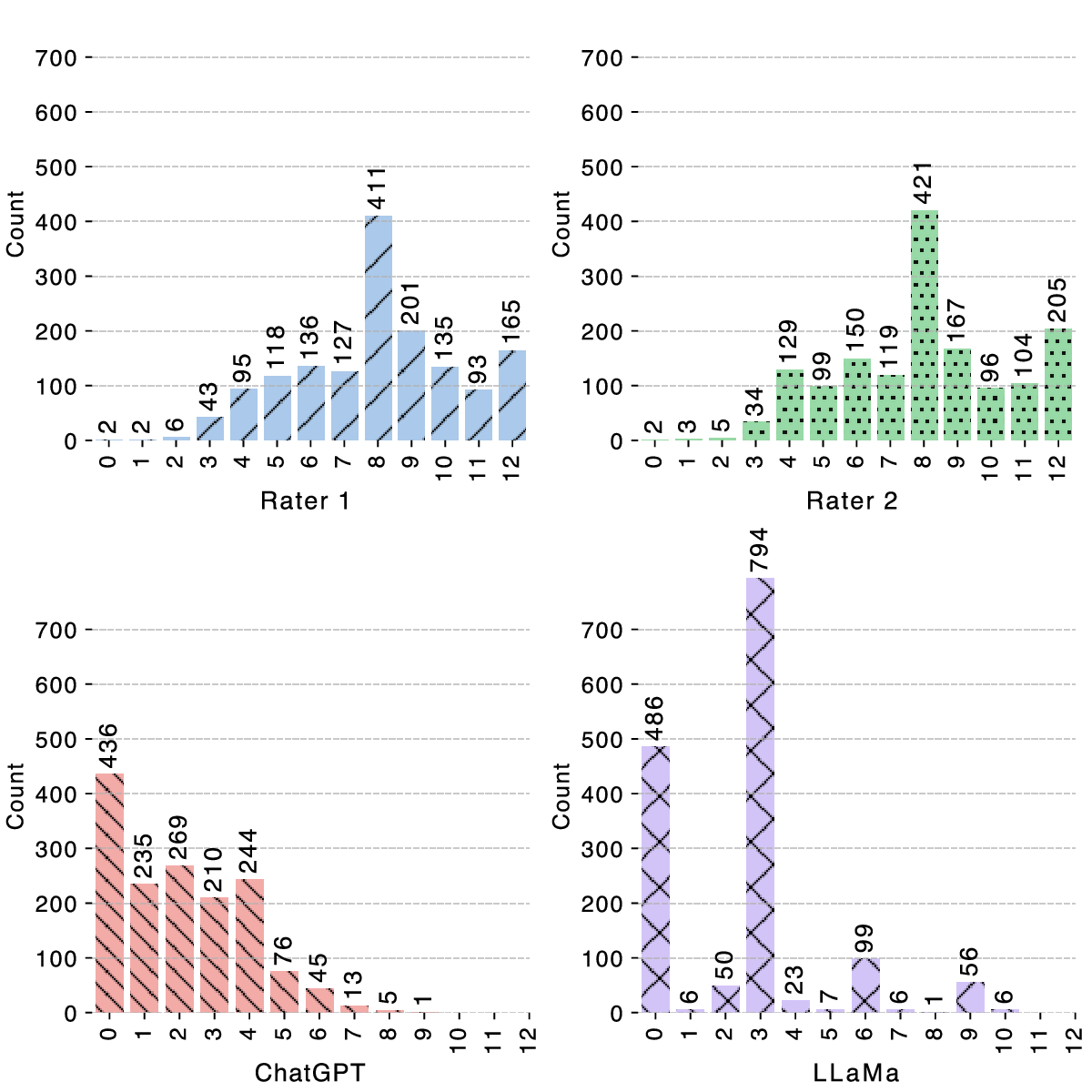}
    \caption{Overall score distribution of human raters and LLMs for \Tseven}
    \label{score-dist-p7}
\end{figure}

In \Tseven, \CGPT and \lama generated four trait scores, along with a final overall score that sums all trait scores for a total of $1569$ essays. Table \ref{desc-stat-p7} presents the descriptive statistics for the four grading methods. \CGPT ($M=2.07, SD=1.83$) and \lama ($M=2.66, SD=2.68$) both assigned significantly lower mean scores than both human raters ($M=8.02, SD=2.42$ and $M=8.04, SD=2.52$) revealing a trend similar to that observed in \Tone where LLMs tended to assign lower scores than human raters. Notably, \CGPT and \lama assigned out-of-range trait scores to 19 and 16 samples, respectively. This discrepancy may be attributed to the complexity of \Tseven's rubric compared to that of \Tone. Figure \ref{score-dist-p7} illustrates the score distribution for this task, excluding the outlier samples when constructing the score distribution graphs. Detailed trait-wise score distribution and statistics are shown in Appendix \ref{secB}.

\subsubsection{LLM and human scores do not correlate} \label{human-llm-donot-correlate}
Tables \ref{human-llm-corr-p1} indicates that \CGPT scores show a weak positive correlation with the scores provided by both human raters ($r_{\text{rater1}}=0.23, p< 0.001, d\footnote{Effect size details can be found in Appendix \ref{cohenD}}=-3.30$ and $r_{\text{rater2}}=0.21, p< 0.001, d=-3.38$) in \Tone.  While \lama scores moderately positively correlate ($r_{\text{rater1}}=0.59, p< 0.001, d=-0.85$ and $r_{\text{rater2}}=0.58, p< 0.001, d=-0.87$) with human scores exceptionally for \Tone . Also, we observed no strong inter-LLM correlation ($r=0.32, p<0.001, d=2.8$) between \lama and  \CGPT. All the detailed statistical analyses can be found in Appendix \ref{appendix-section-statistics}.

\begin{table}[h!]
    \centering
    \captionsetup{width=0.9\textwidth}
    \caption{Correlations between human rater and LLM scores for \Tone }
    \begin{tabular}{lcccc}
        \textbf{Score} & \textbf{Rater 1 } & \textbf{Rater 2 } & \textbf{ChatGPT } & \textbf{\lama} \\ \hline
        \textbf{\rone} & 1.00 & 0.72 & 0.23 & 0.59  \\ 
        \textbf{\rtwo} & 0.72 & 1.00 & 0.21 & 0.58  \\ 
        \textbf{\CGPT} & 0.23 & 0.21 & 1.00 & 0.32  \\
        \textbf{\lama} & 0.59 & 0.58 & 0.32 & 1.00  \\ \hline

    \end{tabular}
    \label{human-llm-corr-p1}
\end{table}

A similar trend is observed in Table \ref{human-llm-corr-p7} for \CGPT in \Tseven, there is negligible positive correlation across all traits with human raters ($r_{\text{rater1}}=0.07 - 0.18, p< 0.001, d_{\text{overall}}=-2.77$ and $r_{\text{rater2}}=0.10-0.19, p< 0.001, d_{\text{overall}}=-2.71$). However, \lama shows a different trend than in \Tone. In \Tseven \lama's correlation also weak with both human raters ($r_{\text{rater1}}=0.27 - 0.38, p< 0.001, d_{\text{overall}}=-2.10$ and $r_{\text{rater2}}=0.28-0.40, p< 0.001, d_{\text{overall}}=-2.07$). Again, we observed no strong inter-LLM correlation ($r=0.20 - 0.30, p<0.001, d=0.26$) between \lama and \CGPT for \Tseven.

\begin{table}[h]
    \centering
    \captionsetup{width=0.9\textwidth}
    \caption{Correlations between human and LLM scores for \Tseven traits}
    \begin{tabular}{llccc}
        \textbf{Traits} & \textbf{LLM} & \textbf{Rater 1} & \textbf{Rater 2} & \textbf{ChatGPT} \\ \hline
         Ideas \& content & ChatGPT & 0.07 & 0.10 & 1.00  \\
               & \lama & 0.27 & 0.29 & 0.21  \\ \hline
        Organization & ChatGPT & 0.10 & 0.10 & 1.00  \\
                    & \lama & 0.32 & 0.36 & 0.24  \\ \hline
        Style & ChatGPT & 0.12 & 0.15 & 1.00  \\
                & \lama & 0.27 & 0.28 & 0.11  \\ \hline
        Convention & ChatGPT & 0.19 & 0.18 & 1.00  \\
                & \lama & 0.34 & 0.37 & 0.20  \\ \hline
        Overall & ChatGPT  & 0.18 & 0.19 & 1.00 \\ 
                & \lama  & 0.38 & 0.40 & 0.30  \\ \hline
    \end{tabular}
    \label{human-llm-corr-p7}
\end{table}


In summary, while human raters agree reasonably well with each other ($r=0.72, p<0.001, d=0.008$), LLMs (\lama and \CGPT) exhibit a weaker statistically significant alignment with both human raters, suggesting that \CGPT is not a suitable scoring model using zero-shot with rubric guideline prompts. Though in \Tone,  \lama shows some alignment with human scores, in \Tseven correlations remain weak ($r <= 0.4$) for both \CGPT and \lama, suggesting that both \lama and \CGPT diverge significantly from human raters. The lack of strong correlations underscores the challenge of automated scoring systems using LLMs in capturing nuanced writing qualities. The scoring differences indicate that \CGPT and \lama may have a different understanding of the rubric.

\subsubsection{ASAP++ traits: weak with \CGPT, strong with \lama} \label{traits-llm-donot-correlate}

Moving forward, we assess the relationship between various essay trait scores provided by human raters in the ASAP++ dataset for \Tone and scores given by human raters in the ASAP dataset and LLMs. Table \ref{asap-plus-corr-grader} demonstrates that all the ASAP++ trait scores given by human raters show a strong positive correlation with the overall scores provided by human raters in the ASAP dataset ($r_{\text{rater1}}=0.63-0.67, p< 0.001, d_{\text{mean}}=-0.56$ and $r_{\text{rater2}}=0.62-0.67, p< 0.001, d_{\text{mean}}=-0.58$). Moreover, we observed that trait scores are strongly correlated with \lama scores ($r=0.59-0.61, p<0.001, d_{\text{mean}}=0.16$) than \CGPT scores ($r=0.33-0.36, p<0.001, d_{\text{mean}}=2.35$). This finding further suggests that \CGPT and human raters may approach the essay scoring task very differently, and \lama can mimic human raters better than \CGPT. 

\begin{table}[!h]
    \centering
    \captionsetup{width=0.9\textwidth}
    \caption{ASAP++ traits weakly correlate with \CGPT but reasonably strong with \lama}
    \begin{tabular}{lcccc}
        \textbf{ Traits } & 
        \rotB{\textbf{ASAP  \rone}} & 
        \rotB{\textbf{ASAP \rtwo}} & 
        \rotB{\textbf{\CGPT}} & 
        \rotB{\textbf{\lama}} \\ \\ \hline
        \textbf{Ideas \& Content} & 0.66 & 0.67 & 0.34 & 0.61   \\ 
        \textbf{Organization} & 0.63 & 0.63 & 0.36 & 0.60   \\ 
        \textbf{Word Choice} & 0.67 & 0.67 & 0.33 & 0.59   \\ 
        \textbf{Sentence Fluency} & 0.64 & 0.62 & 0.36 & 0.60  \\ 
        \textbf{Conventions} & 0.63 & 0.62 & 0.35 & 0.59   \\ \hline
    \end{tabular}
    \label{asap-plus-corr-grader}
\end{table}

\subsection{RQ2: What are the possible reasons behind the similarity/difference in scores?}

To delve into the reasons behind the weak correlation between human and LLM scores, we have examined various essay features, including the count of language errors, readability indices and length-related attributes such as the number of sentences and tokens. More details on the extraction of these features are outlined in section \ref{sec-essay-feature-extraction}. All the detailed statistical analyses can be found in Appendix \ref{appendix-section-statistics}.

\subsubsection{Human scores highly correlate with essay features}

\begin{table}[!h]
    \centering
    \captionsetup{width=0.9\textwidth}
    \caption{Strong correlation between human scores and essay length }
    \begin{tabular}{llcccccc}
        \textbf{} & 
        \textbf{Score} & 
        \rotB{\textbf{Essay Length (sentences)}} & 
        \rotB{\textbf{Essay Length (tokens)}} & 
        \rotB{\textbf{FANBOYS (total)}} & 
        \rotB{\textbf{FANBOYS (unique)}} & 
        \rotB{\textbf{Transition Phrases (total)}} & \rotB{\textbf{Transition Phrases (unique)}} \\ \\
        \hline
        \textbf{} & \textbf{\rone} & \textbf{0.63} & \textbf{0.74} & 0.50 & 0.35 & 0.34 & 0.35 \\ 
       \textbf{\Tone } & \textbf{\rtwo} & \textbf{0.65} & \textbf{0.75} & 0.49 & 0.35 & 0.37 & 0.39 \\ 
        \textbf{} & \textbf{\CGPT} & 0.16 & 0.20 & 0.12 & 0.09 & 0.05 & 0.08 \\ 
        \textbf{} & \textbf{\lama} & 0.59 & \textbf{0.66} & 0.45 & 0.35 & 0.32 & 0.36 \\ 
         \hline
        \textbf{} & \textbf{\rone} & \textbf{0.61} & \textbf{0.62} & 0.39 & 0.31 & 0.36 & 0.43  \\ 
        \textbf{\Tseven} & \textbf{\rtwo} & \textbf{0.61} & \textbf{0.63} & 0.40 & 0.32 & 0.37 & 0.45 \\ 
        \textbf{} & \textbf{\CGPT} & -0.02 & 0.00 & -0.01 & 0.08 & 0.03 & 0.09  \\ 
        \textbf{} & \textbf{\lama} & 0.42 & 0.45 & 0.27 & 0.21 & 0.24 & 0.26 \\
    \label{essay-features-p1p7}
    \end{tabular}
\end{table}

From Table \ref{essay-features-p1p7}, it is evident that human scores exhibit a strong positive correlation with length-related essay features (number of tokens and sentences) for both tasks. This suggests that human raters tend to assign higher scores to longer essays. It goes without saying that better writers are capable of producing more high-quality sentences and longer sentences than average and weaker writers. Therefore, Occam's razor would lead us to believe that the length of the essay has no real bearing on its score. Instead, all that we observe is that good-quality essays are longer because they were written by better students who could accomplish more in the allotted time and so graded with higher scores. Additionally, we observed that while \lama scores show mostly a moderate correlation ($r=0.42-0.66$) with essay length features (both the number of tokens and sentences), including \Tone being an exception where the correlation is exceptionally high. \CGPT grades do not show any strong correlation ($r=0.0-0.20$) with essay features. This observation indicates that \CGPT is not lenient towards longer essays in scoring compared to the human raters and \lama. 

In analyzing the use of transition and ``FANBOYS" words showing Table \ref{essay-features-p1p7}, a weak to moderate positive correlation ($r=0.31-0.50$) has been found with human scores across both tasks. The higher correlation with the total count of ``FANBOYS" words, as opposed to the unique count, suggests that students tend to rely repeatedly on these connecting words (such as ``for, and, but, or, yet, so") in their writing. The more students join clauses and write complex structures of sentences the higher scores they get. Conversely, the higher correlation with the unique count of transition words compared to the total count indicates that employing diverse transition phrases may reflect better storytelling abilities, thus leading to higher scores. However, \lama exhibits a weak correlation ($r=0.21-0.45$), and \CGPT shows an even weaker correlation ($r=0.01-0.12$) compared to human raters, suggesting that, unsurprisingly, humans are better at evaluating the logical progression of ideas and narrative coherence through the adept use of these connecting words than LLMs.

\begin{table}[!h]
    \centering
    \captionsetup{width=1\textwidth}
    \caption{Strong correlation between ASAP++ traits and essay length for \Tone}
    \begin{tabular}{cccccccc}
        \textbf{Traits } & 
        \rotB{\textbf{Essay Length (sentences)}} & 
        \rotB{\textbf{Essay Length (tokens)}} &
        \rotB{\textbf{FANBOYS (total)}} & 
        \rotB{\textbf{FANBOYS (unique)}} & 
        \rotB{\textbf{Transition Phrases (total)}} & \rotB{\textbf{Transition Phrases (unique)}} \\ \\ \hline
    
        \textbf{Ideas \& Content} & 0.58 & 0.68 & 0.43 & 0.28 & 0.30 & 0.32 \\ 
        \textbf{Organization} & 0.54 & 0.64 & 0.40 & 0.26 & 0.28 & 0.30 \\ 
        \textbf{Word Choice} & 0.56 & 0.66 & 0.44 & 0.28 & 0.28 & 0.30 \\ 
        \textbf{Sentence Fluency} & 0.54 & 0.63 & 0.42 & 0.28 & 0.26 & 0.28 \\ 
        \textbf{Conventions} & 0.53 & 0.63 & 0.42 & 0.28 & 0.27 & 0.29 \\ 
        \hline
    \end{tabular}
    \label{asap-plus-corr-essay}
\end{table}

We extend our analysis with trait-wise scores in the ASAP++ dataset, examining their correlation with essay features. Table \ref{asap-plus-corr-essay} illustrates a moderate positive correlation between trait scores and the number of sentences ($r=0.53-0.58, p<0.001, d_{\text{mean}}=-3.05$), as well as a strong positive correlation with the number of tokens ($r=0.63-0.68, p<0.001, d_{\text{mean}}=-4.22$). This reaffirms the tendency for longer essays to receive higher scores from human raters. Furthermore, the observed trend in connected words aligns with the patterns seen in overall scores in the ASAP dataset, highlighting human raters' proficiency in recognizing narrative coherence and rewarding it with higher marks.


\subsubsection{LLM and human scores do not correlate with readability indices}

Here, we compare well-known readability indices with scores provided by LLMs and human raters. As demonstrated in Table \ref{grades-readability-corr-p1p7}, we can see that the scores provided by both human raters have very negligible correlations ($r=0.01-0.3$) across all well-known readability indices. This suggests a minimal influence of readability on human scoring.

\begin{table}[!h]
    \centering
    \captionsetup{width=0.9\textwidth}
    \caption{No correlations between scores and various readability indices}
    \begin{tabular}{llcccccccc}
        \textbf{} & \textbf{Score} &
        \rotB{\textbf{Flesch-Kincaid \\ Grade-level}} & 
        \rotB{\textbf{Flesch \\ Reading Ease}} & 
        \rotB{\textbf{Smog Index}} & 
        \rotB{\textbf{Coleman Liau}} & 
        \rotB{\textbf{Gunning Fog}} & 
        \rotB{\textbf{Automated Readability}} & 
        \rotB{\textbf{Linsear Write}} & 
        \rotB{\textbf{Dale Chall}} \\ \\ \hline
        
        \textbf{} &\textbf{\rone} & 0.02 & -0.17 & 0.16 & 0.30 & -0.19 & 0.01 & -0.10 & 0.16 \\ 
        \textbf{\Tone } &\textbf{\rtwo} & 0.01 & -0.15 & 0.14 & 0.28 & -0.21 & -0.01 & -0.12 & 0.16 \\ 
        \textbf{} &\textbf{\CGPT} & 0.10 & -0.23 & 0.18 & 0.17 & -0.09 & 0.00 & -0.08 & -0.01 \\ 
        \textbf{} &\textbf{\lama} & -0.01 & -0.19 & 0.13 & 0.24 & -0.25 & -0.08 & -0.19 & 0.05 \\ \hline
        \textbf{} &\textbf{\rone} & -0.19 & 0.13 & 0.16 & 0.11 & -0.30 & -0.14 & -0.21 & -0.06 \\ 
        \textbf{\Tseven} &\textbf{\rtwo} & -0.19 & 0.13 & 0.16 & 0.10 & -0.31 & -0.14 & -0.21 & -0.09 \\ 
        \textbf{} &\textbf{\CGPT} & 0.09 & -0.20 & 0.18 & 0.11 & -0.04 & 0.01 & -0.04 & -0.21 \\ 
       \textbf{} & \textbf{\lama} & -0.02 & -0.09 & 0.21 & 0.18 & -0.15 & -0.03 & -0.12 & -0.05 \\ 
    \end{tabular}
    \label{grades-readability-corr-p1p7}
\end{table}

Interestingly, the tables also demonstrate a lack of correlation between LLM scores and readability metrics. Since we are using LLMs as a zero-shot learner and since we do not mention readability explicitly in the prompt (See Appendix \ref{secA1} for an example), this result implies that LLM's interpretation of the question and its response are not associated with the numerous mentions to readability found on the web (and very likely used in the pre-training of the LLM models). Furthermore, some correlations with all scoring methods are negative, indicating that both humans and LLMs may assign high grades to less readable texts and low grades to highly readable ones.

\subsubsection{LLM scores negatively correlate with the count of mistakes} 

We utilized LanguageTool and Aspell to detect various language mistakes in the essays, including grammar, spelling, style, punctuation, and capitalization, as described in section \ref{sec-essay-feature-extraction}. Following this, we computed the number of mistakes for all essays and conducted correlation analysis across four scoring methods. In Table \ref{neg-corr-llm-mistake-p1p7}, we observed that \CGPT consistently displayed a weak but negative correlation, whereas other scorers mostly showed a positive correlation with mistake count.

\begin{table}[h]
    \centering
    \caption{Negative correlation between LLM scores and mistake counts}
    \begin{tabular}{llcccccc}
    \textbf{} &
        \diagbox[width=3cm, height=1.3cm]{Score}{ Mistake count }&
        \rotB{\textbf{Aspell \\Mispelling}} &
        \rotB{\textbf{\ltool Grammar}} & 
        \rotB{\textbf{\ltool Spelling}} & 
        \rotB{\textbf{\ltool Style}} & 
        \rotB{\textbf{\ltool Punctuation}} & 
        \rotB{\textbf{\ltool Capitalization}} \\ \\ \hline
        \textbf{} &\textbf{\rone} & 0.14 & 0.09 & 0.11 & 0.09 & 0.02 & 0.11 \\ 
        \textbf{\Tone } &\textbf{\rtwo} & 0.16 & 0.14 & 0.13 & 0.10 & 0.03 & 0.12 \\ 
        \textbf{} &\textbf{\CGPT} & \textbf{-0.12} & \textbf{-0.11} & \textbf{-0.15} & \textbf{-0.02} & \textbf{-0.01} & \textbf{-0.08} \\ 
        \textbf{} &\textbf{\lama} & 0.04 & 0.05 & 0.02 & 0.05 & 0.05 & 0.11 \\ \hline
        
        \textbf{} &\textbf{\rone} & 0.11 & 0.05 & 0.20 & 0.11 & 0.11 & 0.05 \\ 
        \textbf{\Tseven} &\textbf{\rtwo} & 0.10 & 0.06 & 0.21 & 0.09 & 0.10 & 0.05 \\ 
         \textbf{} & \textbf{\CGPT} & \textbf{-0.24} & \textbf{-0.17} & \textbf{-0.21} & \textbf{-0.05} & \textbf{-0.05 }& \textbf{-0.17} \\ 
         \textbf{} &\textbf{\lama} & \textbf{-0.10} & \textbf{-0.03} & 0.08 & 0.06 & 0.01 & \textbf{-0.11} \\
    \hline
    \end{tabular}
    \label{neg-corr-llm-mistake-p1p7}
\end{table}

The negative correlation observed with mistakes aligns with the rubric, where a higher number of language mistakes typically corresponds to lower scores. This logical relationship underscores \CGPT's reliability in accurately identifying mistakes and adjusting scores accordingly. Additionally, \lama also demonstrated a negative correlation in some instances, albeit with mostly negligible positive scores. Conversely, human graders exhibited a surprising positive correlation with mistake count. This suggests potential challenges for human graders in consistently identifying and adjusting grades based on spelling and grammar errors, highlighting the systematic ability of LLMs like \CGPT and \lama in language error detection. On the other hand, this may be attributed to potential inaccuracies in the misspelling count reported by the spelling checker tools. To investigate this matter, we manually inspected the words flagged by \textit{Aspell} as misspelled in randomly selected essays (approx 10) and cross-checked with another spell checker \textit{LanguageTool}. We have also performed a correlation analysis of these two spell checkers, details can be found in section \ref{aspell-ltool-section}. Our findings indicate that it strongly agrees with \textit{Aspell} that the likelihood of the last scenario occurring is unlikely.

We extended our analysis by examining the relationship between mistake count and trait-wise scores in the ASAP++ dataset. In Table \ref{asap-plus-corr-ltool}, we observed a negative correlation between the \ltool's grammar, spelling, and punctuation error counts and trait scores, although the correlation was very weak. Specifically, for the convention trait score, most mechanical error counts displayed a negative correlation. This indicates that trait scores offer detailed insights pinpointing specific areas for skill enhancement, particularly in identifying mechanical mistakes with greater precision than overall scoring by human raters.

\begin{table}[!h]
    \centering
    \captionsetup{width=0.9\textwidth}
    \caption{ASAP++ trait scores negatively correlate with mistake count}
    \begin{tabular}{lccccccc}
        \diagbox[width=3cm, height=1.5cm]{Score}{ Mistake count }&
        \rotB{\textbf{Aspell \\Mispelling}} &
        \rotB{\textbf{\ltool Grammar}} & 
        \rotB{\textbf{\ltool Spelling}} & 
        \rotB{\textbf{\ltool Style}} & 
        \rotB{\textbf{\ltool Punctuation}} & 
        \rotB{\textbf{\ltool Capitalization}} \\ \\ \hline
        
        \textbf{Ideas \& Content}& 0.04 & -0.01 & -0.01 & 0.06 & -0.03 & 0.05 \\ 
        \textbf{Organization}& 0.01 & -0.02 & -0.03 & 0.04 & -0.02 & 0.04 \\ 
        \textbf{Word Choice}& 0.03 & -0.02 & -0.02 & 0.04 & -0.05 & 0.04 \\ 
        \textbf{Sentence Fluency}& -0.05 & -0.05 & -0.09 & 0.02 & -0.02 & 0.00\\ 
        \textbf{Conventions}&-0.05 & -0.04 & -0.09 & 0.02 & -0.01 & -0.00 \\ 
    \end{tabular}
    \label{asap-plus-corr-ltool}
\end{table}

\subsection{RQ3: Do LLMs offer explanations in a tone that reflects their scores?}
\subsubsection{\CGPT gives consistently harsh explanation}
 We have extracted basic statistical features and applied the VADER sentiment analyzer from the \CGPT explanation as described in section \ref{explanation-extraction}. In table \ref{chatgpt-feedback-corr-p1p7}, explanation length (sentences) and explanation length (tokens) refer to the number of sentences and the number of words found in \CGPT explanation respectively. Explanation sentiment (average) refers to the overall compound sentiment score of \CGPT response provided by VADER. Explanation sentiment (max) means sentence-wise maximum compound score in \CGPT response provided by VADER sentiment analyzer. We perform correlation analysis between these four features with human and LLM scores.

\begin{table}[!h]
    \centering
    \caption{Harsh tendency in \CGPT's explanation }
    \begin{tabular}{llp{1.75cm}p{1.75cm}p{1.75cm}p{1.75cm}}
    \textbf{} & \raisebox{-1.25em}{\textbf{Score}} &  \textbf{Explanation Length (sentences)} & 
    \textbf{Explanation Length (tokens)} & \textbf{Explanation Sentiment (average)} & 
    \textbf{Explanation Sentiment (max)} \\ \hline
    
        \textbf{} & \textbf{\rone} & 0.12 & 0.17 & 0.06 & 0.04 \\ 
        \textbf{\Tone } & \textbf{\rtwo} & 0.11 & 0.15 & 0.06 & 0.04  \\ 
        \textbf{} & \textbf{\CGPT} & 0.33 & 0.45 & \textbf{0.33} & \textbf{0.28}  \\ 
        \textbf{} & \textbf{\lama} & 0.14 & 0.19 & 0.14 & 0.12  \\ \hline
    
        \textbf{} & \textbf{\rone} & 0.05 & 0.04 & 0.05 & 0.02  \\ 
        \textbf{\Tseven} & \textbf{\rtwo} & 0.06 & 0.04 & 0.05 & 0.01  \\ 
        \textbf{} & \textbf{\CGPT} & 0.28 & 0.30 & \textbf{0.24} & \textbf{0.13}  \\ 
        \textbf{} & \textbf{\lama} & 0.13 & 0.13 & 0.08 & 0.02  \\ 
    \end{tabular}
    \label{chatgpt-feedback-corr-p1p7}
\end{table}

It's worth noting that the explanation features show a positive weak correlation, but they are relatively stronger with \CGPT's scores compared to human-assigned or \lama scores. This suggests that \CGPT has a better understanding of the prompt and provides explanations that align with the assigned scores, indicating consistency. For instance, lower scores receive more negative explanations, while higher scores get more positive ones. This implies that \CGPT is not generating explanations randomly but is aware of the scores it has provided.

However, we observed that \CGPT scores have smaller coefficient scores ($r=0.13-0.33$), indicating that \CGPT may not be explained in a positive tone. Considering that \CGPT's overall scores are quite low and LLMs have a tendency to generate toxic content~\cite{stochastic_parrots}, it's possible that the tone of the explanations may not appear positive due to the low scores. Since we cannot rule out the possibility that the explanation generated by \CGPT could have been influenced by the score it produced, there might be a side effect of using LLMs as zero-shot classifiers, affected by their tendency towards generating toxic content (despite concerted efforts by LLM developers to curb this tendency).

\subsubsection{\lama gives moderately positive explanation}
We conducted a comprehensive analysis of the \lama explanations using the VADER sentiment analyzer, similar to our approach with the \CGPT explanations. The results detailed in Table \ref{lama-feedback-corr-p1p7}, reveal that the \lama scores exhibit notably higher correlation ($r=0.42-0.54$) with its explanations. This finding suggests that in contrast to \CGPT, \lama generates explanations with a moderately positive sentiment highlighting \lama's ability to convey positive explanations to learners.

\begin{table}[!h]
    \centering
    \caption{\lama's moderately positive explanation}
    \begin{tabular}{llp{1.75cm}p{1.75cm}p{1.75cm}p{1.75cm}}
    \textbf{} &  \raisebox{-1.25em}{\textbf{Score}} &  \textbf{Explanation Length (sentences)} & 
    \textbf{Explanation Length (tokens)} & \textbf{Explanation Sentiment (average)} & 
    \textbf{Explanation Sentiment (max)} \\ \hline
    
        \textbf{} & \textbf{\rone} & 0.26 & 0.26 & 0.30 & 0.24 \\ 
        \textbf{\Tone } & \textbf{\rtwo} & 0.25 & 0.27 & 0.29 & 0.22 \\ 
        \textbf{} & \textbf{\CGPT} & 0.10 & 0.11 & 0.24 & 0.20 \\ 
        \textbf{} & \textbf{\lama} & 0.28 & 0.31 & \textbf{0.54} & \textbf{0.42} \\ 
        \hline
        \textbf{} & \textbf{\rone}  & 0.16 & 0.21 & 0.26 & 0.22 \\ 
        \textbf{\Tseven} & \textbf{\rtwo}  & 0.13 & 0.17 & 0.25 & 0.22 \\ 
        \textbf{} & \textbf{\CGPT}  & 0.11 & 0.13 & 0.13 & 0.14 \\ 
        \textbf{} & \textbf{\lama}  & 0.27 & 0.30 & \textbf{0.50} & \textbf{0.44} \\ 
    \end{tabular}
    \label{lama-feedback-corr-p1p7}
\end{table}

Furthermore, we observed a consistent trend where the correlation between \lama scores and its explanations outweighs that of any other scoring methods, underscoring the consistency and reliability of the explanations provided by \lama in relation to their assigned scores. This alignment highlights the coherent explanations that correspond closely with their evaluations, demonstrating a promising aspect of using \lama in educational assessment tasks.

\subsection{RQ4: Can LLMs correctly identify and assess spelling and grammatical mistakes and score accordingly?} \label{section-RQ4}

To investigate whether the mention of mistakes (misspellings and grammatical errors) in the explanation reflects the quality of the essays, we have exploited the LLM explanation as described in \ref{rule-based-matching} further. Now we have the information for each essay sample on whether LLM has identified any language mistakes. Next, we categorized these LLM responses (for \Tone and \Tseven of the ASAP dataset) based on the mention of grammatical and spelling errors separately into three distinct groups. The first group contains samples without mention of spelling errors, the second group comprises only samples with mentions of misspellings, and the last group contains essay samples with mentions of misspellings along with a qualifier (such as \textit{several, numerous, multiple}). We followed this grouping process for grammatical errors for both \CGPT and \lama. For these three groups, we have calculated the average of spelling mistakes by \ltool and Aspell, LLM scores, and sentiment score of LLM explanation and compared them against each other. We determined whether the differences between group means are statistically significant by conducting the ANOVA test and reported p-values and F-stat.
We have also calculated effect sizes using Omega squared ($\omega^2$)
\footnote{Cohen (1988) classified $\omega^2$ effect sizes as small ($\omega^2=0.01$), medium ($\omega^2 = 0.059$), and large ($\omega^2 \geq 0.138$)}, which is widely recognized as a less biased alternative to eta-squared, particularly when dealing with small sample sizes.

\begin{table}[ht!]
    \centering
    \captionsetup{width=1\textwidth}
    \caption{Changes in average number of misspellings, sentiment scores of LLM explanations and LLM scores across different misspelling categories \Tone}
    \begin{tabular}{llccccc}
        \multicolumn{1}{p{2.3cm}}{\textbf{Misspelling Category}} & 
        \rotB{\textbf{Sample \\Count}} & 
        \rotB{\textbf{Aspell \\ Misspelling}} & 
        \rotB{\textbf{\ltool Spelling}} & 
        \rotB{\textbf{LLM Score}} &
        \rotB{\textbf{Explanation \\ Sentiment (average)}} &
        \rotB{\textbf{Explanation \\ Sentiment (max)}}\\ \\\hline\hline
        \multicolumn{7}{c}{\CGPT}\\
        \hline\hline
        (1) No mention & 1546 & 8.16 & 9.26 & 1.91 & -0.21 & -0.02 \\ 
        (2) Unqualified mention & 65 & 12.14 & 13.42 & 1.91 & -0.43 & -0.04 \\ 
        (3) Qualified mention & 172 & 13.53 & 14.98 & 1.78 & -0.49 & -0.26 \\ \hline
        F-stat $(2,1780)$  & & 59.37 & 58.43 & 3.7 & 24.25 & 12.23\\ 
        P-value & & $p<0.05$ & $p<0.05$ & $p<0.05$ & $p<0.05$ & $p<0.05$ \\ 
        Effect size ($\omega^2$) & & 0.06 & 0.06 & 0.004 & 0.03 & 0.01 \\
        \hline\hline
        \multicolumn{7}{c}{\lama}\\
        \hline\hline
        (1) No mention & 1598 & 8.58 & 9.71 & 3.63 & 0.79 & 0.69 \\ 
        (2) Unqualified mention & 148 & 9.80& 10.95 & 3.77 & 0.84 & 0.67 \\ 
        (3) Qualified mention & 37 & 15.23 & 17.00 & 2.62& 0.23 & 0.17 \\ \hline
        F-stat $(2,1780)$ & & 19.08 & 19.86 & 48.65 & 29.25 & 20.08 \\ 
        P-value & & $p<0.05$ & $p<0.05$ & $p<0.05$ & $p<0.05$ & $p<0.05$ \\ 
        Effect size ($\omega^2$) & & 0.02 & 0.02 & 0.05 & 0.03 & 0.02 \\ 
    \end{tabular}
    \label{misspellcategory-p1}
\end{table}

\begin{table}[ht!]
    \centering
    \captionsetup{width=1.0\textwidth}
    \caption{Changes in average number of misspellings, explanation sentiment scores and LLM scores across different misspelling categories \Tseven}
    \begin{tabular}{llccccc}
        \multicolumn{1}{p{2.2cm}}{\textbf{Misspelling Category}} & 
        \rotB{\textbf{Sample \\Count}} & 
        \rotB{\textbf{Aspell \\ Misspelling}} & 
        \rotB{\textbf{\ltool Spelling}} & 
        \rotB{\textbf{LLM Score}} &
        \rotB{\textbf{Explanation \\ Sentiment (average)}} &
        \rotB{\textbf{Explanation \\ Sentiment (max)}}\\ \\
        
        \hline\hline
        \multicolumn{7}{c}{\CGPT}\\
        \hline\hline
        (1) No mention & 917 & 3.53 & 5.6 & 2.32 & -0.60 & -0.27  \\ 
        (2) Unqualified mention & 203 & 4.08 & 6.3 & 2.44 & -0.57 & -0.24 \\ 
        (3) Qualified mention & 449 & 5.18 & 7.4 & 1.39 & -0.65 & -0.30 \\ \hline
        F-stat $(2,1566)$  & & 22.87 & 17.22 & 45.76 & 3.11 & 1.03\\ 
        P-value & & $p<0.05$ & $p<0.05$ & $p<0.05$ & $p<0.05$& $p > 0.05$\\ 
        Effect size ($\omega^2$) & & 0.03 & 0.02 & 0.05 & 0.004 & 0.001 \\
        \hline\hline
        \multicolumn{7}{c}{\lama}\\
        \hline\hline
        (1) No mention & 667 & 3.39 & 5.42 & 3.19 & -0.28 & -0.13 \\ 
        (2) Unqualified mention & 339 & 4.03 & 6.55 & 3.37 & -0.30 & -0.13 \\ 
        (3) Qualified mention & 563 & 4.92 & 6.92 & 1.60 & -0.60 & -0.37 \\  \hline
        F-stat $(2,1566)$ & & 20.06 & 12.86 & 75.13 & 46.04 & 32.69 \\ 
        P-value & &$p<0.05$&$p<0.05$ & $p<0.05$ & $p<0.05$ & $p<0.05$ \\ 
        Effect size ($\omega^2$) & & 0.02 & 0.02 & 0.09 & 0.06 & 0.04 \\ 
    \end{tabular}
    \label{misspellcategory-p7}
\end{table}

As shown in Tables \ref{misspellcategory-p1} and \ref{misspellcategory-p7} for \Tone and \Tseven respectively, both Aspell and LanguageTool consistently report an increase in the average number of misspellings across these groups. This logical trend aligns with expectations: the first group, with no mention of mistakes, naturally contains fewer mistakes than the second group, where mistakes are mentioned in the explanation. Similarly, the third group, which includes qualifiers indicating multiple mistakes, is expected to have more mistakes than the other two groups. This finding underscores the fact that both LLMs (\CGPT and \lama) pay close attention to the presence of spelling mistakes in essays when delivering explanations. Additionally, this highlights the careful use of qualifiers (e.g., several, numerous, many) in their explanation generation especially when there are more misspellings. 

However, while this focus on misspellings may not significantly affect the scores assigned by \CGPT and \lama, as seen in the small to no difference in average scores between the first two groups. In most cases, there is a statistically significant difference in average LLM scores between the first and third groups as shown in the pairwise comparisons in Appendix \ref{appendix-section-statistics} (Table D42 - D53). This suggests that the LLMs don't just assign scores randomly all the time, but they can assess evaluating the essay quality to some extent.

Additionally, the decrease in VADER sentiment scores of LLM's explanation is observed, which logically aligns with expectations, as the presence of more mistakes and lower scores tends to result in harsher sentiment in the explanation. This correspondence between sentiment and mistakes indicates that as more mistakes accumulate in student essays, the explanations provided by both \CGPT and \lama tend to adopt a more negative tone, demonstrating their ability to adjust explanations appropriately according to essay quality. However, while \lama sentiment scores also decrease with essay mistakes, they remain relatively positive compared to \CGPT.

In Tables \ref{grammaticalcategory-p1} and \ref{grammaticalcategory-p7}, we observe a similar trend as seen for the misspelling categories. All the results depicted in these tables are statistically significant and can have small to medium effects practically. This reaffirms that LLMs can identify spelling and grammatical mistakes and evaluate essays accordingly.

\begin{table}[ht!]
    \centering
    \captionsetup{width=1\textwidth}
    \caption{Changes in average number of grammar mistakes, sentiment scores and LLM score across different grammatical categories \Tone }
    \begin{tabular}{lccccc}
        \multicolumn{1}{p{2.3cm}}{\textbf{ Grammar \mbox{Error Category}}} & 
        \rotB{\textbf{Sample \\Count}} & 
        \rotB{\textbf{\ltool Grammar}} & 
        \rotB{\textbf{LLM Score }}& 
        \rotB{\textbf{Explanation \\ Sentiment (average)}} &
        \rotB{\textbf{Explanation \\ Sentiment (max)}}\\ \\
        
        \hline\hline
        \multicolumn{6}{c}{\CGPT}\\
        \hline\hline
        (1) No mention &1169 & 2.93 & 1.89 & -0.17 & 0.02\\ 
        (2) Unqualified mention & 230& 3.06 & 1.99 &  -0.35 & -0.12  \\ 
        (3) Qualified mention & 384 & 3.51 & 1.85 & -0.43 & -0.22 \\ 
        \hline
        F-stat $(2,1780)$ & & 8.07 & 4.43 & 39.52 & 28.38  \\ 
        P-value & & $p<0.05$ & $p<0.05$ & $p<0.05$ & $p<0.05$ \\ 
        Effect size ($\omega^2$) & & 0.01 & 0.041 & 0.04 & 0.03 \\ 
        \hline\hline
        \multicolumn{6}{c}{\lama}\\
        \hline\hline
        (1) No mention &  1422 & 3.02 & 3.64 & 0.80 & 0.69 \\ 
        (2) Unqualified mention & 298 & 3.15 & 3.72 & 0.82 & 0.67 \\ 
        (3) Qualified mention & 63 &3.94 & 2.70 & 0.34 & 0.29  \\  \hline
        F-stat $(2,1780)$ & & 4.42 & 70.38 & 31.62 & 20.04  \\ 
        P-value & & $p<0.05$ & $p<0.05$ & $p<0.05$ & $p<0.05$ \\ 
        Effect size ($\omega^2$) & & 0.004 & 0.07 & 0.03 & 0.02 \\ 
    \end{tabular}
    \label{grammaticalcategory-p1}
\end{table}

\begin{table}[ht!]
    \centering
    \captionsetup{width=1.0\textwidth}
    \caption{Changes in average number of grammar mistakes, sentiment scores and LLM score across different grammatical categories \Tseven}
    \begin{tabular}{lccccc}
        \multicolumn{1}{p{2.5cm}}{\textbf{Grammar \mbox{Error Category}}} & 
        \rotB{\textbf{Sample \\Count}} & 
        \rotB{\textbf{\ltool Grammar}} & 
        \rotB{\textbf{LLM Score }}& 
        \rotB{\textbf{Explanation \\ Sentiment (average)}} &
        \rotB{\textbf{Explanation \\ Sentiment (max)}}\\ \\
        \hline\hline
        \multicolumn{6}{c}{\CGPT}\\
        \hline\hline
        (1) No mention &332 & 1.08 & 2.35 & -0.56 & -0.26 \\ 
        (2) Unqualified mention & 386 & 1.10 & 2.76 & -0.54 & -0.21 \\ 
        (3) Qualified mention & 851 & 1.47 & 1.65 & -0.66 & -0.31 \\ \hline
        F-stat $(2,1566)$ & & 11.7 & 57.7 & 13.11 & 5.22  \\ 
        P-value & & $p<0.05$ & $p<0.05$ & $p<0.05$ & $p<0.05$\\ 
        Effect size ($\omega^2$) & & 0.01 & 0.07 & 0.02 & 0.001 \\ 
        \hline\hline
        \multicolumn{6}{c}{\lama}\\
        \hline\hline
        (1) No mention & 802 & 1.17 & 3.25 & -0.30 & -0.15 \\ 
        (2) Unqualified mention & 303 & 1.33 & 3.00 & -0.37 & -0.17 \\ 
        (3) Qualified mention & 464 & 1.48 & 1.41 & -0.58 & -0.36  \\  \hline
        F-stat $(2,1566)$ & & 6.04 & 79.11 & 28.58 & 21.65  \\ 
        P-value & & $p<0.05$ & $p<0.05$ & $p<0.05$ & $p<0.05$\\ 
        Effect size ($\omega^2$) & & 0.01 & 0.09 & 0.03 & 0.03 \\ 
    \end{tabular}
    \label{grammaticalcategory-p7}
\end{table}

Furthermore, we categorized essay samples into three groups based on the scores provided by human raters, with each group representing a range of average scores from two raters. Tables \ref{humancategory-p1} and \ref{humancategory-p7} illustrate that LLMs assign lower scores to lower-grade groups and relatively higher scores to higher-grade groups. This logical trend suggests that LLMs may possess the potential to effectively assess essay quality. However, further fine-tuning and the implementation of additional techniques that enable contextual learning may enhance their performance in grading tasks.

\begin{table}[h!]
    \centering
    \captionsetup{width=1\textwidth}
    \caption{Comparison of average grades assigned by LLMs across human rater grade categories \Tone }
    \begin{tabular}{lccc}
    \multicolumn{1}{p{1.3cm}}{\textbf{Human Score Class}} & 
        \rotB{\textbf{Sample \\Count}} &
        \rotB{\textbf{ChatGPT Score}} & 
        \rotB{\textbf{\lama \\ Score }} \\  \\ \hline
        Score 1-2 & 28 & 1.21 & 1.68  \\ 
        Score 2-4 & 949 & 1.82 & 3.39 \\ 
        Score 4-6 & 806 & 2.01 & 3.95 \\ 
        \hline
        F-stat(2,1780) & & 46.46 & 399.13 \\
        P-values & & $p < 0.001$ & $p < 0.001$ \\
        Effect size ($\omega^2$) & & 0.05 & 0.31\\
    \end{tabular}
    \label{humancategory-p1}
\end{table}

\begin{table}[h!]
    \centering
    \captionsetup{width=1\textwidth}
    \caption{Comparison of average grades assigned by LLMs across human rater grade categories \Tseven}
    \begin{tabular}{lccc}
    \multicolumn{1}{p{1.3cm}}{\textbf{Human Score Class}} & 
        \rotB{\textbf{Sample \\Count}} &
        \rotB{\textbf{ChatGPT Score}} & 
        \rotB{\textbf{\lama \\ Score }} \\  \\ \hline
        Score 0-4 & 108 & 0.84 & 0.89  \\ 
        Score 4-8 & 717 & 1.93 & 1.93 \\ 
        Score 8-15 & 744 & 2.38 & 3.62 \\ 
        \hline
        F-stat(2,1566) & & 39.05 & 111.2 \\
        P-values & & $p < 0.001$ & $p < 0.001$ \\
        Effect size ($\omega^2$) & & 0.05 & 0.1\\
    \end{tabular}
    \label{humancategory-p7}
\end{table}

\subsection{Summary with Canonical Correlation}

Canonical correlation analysis (CCA) is used to identify and measure the associations between two sets of variables. It does it by determining a set of canonical variates (orthogonal linear combinations of the variables within each set) that best explain the correlation both within and between sets. For example, we have two matrices $X = (x_1, …, x_n)$ with n features, and $Y = (y_1, …,y_m)$ with m features. CCA will first find a linear combination of variables in $X$ (let’s call it U1) and a linear combination of variables in Y (let’s call it V1) so that U1 and V1 are maximally correlated with each other. These linear combinations are called canonical variates. 

\begin{figure}[!h]%
    \centering
    \captionsetup{justification=centering}
    \includegraphics[width=1\textwidth]{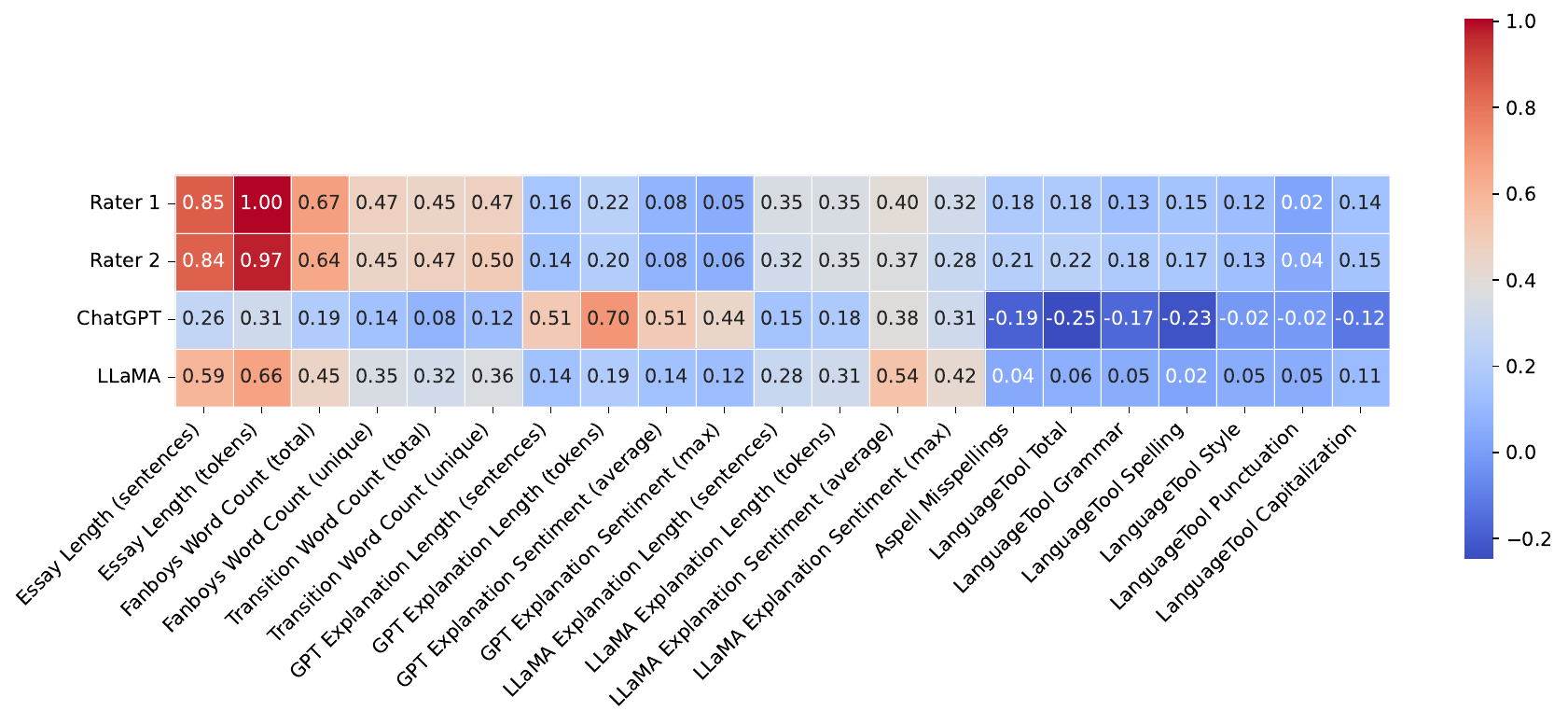}
    \caption{Canonical Correlation of Task 1}
    \label{canonical-corr-p1}
\end{figure}

We have considered all the extracted features in one group and performed canonical correlation analysis with the scores provided by LLMs and human raters. Figure \ref{canonical-corr-p1} illustrates which variables from both groups have the most influence on their first canonical variate. Our analysis revealed consistent findings with our previous experiments, where we considered individual features independently. We observed distinct differences in how human raters and LLMs approach the essay scoring task, with \CGPT showing the most disparity compared to \lama, which demonstrated closer alignment with human grading. 

Notably, there was a strong association between human rater scores and essay length features on the first canonical variate, a pattern also observed with \lama. Interestingly, \CGPT scores showed a positive correlation with the sentiment of its own explanation, suggesting that \CGPT can generate text explanations congruent with the scores it assigns.

\section{Results with Prompt Engineering} \label{sec-prompt-eng}

We delve deeper into our research findings with additional experiments and offer new interpretations of the results.  We explore the impact of prompt engineering by incorporating grade-level information and few-shot examples to enhance the effectiveness of our approach. 

To explore the impact of different prompt formulations on the correlation between LLM scores and human rater scores, we conducted experiments using subsets of samples from our dataset. To ensure balance across different score ranges, we selected samples from each score point. For \Tone, which has a score range of 1 to 6, we selected a total of 100 essay samples (20 samples from each score point from 2 to 6). Since Score 1 had a minimal number of samples compared to the overall distribution, we excluded it from our selection. Details of the data distribution can be found in Figure \ref{score-dist-p1}. Similarly, for \Tseven, which has a score range of 0 to 15, we observed that Scores 0 to 3 had very few samples, while the maximum score given by human raters was 12. Thus, we selected 20 samples each from Scores 3 to 12, totalling 200 samples for \Tseven. We performed analyses on these samples by varying the prompt information, as described in the following subsections. Appendix \ref{secA1} includes a detailed prompt design for ChatGPT and \lama to generate the response.

\subsection{Providing Students' Grade level} 
As shown in Table \ref{grade-level-adding}, there is a slight increase in correlation scores in most cases, indicating a potential improvement with the addition of grade levels in the prompt. This suggests that including grade levels may help ChatGPT to provide less harsh evaluations. Interestingly, for \lama, we observed a decrease or no change in correlation for \Tone, suggesting it may have been lenient previously. However, for \Tseven, we observed an increasing trend in correlation scores.

\begin{table}[!ht]
    \centering
    \captionsetup{width=0.9\textwidth}
    \caption{Changes in correlation scores after adding grade level of students to the prompt (Increase denoted by and $\uparrow$ and decrease denoted by $\downarrow$)}
    \begin{tabular}{|c|c|cc|cc|}
    \hline
        \textbf{} & \textbf{} & \textbf{\Tone  } & (100 samples) & \textbf{\Tseven} & (200 samples) \\ \hline
        \textbf{} & ~ & Before & After & Before & After \\  \hline
        \textbf{} & Rater 1 & 0.36 & {0.40} $\uparrow$ & 0.28 & {0.37} $\uparrow$ \\ 
        \textbf{ChatGPT} & Rater 2 & 0.39 & {0.41} $\uparrow$ & 0.23 & {0.36} $\uparrow$ \\ 
        \textbf{} & \lama & 0.43 & {0.50} $\uparrow$ & 0.36 & {0.45} $\uparrow$ \\  \hline
        \hline
        \textbf{} & Rater 1 & 0.79 & {0.71} $\downarrow$ & 0.46 & {0.59} $\uparrow$ \\ 
        \textbf{\lama} & Rater 2 & 0.80 & {0.76} $\downarrow$ & 0.51 & {0.60} $\uparrow$ \\ 
        \textbf{} & ChatGPT & 0.43 & 0.43 & 0.36 & {0.37} $\uparrow$ \\ 
    \hline
    \end{tabular}
    \label{grade-level-adding}
\end{table}

\subsection{Few-shot learning}
After incorporating two-shot examples, we noticed a significant enhancement in the correlation between \CGPT and human scoring as shown in Table \ref{fewshot-adding}. These examples were carefully chosen to align with human scores, with one representing a lower score range and the other a higher score range. However, we encountered limitations with \lama, as it failed to respond when the prompt exceeded approximately 2.5k tokens, even though it did not reach the 4k token context length limit. Consequently, we experimented with using only one-shot examples to avoid this issue. For \lama, we do not see much difference if we try one-shot learning.
\begin{table}[!ht]
    \centering
    \captionsetup{width=0.9\textwidth}
    \caption{Changes in correlation scores after adding few-shot examples to the prompt (Increase denoted by and $\uparrow$ and decrease denoted by $\downarrow$)}
    \begin{tabular}{|c|c|cc|cc|}
    \hline
        \textbf{} & \textbf{} & \textbf{\Tone} & (100 samples) & \textbf{\Tseven} & (200 samples) \\ \hline
        \textbf{} & ~ & Before & After & Before & After \\  \hline
        \textbf{} & Rater 1 & 0.36 & {0.65} $\uparrow$ & 0.28 & {0.50} $\uparrow$ \\ 
        \textbf{ChatGPT (Two-shot)} & Rater 2 & 0.39 & {0.59} $\uparrow$ & 0.23 & {0.47} $\uparrow$ \\ 
        \textbf{} & \lama & 0.43 & {0.64} $\uparrow$ & 0.36 & {0.54} $\uparrow$ \\  \hline
        \hline
        \textbf{} & Rater 1 & 0.79 & {0.69} $\downarrow$ & 0.49 & {0.52} $\uparrow$\\ 
        \textbf{\lama (One-shot)} & Rater 2 & 0.80 & {0.71} $\downarrow$ & 0.51 & {0.56} $\uparrow$ \\ 
        \textbf{} & ChatGPT & 0.43 & {0.30} $\downarrow$ & 0.36 & {0.41} $\uparrow$\\ 
    \hline
    \end{tabular}
    \label{fewshot-adding}
\end{table}

\section{Conclusions} \label{conclusion}
Our research aims to evaluate the effectiveness of Large Language Models (LLMs) in assessing essay quality, generally, and in Automatic Essay Scoring (AES), specifically. By comparing scores generated by LLMs to those given by human raters, we have uncovered several insights into the performance and alignment of these models.

We found that while \CGPT and \lama exhibit remarkable capabilities in understanding essay prompts and generating self-consistent numerical and textual responses, there is a significant disparity between their assessments and those of human raters. Human raters show strong inter-rater agreement, often awarding higher scores to longer essays and excelling in evaluating the logical progression of ideas. In contrast, LLMs demonstrate distinct grading behaviors, with \CGPT  being stricter in its scoring compared to both human raters and \lama. On the other hand, \lama shows a closer alignment to human scoring patterns.

Both human and LLM scores seem minimally influenced by readability indices; however, essays that were harder to read tended to receive higher marks from both humans and LLMs. LLMs demonstrated a strong ability to detect and account for spelling and grammatical errors, highlighting a key difference from human raters, who seem to overlook such clerical mistakes in grading. The reason for this behaviour is impossible to obtain from the data at our disposal. Nevertheless, our findings suggest that future AES research needs to use richer data that can help capture the nuances of the human grading process. However, specific trait-wise human scores have shown promise in identifying mechanical mistakes more effectively than an overall score provided by human raters.

In terms of providing explanations for the grades, \CGPT often delivered explanations in a harsh tone, whereas \lama used generally less negative language in their output. Both LLMs showed a better understanding of the prompt and provided explanations that aligned with the assigned scores, indicating a degree of consistency. Lower scores are accompanied by more negative explanations, while higher scores have led to more positive explanations, suggesting that LLMs are aware of the scores they provide and do not generate explanations randomly.

Overall, our findings suggest that while LLMs hold potential for automatic essay scoring (AES) tasks, they should be used with human supervision. LLMs may not completely replace human raters, but their ability to understand guidelines, coupled with their consistent explanation capabilities, make them valuable tools for educators facing the demands of modern education.
Additionally, our experiment with one of the more recent LLMs \lamatri reveals that in \Tseven, \lamatri demonstrates a 29.5\% improvement in correlation with human scores compared to \lama-2 and a remarkable 173\% improvement compared to ChatGPT-3.5-Turbo. In Task 1, although \lamatri is 16\% less correlated with human scores than \lama-2, it still achieves a 130\% improvement in correlation with human raters compared to ChatGPT-3.5-Turbo. Overall, \lamatri shows substantial advancements in performance compared to ChatGPT-3.5-Turbo across both tasks, despite some variability compared to \lama-2.

\subsection{Limitations and Future work} 

One limitation of this study is that we evaluated the explanations generated by LLMs based solely on their ability to justify the provided grade, without considering how these explanations might affect learners. This narrow focus overlooks the potential impact of formative feedback on student learning and motivation, which is crucial for educational applications.

Another limitation is the quality and nature of the dataset used. The dataset primarily contains numeric scores, which tend to align with basic length features of the essays. This suggests that human graders may prioritize certain aspects of the rubric over others. For example, longer essays with numerous spelling mistakes often receive higher scores from human raters, potentially because they value coherence and interest over technical accuracy. This inconsistency indicates that the human graders' evaluations may not perfectly adhere to the rubric, which could lead to misleading conclusions when comparing human and LLM scores. 

Additionally, LLMs tend to apply the rubric strictly and may not compensate for errors in the same way human graders do. This could explain why LLMs, particularly \CGPT, tend to be harsher and less aligned with human scores. Human graders might overlook certain mistakes in favor of overall essay quality, while LLMs strictly follow the rubric guidelines. Therefore, comparing LLM scores with human scores from the ASAP dataset, as done in our study, or training machine learning models to replicate these scores, as done in previous studies, could be misguided. Such comparisons might not accurately reflect the nuanced judgments made by human graders.

Future research should investigate the impact of LLM-generated explanations on student revisions and writing improvement. Additionally, our study focused on \CGPT and \lama; future work should include comparisons with other models like Google Gemini \cite{team2023gemini} and similar tools to provide a more comprehensive evaluation of automatic scoring in education. Further exploration can be done on the effects of fine-tuning models or prompting LLMs with more annotated scores and explanations data by human raters to better mimic human grading practices and provide valuable textual insights. It is also important to note that the performance of AI tools is subject to their development stage at the time of the experiment. As these tools continue to evolve with new training data, future studies should reassess their effectiveness and alignment with human grading to ensure continued relevance and accuracy in educational assessments.

\section{Conflict of Interest} 

All authors declare that we have no conflicts of interest to disclose related to this research. 
This article is based primarily on the first author, Anindita Kundu’s Masters thesis. Partial financial support was awarded through the Deepmind scholarship program as part of her graduate studies.

\bibliographystyle{abbrvnat}
\bibliography{bibliography}

\begin{appendices}

\section{Sample Prompt and Output from \CGPT and \lama}\label{secA1}
\subsection{\Tone  Question (given to the students)} \label{tas1-ques}
\begin{lstlisting}
More and more people use computers, but not everyone agrees that this benefits society. Those who support advances in technology believe that computers have a positive effect on people. They teach hand-eye coordination, give people the ability to learn about faraway places and people, and even allow people to talk online with other people. Others have different ideas. Some experts are concerned that people are spending too much time on their computers and less time exercising, enjoying nature, and interacting with family and friends. 

Write a letter to your local newspaper in which you state your opinion on the effects computers have on people. Persuade the readers to agree with you.
\end{lstlisting}

\subsection{\Tone  Rubric} \label{tas1-rubric}
\begin{lstlisting}
Score Point 1: An undeveloped response that may take a position but offers no more than very minimal support. Typical elements:
-   Contains few or vague details.
-   Is awkward and fragmented.
-   May be difficult to read and understand.
-   May show no awareness of audience.

Score Point 2: An under-developed response that may or may not take a position. Typical elements:
-       Contains only general reasons with unelaborated and/or list-like details.
-       Shows little or no evidence of organization.
-       May be awkward and confused or simplistic.
-       May show little awareness of audience.

Score Point 3: A minimally-developed response that may take a position, but with inadequate support and details. Typical elements:
-       Has reasons with minimal elaboration and more general than specific details.
-       Shows some organization.
-       May be awkward in parts with few transitions.
-       Shows some awareness of audience.

Score Point 4: A somewhat-developed response that takes a position and provides adequate support. Typical elements:
-       Has adequately elaborated reasons with a mix of general and specific details.
-       Shows satisfactory organization.
-       May be somewhat fluent with some transitional language.
-       Shows adequate awareness of audience.

Score Point 5: A developed response that takes a clear position and provides reasonably persuasive support. Typical elements:
-       Has moderately well elaborated reasons with mostly specific details.
-       Exhibits generally strong organization.
-       May be moderately fluent with transitional language throughout.
-       May show a consistent awareness of audience.

Score Point 6: A well-developed response that takes a clear and thoughtful position and provides persuasive support. Typical elements:
-       Has fully elaborated reasons with specific details.
-       Exhibits strong organization.
-       Is fluent and uses sophisticated transitional language.
-       May show a heightened awareness of audience.
\end{lstlisting}

\subsection{\Tone  Prompt (Zero-shot)}
\begin{lstlisting}
# Question: {Question Appendix A.1}
Given the criteria: {Rubric Appendix A.2}
 
How would you grade the following essay? Provide a short explanation. 

Here is the essay
Essay: People are probly on the computers chating there e-mail or making a websit or they are on youtube. Maybe the person does not like to go outside and enjoy nature or spending time with there family there are probly checking out new websites and playing games or the problay on the comper to book there vactions of seeing what is on sell a welmart. They are probly looking at or buy a new truk or they are looking at the history on all the wars and how they were started.

Note: Please ensure that the score and explanation are formatted like the example given below. 
Score: [Score according to your evaluation] 
Explanation: [Your detailed feedback].

Please select a score between 1 and 6 for the essay based on the provided criteria and explain your reasoning in detail.

# Answer:
\end{lstlisting}

\subsection{\Tone : Prompting with the grade level of students}\label{prompt}

\begin{lstlisting}
System prompt: You are a helpful essay grading assistant. Answer the question and follow the instructions carefully.

# Question: {Question Appendix A.1}
Given the criteria: {Rubric Appendix A.2}

How would you grade the following essay? This essay is written by a student of grade 8. Provide a short explanation. 

Here is the essay
Essay: People are probly on the computers chating there e-mail or making a websit or they are on youtube. Maybe the person does not like to go outside and enjoy nature or spending time with there family there are probly checking out new websites and playing games or the problay on the comper to book there vactions of seeing what is on sell a welmart. They are probly looking at or buy a new truk or they are looking at the history on all the wars and how they were started.

Note: Please ensure that the score and explanation are formatted like the example given below. 
Score: [Score according to your evaluation] 
Explanation: [Your detailed feedback].

You need to grade according to the expected level of grade 8 writing skills. Please select a score between 1 and 6 for the essay based on the provided criteria and explain your reasoning in detail.

# Answer: 
\end{lstlisting}

\subsection{\Tone : Few-shot learning}
\begin{lstlisting}
# Question: {Question Appendix A.1}
Given the criteria: {Rubric Appendix A.2}
 

* Example Essay 1: 
Coumputors are mind browning on what they can do. They can make or break somone financhely. They also can perduce amasing graphis that aritist cant perduce. Coumputrs are also for great for recer antal acitivitys, and they also take your life away. I not talking about a real life @CAPS1 talking about the activitys you do away from home. Now lets see the benifits and problems we come so far on how we handle money. We use to put it in chets and hide them now all we do is click and type in a securty number. people can also invist in items but invisting is tricky you dont always win. You can win small or lose big they say @CAPS2 coumputers today are hd. The screens are made of a liquid crystal fourmula that hardens. The lights in the back pixels do most of the wort to bring stunning graphis to life. On coumputers today people make money by selling graphis they made on coumputers. So this would be another plus on what a coumputor dose to a human. I went to a school to lern how to perduce great graphis. All teens kids and adults have used a coumputer for recerantol acivitys. On todays coumputors you can play games watch videos, and listen on coumputors. People do this to engoy then selfs. Some people ran get in trouble for fooling around on the web. But there is nothing stressed out and just playing some games for a while, personly I like to go on youtube and find funny vidos after school. I use to own a youtube account and get many views. The worst thing about coumpurts is it can ruin plan gobo or somones socal life some people get addicted and cant stop being on the coumputor until it is takin away. For example some people play world of war craft get addicted and cant stop! I had a problem when I had to chek you tube a lot. Well I hoped you readers. No matter what happens I think coumputors bennafit more than the problems
Human rater score 1: 4
Example Answer 1: 
Score: 4
Explanation: The essay presents a clear position on the effects of computers on people, arguing that computers have both positive and negative effects. The writer provides some reasons and details to support their position, such as the ability to learn new skills, produce amazing graphics, and enjoy recreational activities. However, they also acknowledge the potential negative effects, such as addiction and social isolation.
The organization of the essay is satisfactory, with a clear introduction, body, and conclusion. The writer uses some transitional phrases to connect their ideas, but the transitions could be smoother and more cohesive. The essay shows some awareness of the audience, with the writer addressing the readers directly and using a conversational tone.
One area for improvement is the use of language. The essay contains several spelling and grammar errors, which detract from the overall quality of the writing. Additionally, the writer could benefit from providing more specific and detailed examples to support their points. For instance, they could describe a particular skill they learned from using a computer or provide an example of a successful graphic design project.
Overall, the essay is somewhat developed but could benefit from more detail and refinement in the writing. Therefore, a score of 4 is appropriate.


* Example Essay 2: 
The computers are cool. Do you now I werpsite is @NUM1.com. We nede to time the lemt lo-menteno it. I love the compueters. We nede @NUM2 make a computers are grene compurers.
Human rater score 2: 1
Example Answer 2: 
Score: 1
Explanation:The essay provided is undeveloped and lacks clear reasoning or persuasive support. It contains very minimal details and is awkward and fragmented, making it difficult to read and understand. The essay does not provide any specific examples or reasons to support the author's claim that computers are "cool". The language used is also inappropriate and contains spelling and grammatical errors. The essay does not show any awareness of the audience or attempt to persuade the readers to agree with the author's position. Therefore, it scores a 1. 
It is worth noting that this essay is difficult to read and understand due to its poor writing quality. Therefore, it is not possible to provide detailed feedback on its content.


How would you grade the following essay? Provide a short explanation. 

Here is the essay
Essay: People are probly on the computers chating there e-mail or making a websit or they are on youtube. Maybe the person does not like to go outside and enjoy nature or spending time with there family there are probly checking out new websites and playing games or the problay on the comper to book there vactions of seeing what is on sell a welmart. They are probly looking at or buy a new truk or they are looking at the history on all the wars and how they were started.

Note: Please ensure that the score and explanation are formatted like the example given below. 
Score: [Score according to your evaluation] 
Explanation: [Your detailed feedback].

Please select a score between 1 and 6 for the essay based on the provided criteria and explain your reasoning in detail.

# Answer: 
\end{lstlisting}
\subsection{\Tseven Question (given to the students)}
\begin{lstlisting}
Write about patience. Being patient means that you are understanding and tolerant. A patient person experience difficulties without complaining. 

Do only one of the following: write a story about a time when you were patient OR write a story about a time when someone you know was patient OR write a story in your own way about patience.
\end{lstlisting}

\subsection{\Tseven Rubric}
\begin{lstlisting}
A rating of 0-3 on the following four traits:
# Ideas (points doubled)
Score 3: Tells a story with ideas that are clearly focused on the topic and are thoroughly developed with specific, relevant details. 
Score 2: Tells a story with ideas that are somewhat focused on the topic and are developed with a mix of specific and/or general details. 
Score 1: Tells a story with ideas that are minimally focused on the topic and developed with limited and/or general details. Score 0: Ideas are not focused on the task and/or are undeveloped.

# Organization
Score 3: Organization and connections between ideas and/or events are clear and logically sequenced. 
Score 2: Organization and connections between ideas and/or events are logically sequenced. 
Score 1: Organization and connections between ideas and/or events are weak. 
Score 0: No organization evident.

# Style
Score 3: Command of language, including effective and compelling word choice and varied sentence structure, clearly supports the writer's purpose and audience. 
Score 2: Adequate command of language, including effective word choice and clear sentences, supports the writer's purpose and audience. 
Score 1: Limited use of language, including lack of variety in word choice and sentences, may hinder support for the writer's purpose and audience.
Score 0: Ineffective use of language for the writer's purpose and audience.

# Conventions
Score 3: Consistent, appropriate use of conventions of Standard English for grammar, usage, spelling, capitalization, and punctuation for the grade level. 
Score 2: Adequate use of conventions of Standard English for grammar, usage, spelling, capitalization, and punctuation for the grade level. 
Score 1: Limited use of conventions of Standard English for grammar, usage, spelling, capitalization, and punctuation for the grade level.
Score 0: Ineffective use of conventions of Standard English for grammar, usage, spelling, capitalization, and punctuation.
\end{lstlisting}

\subsection{\Tseven: Prompting with the grade level of students}
\begin{lstlisting}
# Question: {Question Appendix A.6}
Given the criteria: {Rubric Appendix A.7}

How would you grade the following essay? This essay is written by a student of grade 7. Provide a short explanation. 

Here is the essay
Essay: I am patent i always have to wait for somthing or earn it. I go to my grouping with my parents but I have to wat. I have to wait to go to my aunts. The worst time to wat is going to @CAPS1/a up @LOCATION1 and camping.

Note: Please ensure that the score and explanation are formatted like the example given below. 
Score: [Score according to your evaluation] 
Explanation: [Your detailed feedback].

You need to grade according to the expected level of grade 7 writing skills. Please select a score between 0 and 15 for the essay based on the provided criteria and explain your reasoning in detail.
\end{lstlisting}

\subsection{\Tseven: Few-shot Learning}
\begin{lstlisting}
# Question: {Question Appendix A.6}
Given the criteria: {Rubric Appendix A.7}

* Example Essay 1: 
Being patient, what does it mean? It means that; "you are understanding and tolerant. A patient person experiences difficulties without complaining." @CAPS1, being a patient person is difficult. One time, it was hard for me to be a patient person, I didn't complain though. I was overjoyed to be going to a water park on my birthday. I wanted to party and get my swim on but it was almost @NUM1 weeks away. I didn't want to wait any longer, but I knew that being patient would make my experience even better. I was tolerant, understanding, and patient. The week dragged on. I wanted to complain and rip my hair out because it was like the earth was turning @NUM2, @NUM3 times slower, but I knew that I couldn't! When the thought of me going to the water park got stuck in my brain like glue, the problem was clear. If I didn't think about it, time would fly by again. I had to be patient because I didn't want to annoy anyone with my complaints. When the day came, I had a blast. Being patient always turns out good!

Human rater provided score on this essay: ideas: 2 organization: 2 style: 2 conventions: 2 total: 8
Example Answer 1: 
Score: ideas: 2 organization: 2 style: 2 conventions: 2 total: 8
Explanation: Ideas: The essay tells a story about a time when the writer had to be patient. It focuses on the topic and provides some specific details, such as wanting to go to a water park on their birthday. However, the ideas could have been more thoroughly developed with more specific and relevant details. Organization: The essay has a clear introduction and conclusion. The events are logically sequenced, with the writer describing their desire to go to the water park, the challenges they faced in being patient, and the outcome of their patience. However, there could have been more cohesion and transitions between the events. Style: The writer demonstrates an adequate command of language. They use some effective word choices and clear sentences. However, there lacks variety in word choice and sentence structure, and the language could have been more compelling to support the writer's purpose and audience. Conventions: The writer shows an adequate use of conventions of Standard English. There are some errors in grammar, usage, spelling, capitalization, and punctuation, but they do not significantly hinder the reader's understanding. Overall, the essay meets the basic requirements but could benefit from more development in ideas, stronger organization and transitions, more compelling language, and improved conventions.

* Example Essay 2: 
I was patient on the trip to @LOCATION1 it took two days to get down there I got hungery and had to go to the bathroom when ever I was hungery I had to what until next stop when we got gas so when it was on full I had to be patient with out complaining. So, I had to be patient until the gas tank was empty that's when we got gas and I could eat something or I had to be patient until there was a @CAPS1 on the right hand or a test area because the have a vending machine there and bathrooms there too. On the right hand side because we can get off on get back on the high way. That's when I had to be patient.

Human rater provided score on this essay: ideas: 1 organization: 1 style: 1 conventions: 1 total: 4
Example Answer 2: 
Score: ideas: 1 organization: 1 style: 1 conventions: 1 total: 4
Explanation: The ideas presented in the essay are minimally focused and underdeveloped. There is limited detail provided to support the topic of patience. Additionally, the organization of the essay is weak, as there is no clear sequencing of events or logical connections between ideas. The style of writing is limited, with lack of variety in word choice and sentence structure. There are also several grammatical errors and lack of proper punctuation throughout the essay. Overall, the essay lacks coherence and fails to effectively convey the concept of patience.

How would you grade the following essay? Provide a short explanation. 

Here is the essay
Essay: Patience is usually good or bad for me. Here are a few reasons why patience can be hard or stresful. If you are waiting to get a shot or life threat results. A way to avoid normal patience like being board is to get some notes or school work to do or an ipod. Patience is everywere. For example carrides, birthday parties, vacation commercial, school being over, restaurants, holidays or whenever you say I cannot wait, and etc. to sam up a few ways about patience in my way of description.

Note: Please ensure that the score and explanation are formatted like the example given below. 
Score: [Score according to your evaluation] 
Explanation: [Your detailed feedback].

Please select a score between 0 and 15 for the essay based on the provided criteria and explain your reasoning in detail.

# Answer: 
\end{lstlisting}
\subsection{Response from OpenAI ChatGPT:}\label{response}
\begin{lstlisting}
Score: 1
Explanation: This essay is an undeveloped response that offers very minimal support. The essay lacks organization and coherence. The language is fragmented and contains many grammatical errors. The ideas are unclear and the details provided are vague and nonspecific. The essay does not demonstrate any awareness of the audience. Overall, it is difficult to read and understand.
\end{lstlisting}
\subsection{Response from \lama:}\label{response}
\begin{lstlisting}
Score: 1
Explanation: The essay provides minimal support for the position that people are spending too much time on their computers and not enough time on other activities. The essay merely states that people are probably using their computers for various activities without providing any specific examples or evidence. There is no clear organization, and the writing is awkward and difficult to read in some parts. The essay shows little awareness of the audience and does not attempt to persuade the readers to agree with the writer's position. Therefore, a score of 1 is appropriate for this essay.
\end{lstlisting}

\newpage

\section{\Tseven trait-wise detail information}\label{secB}
\subsection{\Tseven trait-wise descriptive statistics}
\begin{table}[!ht]
    \centering
    \caption{Descriptive statistics \Tseven: Ideas}
    \begin{tabular}{lllll}
        \textbf{} & \textbf{Rater 1 } & \textbf{Rater 2 } & \textbf{ChatGPT } & \textbf{\lama } \\ \hline
        \textbf{Mean} & 1.84 & 1.85 & 0.6 & 0.88 \\ 
        \textbf{SD} & 0.86 & 0.88 & 0.6 & 0.96 \\ 
        \textbf{Min} & 0.00 & 0.00 & 0.0 & 0.00 \\ 
        \textbf{Median} & 2.00 & 2.00 & 1.0 & 1.00 \\ 
        \textbf{Max} & 3.00 & 3.00 & 2.0 & 8.00 \\ 
    \end{tabular}
\end{table}

\begin{table}[!ht]
    \centering
    \caption{Descriptive statistics \Tseven: Organization}
    \begin{tabular}{lllll}
        \textbf{} & \textbf{Rater 1 } & \textbf{Rater 2 } & \textbf{ChatGPT } & \textbf{\lama } \\ \hline
        \textbf{Mean} & 2.02 & 2.03 & 0.57 & 0.89 \\ 
        \textbf{SD} & 0.72 & 0.72 & 0.79 & 0.87 \\ 
        \textbf{Min} & 0.00 & 0.00 & 0.00 & 0.00 \\ 
        \textbf{Median} & 2.00 & 2.00 & 0.00 & 1.00 \\ 
        \textbf{Max} & 3.00 & 3.00 & 5.00 & 6.00 \\ 
    \end{tabular}
\end{table}

\begin{table}[!ht]
    \centering
    \caption{Descriptive statistics \Tseven: Style}
    \begin{tabular}{lllll}
        \textbf{} & \textbf{Rater 1 } & \textbf{Rater 2 } & \textbf{ChatGPT } & \textbf{\lama } \\ \hline
        \textbf{Mean} & 1.99 & 2.00 & 0.37 & 0.22 \\ 
        \textbf{SD} & 0.61 & 0.64 & 0.51 & 0.67 \\ 
        \textbf{Min} & 0.00 & 0.00 & 0.00 & 0.00 \\ 
        \textbf{Median} & 2.00 & 2.00 & 0.00 & 0.00 \\ 
        \textbf{Max} & 3.00 & 3.00 & 2.00 & 7.00 \\ 
    \end{tabular}
\end{table}

\begin{table}[!ht]
    \centering
    \caption{Descriptive statistics \Tseven: Conventions}
    \begin{tabular}{lllll}
        \textbf{} &\textbf{Rater 1 } & \textbf{Rater 2 } & \textbf{ChatGPT } & \textbf{\lama } \\ \hline
        \textbf{Mean} & 2.17 & 2.17 & 0.59 & 0.73 \\ 
        \textbf{SD} & 0.69 & 0.69 & 0.61 & 0.63 \\ 
        \textbf{Min} & 0.00 & 0.00 & 0.00 & 0.00 \\ 
        \textbf{Median} & 2.00 & 2.00 & 1.00 & 1.00 \\ 
        \textbf{Max} & 3.00 & 3.00 & 5.00 & 5.00 \\ 
    \end{tabular}
\end{table}

\subsection{\Tseven trait-wise score distribution}
\begin{figure}[H]%
    \centering
    \captionsetup{justification=centering}
    \includegraphics[width=0.75\textwidth]{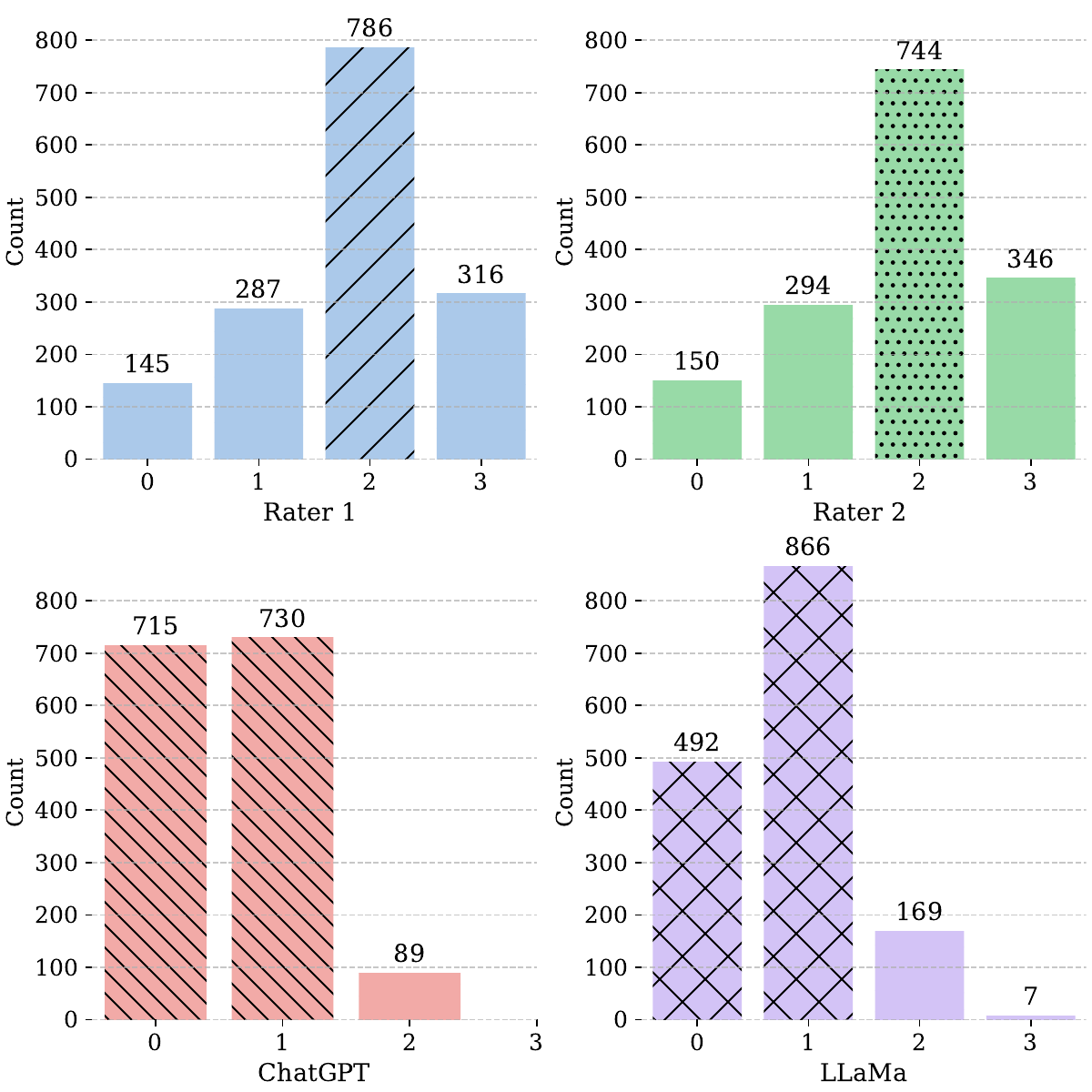}
    \caption{Score distribution of human raters and LLMs for \Tseven (trait ideas)}
\end{figure}

\begin{figure}[H]%
    \centering
    \captionsetup{justification=centering}
    \includegraphics[width=0.75\textwidth]{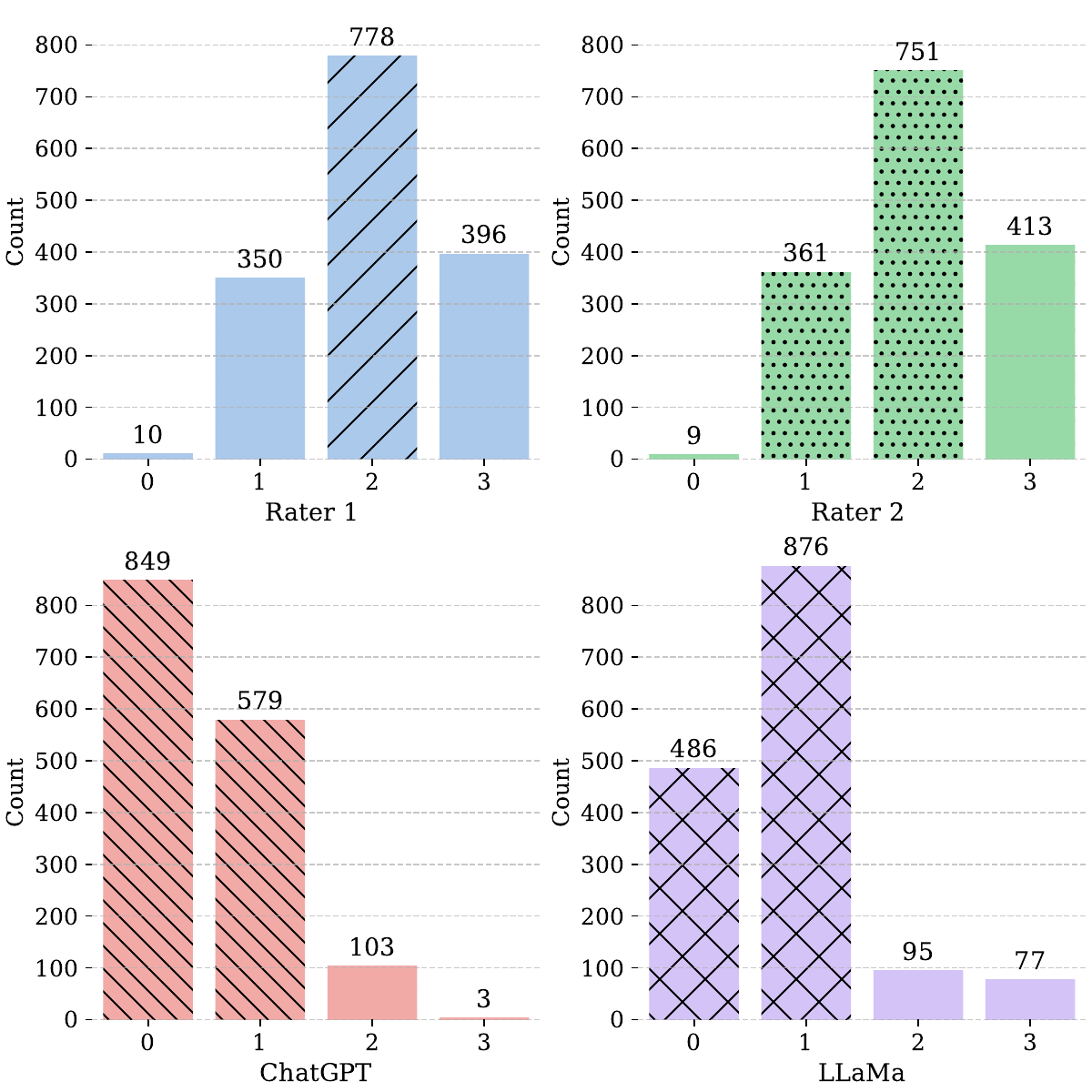}
    \caption{Score distribution of human raters and LLMs for \Tseven (trait organization)}
\end{figure}

\begin{figure}[H]%
    \centering
    \captionsetup{justification=centering}
    \includegraphics[width=0.75\textwidth]{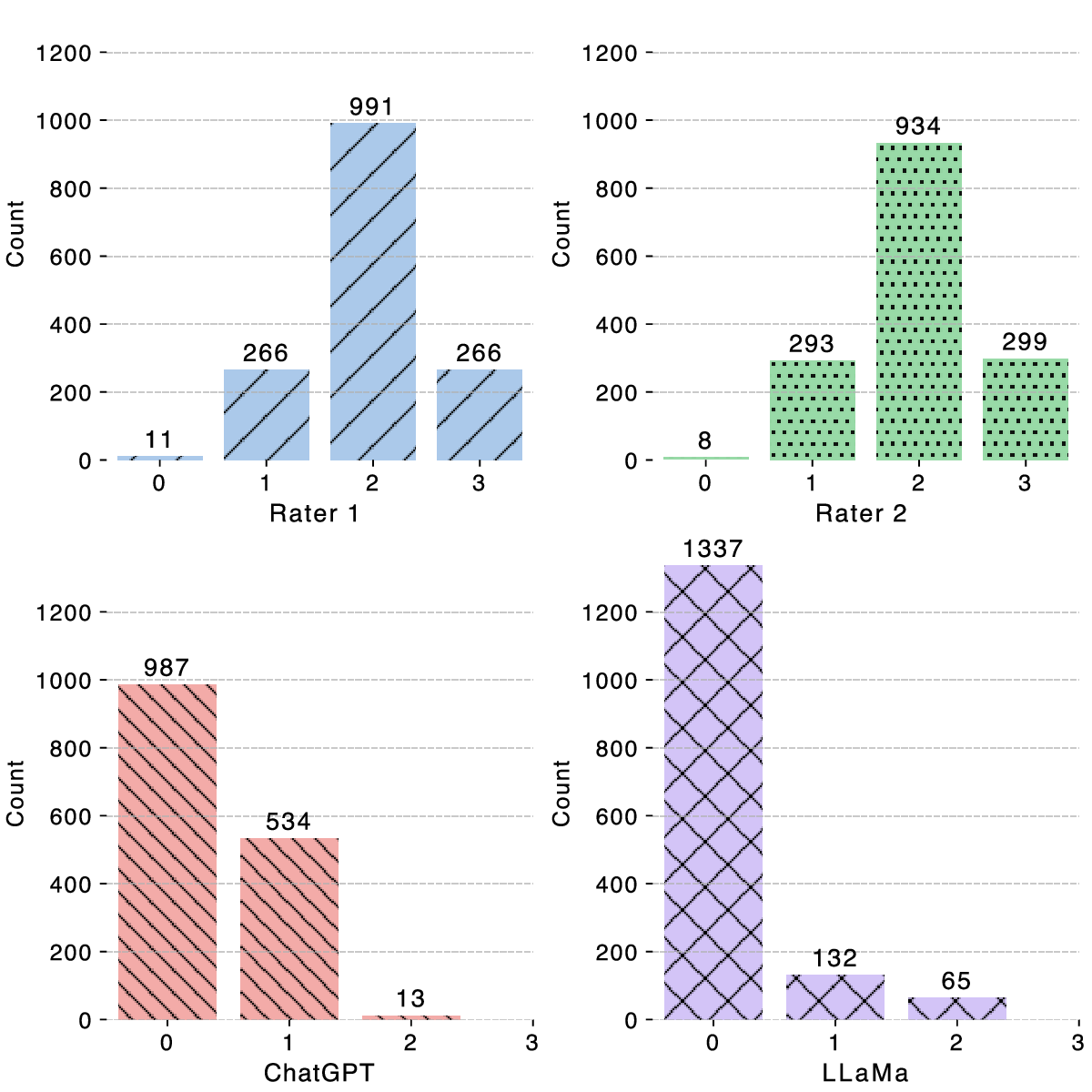}
    \caption{Score distribution of human raters and LLMs for \Tseven (trait style)}
\end{figure}

\begin{figure}[H]%
    \centering
    \captionsetup{justification=centering}
    \includegraphics[width=0.75\textwidth]{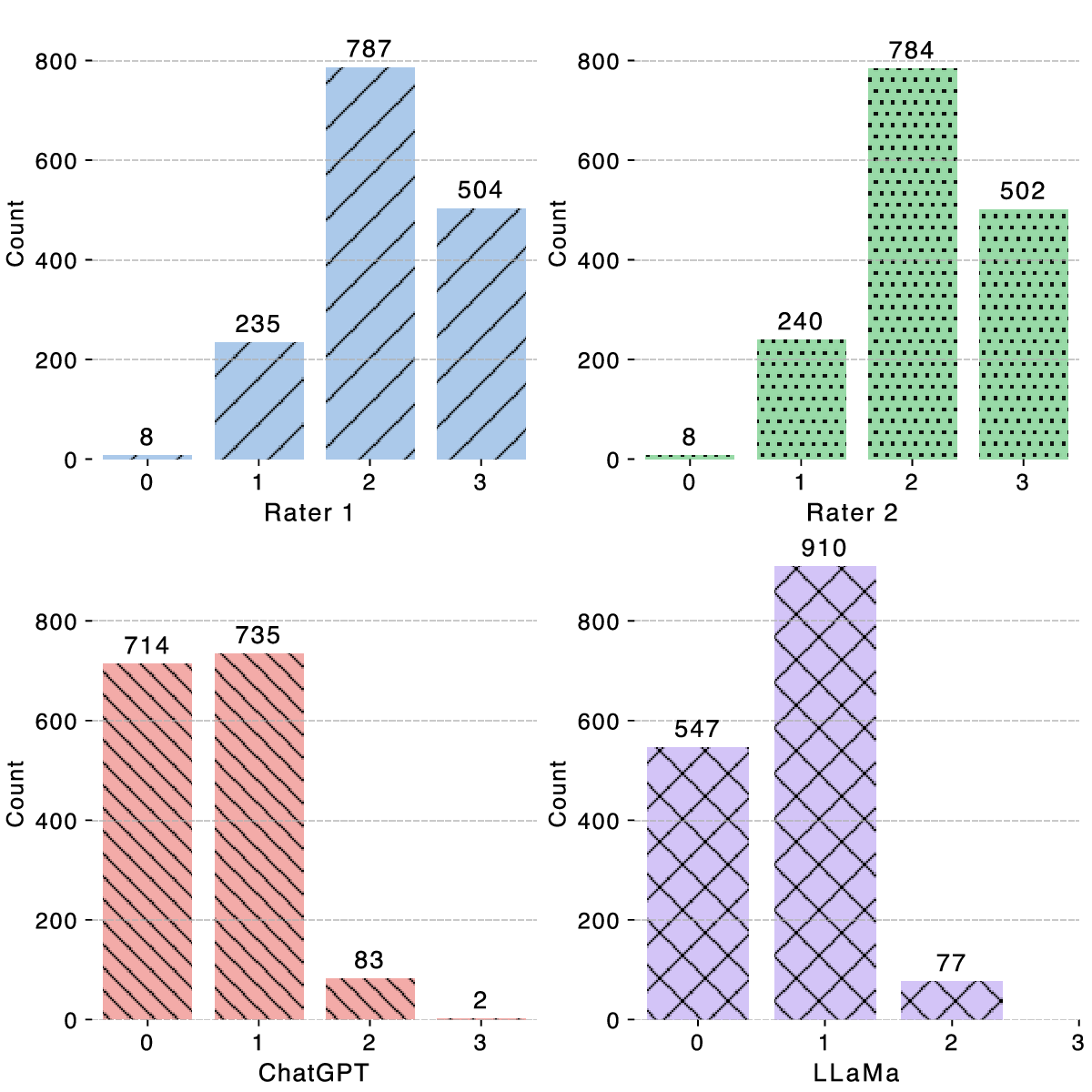}
    \caption{Score distribution of human raters and LLMs for \Tseven (trait conventions)}
\end{figure}

\section{Additional Results with \lamatri} \label{llama3-results}

In this section, we consider one of the more recent and most powerful large language models which is \lamatri. This model was not available when we started our study, so we decided to present them separately in this section for better clarity. Towards the end of our research, new and much larger language models such as GPT-4 \cite{achiam2023gpt}, {\lamatri} \cite{llama3modelcard} became available. Llama 3 70B can be up to 50 times more cost-effective and 10 times faster than GPT-4\footnote{\url{https://www.vellum.ai/blog/llama-3-70b-vs-gpt-4-comparison-analysis}}. GPT-4 still has advantages in scenarios that need longer context or special features like image support and function calling \cite{achiam2023gpt}. However, for many tasks, Llama 3 70B is catching up as a strong competitor and producing comparable results \cite{open-llm-leaderboard}. In the interest of time, we were able to run \textit{Meta-Llama-3-70B-Instruct} \footnote{\url{https://huggingface.co/meta-llama/Meta-Llama-3-70B-Instruct}} lately because we could deploy it locally. In this section, we are going to revisit some of our findings now using Llama 3 70B.

\subsection{RQ1: Do human scores align with the LLM scores? }

\begin{figure}[H]%
    \centering
    \captionsetup{justification=centering}
    \includegraphics[width=0.75\textwidth]{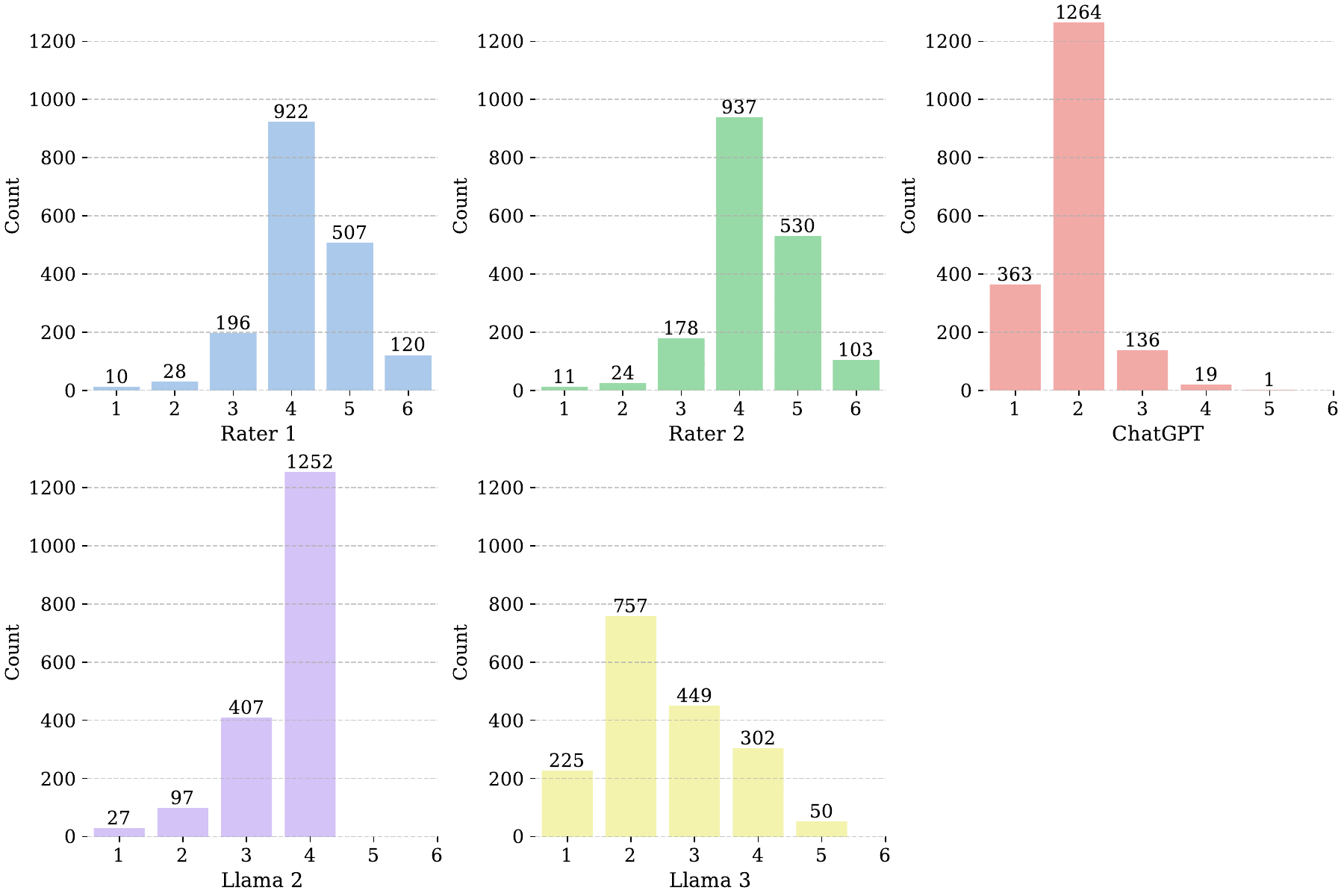}
    \caption{Score distribution of human raters and LLMs including \lamatri for \Tone }
    \label{llama3-score-dist-p1}
\end{figure}

\begin{figure}[H]%
    \centering
    \captionsetup{justification=centering}
    \includegraphics[width=0.75\textwidth]{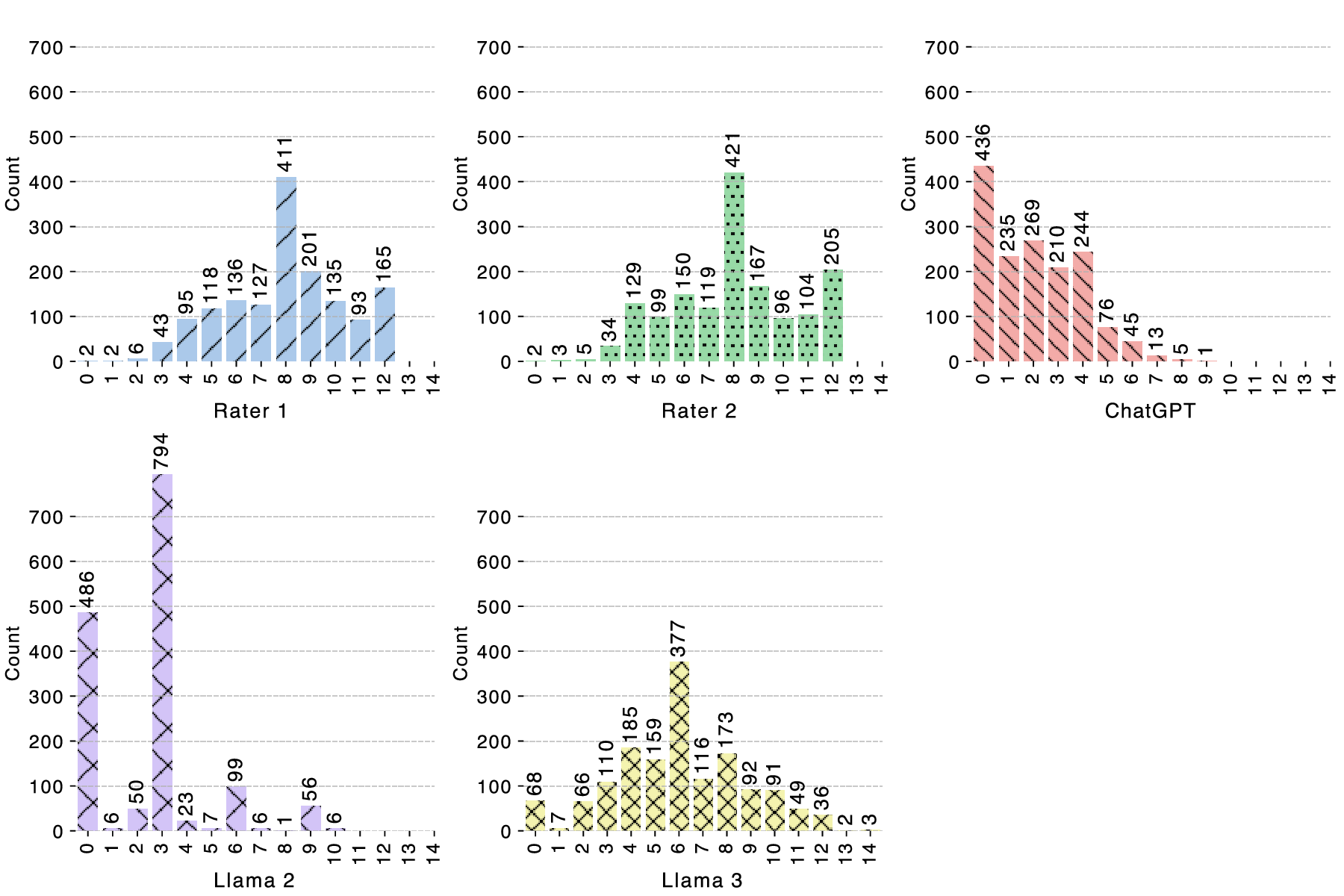}
    \caption{Score distribution of human raters and LLMs including \lamatri for \Tseven}
    \label{llama3-score-dist-p7}
\end{figure}

 As shown in Table \ref{llama3-desc-stat-p1}, for \Tone \lamatri assigned valid overall scores to all $1783$ essays, with scores consistently ranging from 1 to 6. In \Tseven, Table \ref{llama3-desc-stat-p7} reveals that \lamatri surprisingly assigned higher scores ($M=6.05, SD=2.73$) than \CGPT-3.5T or \lama-2, and these scores are much closer to those of the human raters ($M=8.02, SD=2.42$ and $M=8.04, SD=2.52$). \lamatri demonstrated impressive results by scoring only 4 samples out of range, all of which occurred while predicting the organization trait score. We can see a more normalized distribution of scores in Figures \ref{llama3-score-dist-p1} and \ref{llama3-score-dist-p7}.

\begin{table}[h]
    \centering
    \captionsetup{width=0.75\textwidth}
    \caption{Descriptive statistics with \lamatri and other scoring methods for \Tone }
    \begin{tabular}{lccccc}
        \textbf{} & \textbf{Rater 1 } & \textbf{Rater 2 } & \textbf{\CGPT-3.5T } & \textbf{\lama-2 } & \textbf{\lamatri}\\ \hline
        \textbf{Mean} & 4.26 & 4.27 & 1.90 & 3.62 & 2.55\\ 
        \textbf{SD} & 0.84 & 0.82 & 0.56 & 0.66 & 1.00\\ 
        \textbf{Min} & 1.00 & 1.00 & 1.00 & 1.00 & 1.00 \\ 
        \textbf{Median} & 4.00 & 4.00 & 2.00 & 4.00 & 2.00\\ 
        \textbf{Max} & 6.00 & 6.00 & 5.00 & 4.00 & 5.00\\ 
    \end{tabular}
    \label{llama3-desc-stat-p1}
\end{table}

\begin{table}[h]
    \centering
    \captionsetup{width=0.75\textwidth}
    \caption{Descriptive statistics with \lamatri and other scoring methods for  \Tseven}
    \begin{tabular}{lccccc}
        \textbf{} & \textbf{Rater 1} & \textbf{Rater 2} & \textbf{\CGPT-3.5T} & \textbf{\lama-2} & \textbf{\lamatri}\\ \hline
        \textbf{Mean} & 8.02 & 8.04 & 2.07 & 2.66 & \textbf{6.05}\\ 
        \textbf{SD} & 2.42 & 2.52 & 1.83 & 2.68 & 2.73\\ 
        \textbf{Min} & 0.00 & 0.00 & 0.00 & 0.00 & 0.00 \\ 
        \textbf{Median} & 8.00 & 8.00 & 2.00 & 3.00 & 6.00\\ 
        \textbf{Max} & 12.00 & 12.00 & 9.00 & 26.00 & 14.00\\ 
    \end{tabular}
    \label{llama3-desc-stat-p7}
\end{table}

\begin{table}[h]
    \centering
    \captionsetup{width=0.75\textwidth}
    \caption{Moderate correlation between human rater and \lamatri scores in \Tone}
    \begin{tabular}{lcccc}
        \textbf{Score} & \textbf{Rater 1 } & \textbf{Rater 2 } & \textbf{\CGPT-3.5T }  & \textbf{\lama-2 } \\ \hline
        \textbf{\CGPT-3.5T } & 0.23 & 0.21 & 1.00  & 0.32\\ 
        \textbf{\lama-2 } & 0.59 & 0.58 & 0.32 & 1.00 \\
        \textbf{\lamatri } & 0.52 & 0.49 & 0.46  & 0.54\\\hline
    \end{tabular}
    \label{llama3-human-llm-corr-p1}
\end{table} 
In terms of agreement with human raters in \Tone, as shown in Table \ref{llama3-human-llm-corr-p1}, \lama-2 has a correlation that is $13.5\%$ higher with rater 1 and $18.4\%$ higher with rater 2 compared to \lamatri. On average, \lamatri is $16\%$ less correlated with human scores than \lama-2. However, \lamatri shows significantly better agreement with human raters compared to \CGPT-3.5T, with a $126\%$ higher correlation with rater 1 and a $133.3\%$ higher correlation with rater 2, resulting in an average improvement of $130\%$ in correlation compared to \CGPT-3.5T.
\begin{table}[h]
    \centering
    \captionsetup{width=0.75\textwidth}
    \caption{Moderate correlation between human and \lamatri scores in \Tseven}
    \begin{tabular}{llcccc}
        \textbf{Traits} & \textbf{LLM} & \textbf{Rater 1} & \textbf{Rater 2} & \textbf{\CGPT-3.5T} & \textbf{\lama-2}\\ \hline
         Ideas \& content 
            & \CGPT-3.5T & 0.07 & 0.10 & 1.00  & 0.21\\
            & \lama-2 & 0.27 & 0.29 & 0.21  & 1.00\\
            & \lamatri & 0.42 & 0.44 & 0.27 & 0.44\\ \hline
        Organization 
                & \CGPT-3.5T & 0.10 & 0.10 & 1.00 & 0.24  \\
                & \lama-2 & 0.32 & 0.36 & 0.24 & 1.00 \\
                & \lamatri & 0.38 & 0.39 & 0.26 & 0.41\\ \hline
        Style 
            & \CGPT-3.5T & 0.12 & 0.15 & 1.00  & 0.11 \\
            & \lama-2 & 0.27 & 0.28 & 0.11 & 1.00 \\ 
            & \lamatri & 0.39 & 0.43 & 0.22 & 0.36 \\\hline
        Convention 
            & \CGPT-3.5T & 0.19 & 0.18 & 1.00  & 0.20\\
            & \lama-2 & 0.34 & 0.37 & 0.20 & 1.00 \\ 
            & \lamatri & 0.31 & 0.36 & 0.20 & 0.45\\ \hline
        Overall 
            & \CGPT-3.5T  & 0.18 & 0.19 & 1.00 & 0.30 \\ 
            & \lama-2  & 0.38 & 0.40 & 0.30  & 1.00\\
            & \lamatri & 0.49 & 0.52 & 0.40 & 0.52 \\ \hline
    \end{tabular}
    \label{llama3-human-llm-corr-p7}
\end{table}

On the other hand, in \Tseven, as shown in Table \ref{llama3-human-llm-corr-p7}, \lamatri demonstrates a correlation that is $29\%$ higher with rater 1 and $30\%$ higher with rater 2 compared to \lama-2. On average, \lamatri shows an $29.5\%$ better correlation with human scores than \lama-2. Additionally, \lamatri significantly outperforms \CGPT-3.5T, with a $172\%$ higher correlation with rater 1 and a $173.7\%$ higher correlation with rater 2, resulting in an overall average improvement of $173\%$ in correlation compared to \CGPT-3.5T. Table \ref{llama3-asap-plus-corr-grader} shows that similar to \lama-2, \lamatri has strong correlation with ASAP++ trait scores by human raters. Notably, correlations for \lamatri are higher than both \lama-2 and \CGPT-3.5T across all traits.

\begin{table}[h]
    \centering
    \captionsetup{width=0.75\textwidth}
    \caption{Strong correlation between ASAP++ traits and \lamatri scores in \Tone }
    \begin{tabular}{lccc}
        \textbf{ Traits } & 
        \rotB{\textbf{\CGPT-3.5T}} & 
        \rotB{\textbf{\lama-2}} &
        \rotB{\textbf{\lamatri}}
        \\ \\ \hline
        \textbf{Ideas \& Content} & 0.34 & 0.61 & 0.64  \\ 
        \textbf{Organization}  & 0.36 & 0.60  & 0.60  \\ 
        \textbf{Word Choice}  & 0.33 & 0.59  & 0.63  \\ 
        \textbf{Sentence Fluency}  & 0.36 & 0.60 & 0.64 \\ 
        \textbf{Conventions}  & 0.35 & 0.59 & 0.63  \\ \hline
    \end{tabular}
    \label{llama3-asap-plus-corr-grader}
\end{table}


\subsection{RQ2: What are the possible reasons behind the similarity/ difference in scores?}
We perform similar correlation analyses with essay features, readability indices, and language-checking tools. 
Table \ref{llama3-essay-features-p1p7} reveals that \lamatri has a moderate correlation with essay features, which is weaker than \lama-2 but stronger than \CGPT-3.5T. Comparatively, \lamatri shows stronger correlations with some readability indices, as seen in Table \ref{llama3-grades-readability-corr-p1p7}, than the other LLMs. Similar to \CGPT-3.5T, \lamatri exhibits negative yet weak correlations across all mistake types in \Tone, as shown in Table \ref{llama3-neg-corr-llm-mistake-p1p7}. For \Tseven, \lamatri shows negative correlations only with misspelling, grammar, and capitalization mistake counts, similar to \lama-2.

\begin{table}[!h]
    \centering
    \captionsetup{width=0.75\textwidth}
    \caption{Moderate correlation between \lamatri scores and basic essay features}
    \begin{tabular}{llcccccc}
        \textbf{} & 
        \textbf{Score} & 
        \rotB{\textbf{Essay Length (sentences)}} & 
        \rotB{\textbf{Essay Length (tokens)}} & 
        \rotB{\textbf{FANBOYS (total)}} & 
        \rotB{\textbf{FANBOYS (unique)}} & 
        \rotB{\textbf{Transition Phrases \\(total)}} & \rotB{\textbf{Transition Phrases (unique)}} \\ \\
        \hline
        \textbf{} & \textbf{\CGPT-3.5T} & 0.16 & 0.20 & 0.12 & 0.09 & 0.05 & 0.08 \\ 
        \textbf{\Tone} & \textbf{\lama-2} & 0.59 & 0.66 & 0.45 & 0.35 & 0.32 & 0.36 \\ 
        \textbf{} & \textbf{\lamatri} & 0.44 & 0.53 & 0.31 & 0.22 & 0.20 & 0.25 \\
         \hline
        \textbf{} & \textbf{\CGPT-3.5T} & -0.02 & 0.00 & -0.01 & 0.08 & 0.03 & 0.09  \\ 
        \textbf{\Tseven} & \textbf{\lama-2} & 0.42 & 0.45 & 0.27 & 0.21 & 0.24 & 0.26 \\ 
        \textbf{} & \textbf{\lamatri} & 0.37 & 0.44 & 0.30 & 0.28 & 0.27 & 0.34 \\ \hline
    \label{llama3-essay-features-p1p7}
    \end{tabular}
\end{table}

\begin{table}[h]
    \centering
    \captionsetup{width=0.75\textwidth}
    \caption{Comparatively stronger correlations between \lamatri scores and various readability indices}
    \begin{tabular}{llcccccccc}
        \textbf{} & \textbf{Score} &
        \rotB{\textbf{Flesch-Kincaid \\ Grade-level}} & 
        \rotB{\textbf{Flesch \\ Reading Ease}} & 
        \rotB{\textbf{Smog Index}} & 
        \rotB{\textbf{Coleman Liau}} & 
        \rotB{\textbf{Gunning Fog}} & 
        \rotB{\textbf{Automated Readability}} & 
        \rotB{\textbf{Linsear Write}} & 
        \rotB{\textbf{Dale Chall}} \\ \\ \hline
        
        \textbf{} &\textbf{\CGPT-3.5T} & 0.10 & -0.23 & 0.18 & 0.17 & -0.09 & 0.00 & -0.08 & -0.01 \\ 
        \textbf{\Tone} &\textbf{\lama-2} & -0.01 & -0.19 & 0.13 & 0.24 & -0.25 & -0.08 & -0.19 & 0.05 \\ 
        \textbf{} & \textbf{\lamatri} & \textbf{0.15} & \textbf{-0.39} & \textbf{0.29} & \textbf{0.37} & -0.17 & 0.04 & -0.11 & 0.10 \\ \hline
        \textbf{} &\textbf{\CGPT-3.5T} & 0.09 & -0.20 & 0.18 & 0.11 & -0.04 & 0.01 & -0.04 & -0.21 \\ 
       \textbf{\Tseven} & \textbf{\lama-2} & -0.02 & -0.09 & 0.21 & 0.18 & -0.15 & -0.03 & -0.12 & -0.05 \\
       \textbf{} & \textbf{\lamatri} & 0.05 & -0.16 & \textbf{0.26} & \textbf{0.19} & -0.12 & -0.00 & -0.08 & -0.19 \\ \hline
    \end{tabular}
    \label{llama3-grades-readability-corr-p1p7}
\end{table}

\begin{table}[h]
    \centering
    \captionsetup{width=0.75\textwidth}
    \caption{Negative weak correlation between \lamatri scores and mistake counts}
    \begin{tabular}{llcccccc}
    
    \textbf{} &
        \diagbox[width=3.8cm, height=1.8cm]{Score}{ Mistake count }&
        \rotB{\textbf{Aspell \\Mispelling}} &
        \rotB{\textbf{\ltool Grammar}} & 
        \rotB{\textbf{\ltool Spelling}} & 
        \rotB{\textbf{\ltool Style}} & 
        \rotB{\textbf{\ltool Punctuation}} & 
        \rotB{\textbf{\ltool Capitalization}} \\ \\ \hline
        \textbf{} &\textbf{\rone} & 0.14 & 0.09 & 0.11 & 0.09 & 0.02 & 0.11 \\ 
        \textbf{\Tone } &\textbf{\rtwo} & 0.16 & 0.14 & 0.13 & 0.10 & 0.03 & 0.12 \\ 
        \textbf{} &\textbf{\CGPT-3.5T} & \textbf{-0.12} & \textbf{-0.11} & \textbf{-0.15} & \textbf{-0.02} & \textbf{-0.01} & \textbf{-0.08} \\ 
        \textbf{} &\textbf{\lama-2} & 0.04 & 0.05 & 0.02 & 0.05 & 0.05 & 0.11 \\
        \textbf{} &\textbf{\lamatri} & \textbf{-0.14} & \textbf{-0.15} & \textbf{-0.18} & \textbf{-0.01} & -\textbf{0.10} & \textbf{-0.02 }\\ \hline
        
        \textbf{} &\textbf{\rone} & 0.11 & 0.05 & 0.20 & 0.11 & 0.11 & 0.05 \\ 
        \textbf{\Tseven} &\textbf{\rtwo} & 0.10 & 0.06 & 0.21 & 0.09 & 0.10 & 0.05 \\ 
         \textbf{} & \textbf{\CGPT-3.5T} & \textbf{-0.24} & \textbf{-0.17} & \textbf{-0.21} & \textbf{-0.05} & \textbf{-0.05 }& \textbf{-0.17} \\ 
         \textbf{} &\textbf{\lama-2} & \textbf{-0.10} & \textbf{-0.03} & 0.08 & 0.06 & 0.01 & \textbf{-0.11} \\
         \textbf{} &\textbf{\lamatri} & \textbf{-0.13} & \textbf{-0.05} & 0.01 & 0.01 & 0.06 & \textbf{-0.16} \\
    \hline
    \end{tabular}
    \label{llama3-neg-corr-llm-mistake-p1p7}
\end{table}

    
        


\vspace*{2em}


\subsection{RQ3: Do LLMs offer explanations in a tone that reflects their scores?}

As seen in Tables \ref{llama3-chatgpt-feedback-corr-p1p7} and \ref{llama3-lama-feedback-corr-p1p7}, \lamatri has a weaker correlation with \CGPT-3.5T and \lama-2 explanations than these models have with themselves. However, Table \ref{lama3-feedback-corr-p1p7} reveals that \lamatri's explanations are comparatively strongly correlated with the human raters' scores in \Tone. Furthermore, \lamatri's provided scores are moderately positively correlated with its own explanations.

\begin{table}[h]
    \centering
    \caption{\lamatri scores weakly correlate with \CGPT-3.5T's explanation}
    \begin{tabular}{llcccc}
    \textbf{} & \raisebox{-1.25em}{\textbf{Score}} &  
    \rotB{\textbf{Explanation Length \\ (sentences)}} & 
    \rotB{\textbf{Explanation Length \\(tokens)}} & 
    \rotB{\textbf{Explanation Sentiment (average)}} & 
    \rotB{\textbf{Explanation Sentiment (max)}} \\ \\\hline
    
        \textbf{} & \textbf{\rone} & 0.12 & 0.17 & 0.06 & 0.04 \\ 
        \textbf{\Tone } & \textbf{\rtwo} & 0.11 & 0.15 & 0.06 & 0.04  \\ 
        \textbf{} & \textbf{\CGPT-3.5T} & 0.33 & 0.45 & \textbf{0.33} & \textbf{0.28}  \\ 
        \textbf{} & \textbf{\lama-2} & 0.14 & 0.19 & 0.14 & 0.12  \\
        \textbf{} & \textbf{\lamatri} & 0.16 & 0.25 & 0.18 & 0.13  \\ \hline
    
        \textbf{} & \textbf{\rone} & 0.05 & 0.04 & 0.05 & 0.02  \\ 
        \textbf{\Tseven} & \textbf{\rtwo} & 0.06 & 0.04 & 0.05 & 0.01  \\ 
        \textbf{} & \textbf{\CGPT-3.5T} & 0.28 & 0.30 & \textbf{0.24} & \textbf{0.13}  \\ 
        \textbf{} & \textbf{\lama-2} & 0.13 & 0.13 & 0.08 & 0.02  \\ 
        \textbf{} & \textbf{\lamatri} & 0.13 & 0.15 & 0.10 & 0.02  \\ \hline
    \end{tabular}
    \label{llama3-chatgpt-feedback-corr-p1p7}
\end{table}

\begin{table}[h]
    \centering
    \caption{ \lamatri scores weakly correlate with \lama-2's explanation}
    \begin{tabular}{llcccc}
    \textbf{} & \raisebox{-1.25em}{\textbf{Score}} &  
    \rotB{\textbf{Explanation Length \\ (sentences)}} & 
    \rotB{\textbf{Explanation Length \\(tokens)}} & 
    \rotB{\textbf{Explanation Sentiment (average)}} & 
    \rotB{\textbf{Explanation Sentiment (max)}} \\ \\\hline
    
        \textbf{} & \textbf{\rone} & 0.26 & 0.26 & 0.30 & 0.24 \\ 
        \textbf{\Tone } & \textbf{\rtwo} & 0.25 & 0.27 & 0.29 & 0.22 \\ 
        \textbf{} & \textbf{\CGPT-3.5T} & 0.10 & 0.11 & 0.24 & 0.20 \\ 
        \textbf{} & \textbf{\lama-2} & 0.28 & 0.31 & \textbf{0.54} & \textbf{0.42} \\ 
        \textbf{} & \textbf{\lamatri} & 0.22 & 0.23 & 0.34 & 0.26 \\ 
        \hline
        \textbf{} & \textbf{\rone}  & 0.16 & 0.21 & 0.26 & 0.22 \\ 
        \textbf{\Tseven} & \textbf{\rtwo}  & 0.13 & 0.17 & 0.25 & 0.22 \\ 
        \textbf{} & \textbf{\CGPT-3.5T}  & 0.11 & 0.13 & 0.13 & 0.14 \\ 
        \textbf{} & \textbf{\lama-2}  & 0.27 & 0.30 & \textbf{0.50} & \textbf{0.44} \\
        \textbf{} & \textbf{\lamatri} & 0.23 & 0.27 & 0.31 & 0.30 \\ 
    \end{tabular}
    \label{llama3-lama-feedback-corr-p1p7}
\end{table}

\begin{table}[t]
    \centering
    \caption{\lamatri's moderately positive explanation}
    \begin{tabular}{llcccc}
    \textbf{} & \raisebox{-1.25em}{\textbf{Score}} &  
    \rotB{\textbf{Explanation Length \\ (sentences)}} & 
    \rotB{\textbf{Explanation Length \\(tokens)}} & 
    \rotB{\textbf{Explanation Sentiment (average)}} & 
    \rotB{\textbf{Explanation Sentiment (max)}} \\ \\\hline
    
        \textbf{} & \textbf{\rone} & \textbf{0.35} & 0.40 & 0.23 & 0.21 \\ 
        \textbf{\Tone } & \textbf{\rtwo} & \textbf{0.37} & 0.41 & 0.21 & 0.21 \\ 
        \textbf{} & \textbf{\CGPT-3.5T} & 0.11 & 0.21 & 0.29 & 0.23 \\ 
        \textbf{} & \textbf{\lama-2} & 0.32 & 0.43 & 0.31 & 0.25 \\ 
        \textbf{} & \textbf{\lamatri} & 0.34 & \textbf{0.47} & \textbf{0.49} & \textbf{0.40} \\ 
        \hline
        \textbf{} & \textbf{\rone}  & 0.25 & 0.34 & 0.28 & 0.20 \\ 
        \textbf{\Tseven} & \textbf{\rtwo} & 0.29 & 0.36 & 0.30 & 0.22 \\ 
        \textbf{} & \textbf{\CGPT-3.5T} & 0.15 & 0.24 & 0.24 & 0.17 \\ 
        \textbf{} & \textbf{\lama-2}  & 0.22 & 0.28 & 0.38 & 0.26 \\
        \textbf{} & \textbf{\lamatri} & \textbf{0.34} & \textbf{0.46} & \textbf{0.48} & \textbf{0.32} \\ 
    \end{tabular}
    \label{lama3-feedback-corr-p1p7}
\end{table}


\subsection{RQ4: Can LLMs correctly identify and assess spelling and grammatical mistakes and score accordingly?} \label{llama3-section-RQ4}

Similar to previous analyses, we categorized \lamatri responses based on the mention of grammatical and spelling errors into three distinct groups. As shown in Tables \ref{llama3-misspellcategory-p1}, \ref{llama3-misspellcategory-p7}, \ref{llama3-grammaticalcategory-p1}, and \ref{llama3-grammaticalcategory-p7}, we observe a logical increase in the average number of mistake counts across grammatical and spelling categories for both \Tone and \Tseven. The differences in spelling and grammar mistake counts are prominent in the first and third groups. All results are statistically significant, with \lamatri showing comparatively more medium and large (in bold) effect sizes. There is a noticeable decrease in scores from the first group to the third group as mistakes increase. The sentiment of the explanations provided by \lamatri is moderate, falling between \CGPT-3.5T's harsher tone and \lama-2's more positive tone.

\begin{table}[t]
    \captionsetup{width=0.75\textwidth}
    \caption{Changes in average number of misspellings, \lamatri's scores and explanation sentiment across different misspelling categories in \Tone}
    \begin{tabular}{llccccc}
        \multicolumn{1}{p{2.3cm}}{\textbf{Misspelling Category}} & 
        \rotB{\textbf{Sample \\Count}} & 
        \rotB{\textbf{Aspell \\ Misspelling}} & 
        \rotB{\textbf{\ltool Spelling}} & 
        \rotB{\textbf{LLM Score}} &
        \rotB{\textbf{Explanation \\ Sentiment (average)}} &
        \rotB{\textbf{Explanation \\ Sentiment (max)}}\\ \\\hline\hline
        \multicolumn{7}{c}{\lamatri}\\
        \hline \hline
        (1) No mention & 1504 & 8.01 & 9.15 & 2.59 & 0.30 & 0.31 \\ 
        (2) Unqualified mention & 124 & 13.48 & 14.56 & 2.58 & 0.30 & 0.28\\ 
        (3) Qualified mention & 155 & 12.97 & 14.23 & 2.14 & 0.15 & 0.15 \\ \hline
        F-stat $(2,1780)$ & & 73.62 & 64.69 & 14.13 & 2.77 & 3.88 \\ 
        P-value & & $p<0.05$ & $p<0.05$ & $p<0.05$ & $p>0.05$ & $p<0.05$\\ 
        Effect size ($\omega^2$) & & 0.08 & 0.07 & 0.02 & 0.002 & 0.003 \\ 
    \end{tabular}
    \label{llama3-misspellcategory-p1}
\end{table}

\begin{table}[h!]
    \centering
    \captionsetup{width=1.0\textwidth}
    \caption{Changes in average number of misspellings, \lamatri's scores and explanation sentiment across different misspelling categories in \Tseven}
    \begin{tabular}{llccccc}
        \multicolumn{1}{p{2.2cm}}{\textbf{Misspelling Category}} & 
        \rotB{\textbf{Sample \\Count}} & 
        \rotB{\textbf{Aspell \\ Misspelling}} & 
        \rotB{\textbf{\ltool Spelling}} & 
        \rotB{\textbf{LLM Score}} &
        \rotB{\textbf{Explanation \\ Sentiment (average)}} &
        \rotB{\textbf{Explanation \\ Sentiment (max)}}\\ \\
        


        \hline\hline
        \multicolumn{7}{c}{\lamatri}\\
        \hline\hline
        (1) No mention & 659 & 2.40 & 4.66 & 6.80 & -0.29 & 0.05 \\ 
        (2) Unqualified mention & 396 & 4.03 & 6.30 & 6.49 & -0.33 & 0.14 \\ 
        (3) Qualified mention & 514 & 6.27 & 8.11 & 4.75 & -0.60 & -0.16 \\  \hline
        F-stat $(2,1566)$ & & 138.60 & 63.91 & 99.32 & 45.1 & 25.98 \\ 
        P-value & & $p<0.05$ & $p<0.05$ & $p<0.05$ & $p<0.05$ & $p<0.05$\\ 
        Effect size ($\omega^2$) & & \textbf{0.15} & 0.07 & \textbf{0.11} & 0.05 & 0.03 \\ 
    \end{tabular}
    \label{llama3-misspellcategory-p7}
\end{table}

\begin{table}[h]
    \centering
    \captionsetup{width=0.75\textwidth}
    \caption{Changes in average number of grammar mistakes, \lamatri's scores and explanation sentiment across different grammatical categories in \Tone }
    \begin{tabular}{lccccc}
        \multicolumn{1}{p{2.3cm}}{\textbf{ Grammar \mbox{Error Category}}} & 
        \rotB{\textbf{Sample \\Count}} & 
        \rotB{\textbf{\ltool Grammar}} & 
        \rotB{\textbf{LLM Score }}& 
        \rotB{\textbf{Explanation \\ Sentiment (average)}} &
        \rotB{\textbf{Explanation \\ Sentiment (max)}}\\ \\
        
        
        
        \hline\hline
        \multicolumn{6}{c}{\lamatri}\\
        \hline\hline
        (1) No mention &  751 & 2.65 & 2.92 & 0.48 & 0.48 \\ 
        (2) Unqualified mention & 608 & 3.30 & 2.38 & 0.19 & 0.17 \\ 
        (3) Qualified mention & 424 & 3.49 & 2.14 & 0.08 & 0.14  \\  \hline
        F-stat $(2,1780)$ & & 19.99 & 105.21 & 43.75 &47.87  \\ 

        P-value & & $p<0.05$ & $p<0.05$ & $p<0.05$ & $p<0.05$\\ 
        Effect size ($\omega^2$) & & 0.02 & \textbf{0.11} & 0.05 & 0.05 \\ 
    \end{tabular}
    \label{llama3-grammaticalcategory-p1}
\end{table}

\begin{table}[h]
    \centering
    \captionsetup{width=1.0\textwidth}
    \caption{Changes in average number of grammar mistakes, \lamatri's scores and explanation sentiment across different grammatical categories in \Tseven}
    \begin{tabular}{lccccc}
        \multicolumn{1}{p{2.5cm}}{\textbf{Grammar \mbox{Error Category}}} & 
        \rotB{\textbf{Sample \\Count}} & 
        \rotB{\textbf{\ltool Grammar}} & 
        \rotB{\textbf{LLM Score }}& 
        \rotB{\textbf{Explanation \\ Sentiment (average)}} &
        \rotB{\textbf{Explanation \\ Sentiment (max)}}\\ \\
        \hline\hline
        \multicolumn{6}{c}{\lamatri}\\
        \hline\hline
        (1) No mention & 295 & 1.00 & 6.71 & -0.30 & 0.08 \\ 
        (2) Unqualified mention &  627 & 1.01 & 6.89 & -0.25 & 0.16 \\ 
        (3) Qualified mention &  647& 1.71 & 4.94 & -0.59 & -0.18  \\ \hline
        F-stat $(2,1780)$ & & 41.73 & 103.67 & 55.95 & 43.35  \\ 
        P-value & & $p<0.05$ & $p<0.05$ & $p<0.05$ & $p<0.05$ \\ 
        Effect size ($\omega^2$) & & 0.05 & \textbf{0.12} & 0.07 & 0.05 \\ 
    \end{tabular}
    \label{llama3-grammaticalcategory-p7}
\end{table}

\begin{table}[ht!]
    \centering
    \captionsetup{width=0.8\textwidth}
    \caption{Comparison of average grades assigned by LLMs across human rater grade categories \Tone }
    \begin{tabular}{lcccc}
    \multicolumn{1}{p{1.3cm}}{\textbf{Human Score Class}} & 
        \rotB{\textbf{Sample \\Count}} &
        \rotB{\textbf{\CGPT-3.5T Score}} & 
        \rotB{\textbf{\lama-2 \\ Score }} &
        \rotB{\textbf{\lamatri \\ Score }}\\  \\ \hline
        Score 1-2 & 28 & 1.21 & 1.68 & 1.11 \\ 
        Score 2-4 & 949 & 1.82 & 3.39 & 2.17\\ 
        Score 4-6 & 806 & 2.01 & 3.95 &3.05\\ 
        \hline
        F-stat(2,1780) & & 46.46 & 399.13 &254.96\\
        P-values & & $p<0.05$ & $p<0.05$ & $p<0.05$ \\
        Effect size ($\omega^2$) & & 0.05 & \textbf{0.31} & \textbf{0.22}\\
    \end{tabular}
    \label{llama3-humancategory-p1}
\end{table}

\begin{table}[ht!]
    \centering
    \captionsetup{width=0.75\textwidth}
    \caption{Comparison of average grades assigned by LLMs across human rater grade categories \Tseven}
    \begin{tabular}{lcccc}
    \multicolumn{1}{p{1.3cm}}{\textbf{Human Score Class}} & 
        \rotB{\textbf{Sample \\Count}} &
        \rotB{\textbf{\CGPT-3.5T Score}} & 
        \rotB{\textbf{\lama-2 \\ Score }} &
        \rotB{\textbf{\lamatri \\ Score }}\\  \\ \hline
        Score 0-4 & 108 & 0.84 & 0.89 &  2.61\\ 
        Score 4-8 & 717 & 1.93 & 1.93 & 5.29\\ 
        Score 8-15 & 744 & 2.38 & 3.62 & 7.29  \\ 
        \hline
        F-stat(2,1566) & & 39.05 & 111.2 & 249.93\\
        P-values & & $p<0.05$ & $p<0.05$ & $p<0.05$ \\
        Effect size ($\omega^2$) & & 0.05 & \textbf{0.12} & \textbf{0.24}\\
    \end{tabular}
    \label{llama3-humancategory-p7}
\end{table}

In Tables \ref{llama3-humancategory-p1} and \ref{llama3-humancategory-p7}, we categorized essay samples into three distinct groups based on the scores provided by human raters and calculated the average scores given by LLMs. We observe a more distinct difference in \lamatri scores from the first group to the second group to the third group. \lamatri successfully assigns lower scores to lower-grade groups and higher scores to higher-grade groups. In both tasks, the results are statistically and practically significant.

\subsection{Prompt Engineering Results with \lamatri}

\begin{figure}[!h]%
    \centering
    \captionsetup{justification=centering}
    \includegraphics[width=0.75\textwidth]{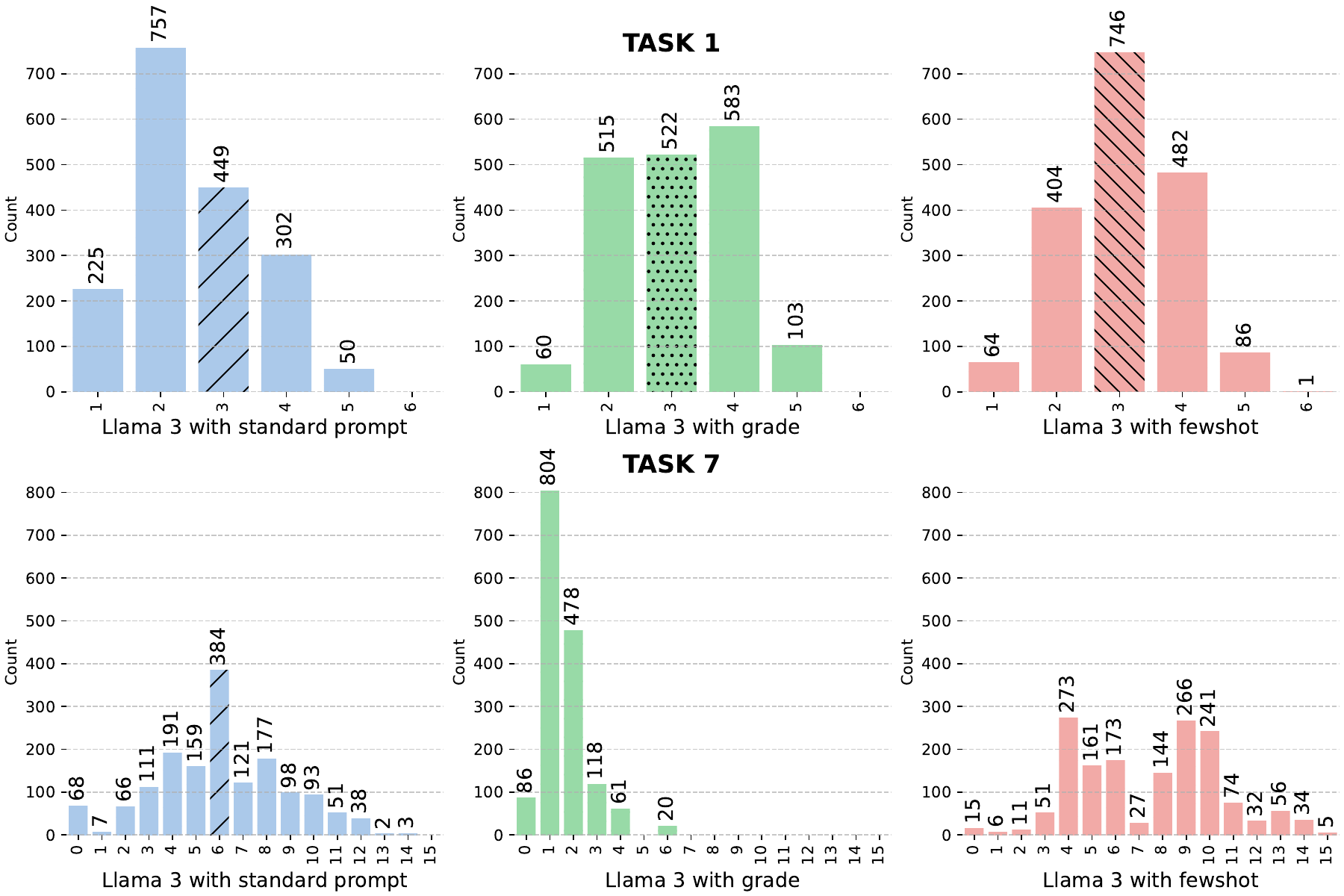}
    \caption{Score distribution across different prompts on \lamatri}
    \label{llama3-score-dist-p1p7-prompting}
\end{figure}

As shown in Table \ref{llama3-prompt-eng}, after incorporating the students' grade level, we observed an average increase of around $5\%$ in the correlation between human raters and \lamatri in \Tone. However, in \Tseven, adding the grade level resulted in a significant drop in the correlation between human raters and \lamatri, averaging $216\%$. This change is evident in the score distribution chart in Figure \ref{llama3-score-dist-p1p7-prompting}, where the normal distribution, initially in blue, moves to a lower grade range in green bars after the grade level is added. The diagram also shows the score distribution for different prompts. Additionally, as seen in Table \ref{llama3-prompt-eng}, after including two-shot examples, we observed an average increase of around $22\%$ in the correlation between human raters and \lamatri in \Tone, and an average increase of around $20\%$ in \Tseven.


\begin{table}[!t]
    \centering
    \captionsetup{width=0.75\textwidth}
    \caption{Changes in correlation scores after adding grade level of students and few-shot examples to the prompt of \lamatri (Increase denoted by and $\uparrow$ and decrease denoted by $\downarrow$)}
    \begin{tabular}{|c|c|cc|cc|}
    \hline
        \textbf{} & \textbf{} & \textbf{\Tone  } &  & \textbf{\Tseven} &  \\ \hline
        \textbf{Score} & ~ & Before & After & Before & After \\  \hline  
        
        \textbf{} & Rater 1 & 0.52 & {0.54} $\uparrow$ & 0.49 & {0.15} $\downarrow$ \\ 
        \textbf{\lamatri} & Rater 2 & 0.49 & {0.52} $\uparrow$ & 0.52 & {0.17} $\downarrow$ \\ 
        (with grade level) & \CGPT-3.5T & 0.46 & {0.44} $\downarrow$ & 0.40 & {0.14} $\downarrow$ \\ 
        \textbf{} & \lama-2 & 0.54 & {0.58} $\uparrow$ & 0.52 & {0.13} $\downarrow$ \\  
        
        \hline \hline

        \textbf{} & Rater 1 & 0.52 & {0.62} $\uparrow$ & 0.49 & {0.59} $\uparrow$ \\ 
        \textbf{\lamatri } & Rater 2 & 0.49 & {0.61} $\uparrow$ & 0.52 & {0.62} $\uparrow$ \\ 
        (two-shot) & \CGPT-3.5T & 0.46 & {0.41} $\downarrow$ & 0.40 & {0.43} $\uparrow$ \\ 
        \textbf{} & \lama-2 & 0.54 & {0.66} $\uparrow$ & 0.52 & {0.60} $\uparrow$ \\  
    \hline
    \end{tabular}
    \label{llama3-prompt-eng}
\end{table}

\section{Statistical Analysis Results} \label{appendix-section-statistics}
\begin{table}[!ht]
    \centering
    \captionsetup{width=0.75\textwidth}
    \caption{P value, Confidence Interval and Effect sizes of Table \ref{human-llm-corr-p1} and \ref{llama3-human-llm-corr-p1}}
\begin{tabular}{lccccc}
        \textbf{Comparison} & \rotB{\textbf{Pearson's r}} & \rotB{\textbf{p-value}} & \rotB{\textbf{95\% CI}} & 
        \rotB{\textbf{Effect Size (Cohen's d)}} & \rotB{\textbf{Cohen's d Interpretation}} \\ \\ \hline
        
        \rone vs. \rtwo & 0.72 & $<0.05$ & [0.7, 0.74] & -0.01 & Small effect size \\ 
        \rone vs. \CGPT & 0.23 & $<0.05$ & [0.19, 0.28] & 3.3 & Large effect size \\ 
        \rone vs. \lama & 0.59 & $<0.05$ & [0.56, 0.62] & 0.85 & Large effect size \\ 
        \rone vs. \lamatri & 0.52 & $<0.05$ & [0.49, 0.55] & 1.85 & Large effect size \\ 
        \rtwo vs. \CGPT & 0.21 & $<0.05$ & [0.16, 0.25] & 3.38 & Large effect size \\ 
        \rtwo vs. \lama & 0.58 & $<0.05$ & [0.55, 0.61] & 0.88 & Large effect size \\ 
        \rtwo vs. \lamatri & 0.49 & $<0.05$ & [0.45, 0.52] & 1.88 & Large effect size \\ 
        \CGPT vs. \lama & 0.32 & $<0.05$ & [0.27, 0.36] & -2.81 & Large effect size \\
        \CGPT vs. \lamatri & 0.46 & $<0.05$ & [0.42, 0.49] & -0.8 & Large effect size \\ 
        \lama vs. \lamatri & 0.54 & $<0.05$ & [0.51, 0.57] & 1.26 & Large effect size \\ \hline
    \end{tabular}
\end{table}

\begin{table}[!ht]
    \centering
    \captionsetup{width=0.75\textwidth}
    \caption{P value, Confidence Interval and Effect sizes of Table \ref{human-llm-corr-p7} and  \ref{llama3-human-llm-corr-p7} (\Tseven overall scores)}
    \begin{tabular}{lccccc}
         \textbf{Comparison} & \rotB{\textbf{Pearson's r}} & \rotB{\textbf{p-value}} & \rotB{\textbf{95\% CI}} & 
        \rotB{\textbf{Effect Size (Cohen's d)}} & \rotB{\textbf{Cohen's d Interpretation}} \\ \\ \hline
        
        \rone vs. \rtwo & 0.72 & $<0.05$ & [0.7, 0.74] & -0.01 & Small effect size \\ 
        \rone vs. \CGPT & 0.18 & $<0.05$ & [0.13, 0.23] & 2.77 & Large effect size \\ 
        \rone vs. \lama & 0.38 & $<0.05$ & [0.34, 0.43] & 2.1 & Large effect size \\ 
        \rone vs. \lamatri & 0.49 & $<0.05$ & [0.45, 0.53] & 0.76 & Large effect size \\ 
        \rtwo vs. \CGPT & 0.19 & $<0.05$ & [0.15, 0.24] & 2.71 & Large effect size \\ 
        \rtwo vs. \lama & 0.4 & $<0.05$ & [0.36, 0.45] & 2.07 & Large effect size \\ 
        \rtwo vs. \lamatri & 0.52 & $<0.05$ & [0.48, 0.55] & 0.76 & Large effect size \\ 
        \CGPT vs. \lama & 0.3 & $<0.05$ & [0.25, 0.34] & -0.26 & Medium effect size \\ 
        \CGPT vs. \lamatri & 0.4 & $<0.05$ & [0.35, 0.44] & -1.71 & Large effect size \\ 
        \lama vs. \lamatri & 0.52 & $<0.05$ & [0.48, 0.56] & -1.25 & Large effect size \\ \hline
    \end{tabular}
\end{table}

\begin{table}[!ht]
    \centering
    \caption{P value, Confidence Interval and Effect sizes of Table \ref{asap-plus-corr-grader} and \ref{llama3-asap-plus-corr-grader}}
    \begin{tabular}{lccccc}
        \textbf{Comparison} & \rotB{\textbf{Pearson's r}} & \rotB{\textbf{p-value}} & \rotB{\textbf{95\% CI}} & 
        \rotB{\textbf{Effect Size (Cohen's d)}} & \rotB{\textbf{Cohen's d Interpretation}} \\ \\ \hline
        Content vs. \rone & 0.66 & $<0.05$ & [0.63, 0.68] & -0.45 & Medium effect size \\ 
        Content vs. \rtwo & 0.67 & $<0.05$ & [0.64, 0.69] & -0.46 & Medium effect size \\ 
        Content vs. \CGPT & 0.34 & $<0.05$ & [0.3, 0.38] & 2.42 & Large effect size \\ 
        Content vs. \lama & 0.61 & $<0.05$ & [0.58, 0.64] & 0.27 & Medium effect size \\ 
        Content vs. \lamatri & 0.64 & $<0.05$ & [0.61, 0.66] & 1.3 & Large effect size \\ 
        Organization vs. \rone & 0.63 & $<0.05$ & [0.6, 0.66] & -0.58 & Large effect size \\
\hline
    \end{tabular}
\end{table}

\begin{table}[!ht]
    \centering
    \caption{P value, Confidence Interval and Effect sizes of Table \ref{essay-features-p1p7} and \ref{llama3-essay-features-p1p7} \Tone}
    \begin{tabular}{lccccc}
        \textbf{Comparison} & \rotB{\textbf{Pearson's r}} & \rotB{\textbf{p-value}} & \rotB{\textbf{95\% CI}} & 
        \rotB{\textbf{Effect Size (Cohen's d)}} & \rotB{\textbf{Cohen's d Interpretation}} \\ \\ \hline
        
        \rone vs. Essay Length (sentences) & 0.63 & $<0.05$ & [0.6, 0.66] & -2.97 & Large effect size \\ 
        \rone vs. Essay Length (tokens) & 0.74 & $<0.05$ & [0.72, 0.76] & -4.21 & Large effect size \\ 
        \rone vs. FANBOYS (total) & 0.5 & $<0.05$ & [0.47, 0.54] & -2.52 & Large effect size \\ 
        \rone vs. FANBOYS (unique) & 0.35 & $<0.05$ & [0.31, 0.39] & 0.28 & Medium effect size \\ 
        \rone vs. Transition Phrases (total) & 0.34 & $<0.05$ & [0.3, 0.38] & -1.27 & Large effect size \\ 
        \rone vs. Transition Phrases (unique) & 0.35 & $<0.05$ & [0.31, 0.39] & -0.11 & Small effect size \\
\hline
    \end{tabular}
\end{table}

\begin{table}[!ht]
    \centering
    \caption{P value, Confidence Interval and Effect sizes of Table \ref{essay-features-p1p7} and \ref{llama3-essay-features-p1p7} \Tseven (*no means $ p \ge 0.05$)}
    \begin{tabular}{lccccc}
        \textbf{Comparison} & \rotB{\textbf{Pearson's r}} & \rotB{\textbf{p-value}} & \rotB{\textbf{95\% CI}} & 
        \rotB{\textbf{Effect Size (Cohen's d)}} & \rotB{\textbf{Cohen's d Interpretation}} \\ \\ \hline
        
        \rone vs. essay sent & 0.61 & $<0.05$ & [0.57, 0.64] & -0.88 & Large effect size \\ 
        \rone vs. essay tok & 0.62 & $<0.05$ & [0.59, 0.65] & -2.59 & Large effect size \\ 
        \rone vs. fanboys t & 0.39 & $<0.05$ & [0.35, 0.43] & -0.31 & Medium effect size \\ 
        \rone vs. fanboys u & 0.31 & $<0.05$ & [0.27, 0.36] & 2.62 & Large effect size \\ 
        \rone vs. transition t & 0.36 & $<0.05$ & [0.32, 0.4] & 0.9 & Large effect size \\ 
        \rone vs. transition u & 0.43 & $<0.05$ & [0.38, 0.47] & 2.15 & Large effect size \\
\hline
    \end{tabular}
\end{table}

\begin{table}[!ht]
    \centering
    \caption{P value, Confidence Interval and Effect sizes of Table \ref{asap-plus-corr-essay} }
        \begin{tabular}{lccccc}
        \textbf{Comparison} & \rotB{\textbf{Pearson's r}} & \rotB{\textbf{p-value}} & \rotB{\textbf{95\% CI}} & 
        \rotB{\textbf{Effect Size (Cohen's d)}} & \rotB{\textbf{Cohen's d Interpretation}} \\ \\ \hline
        Content vs. Essay Length (sentences) & 0.58 & $<0.05$ & [0.55, 0.61] & -3.03 & Large   \\ 
        Content vs. Essay Length (tokens) & 0.68 & $<0.05$ & [0.65, 0.7] & -4.22 & Large \\ 
        Content vs. FANBOYS (total) & 0.43 & $<0.05$ & [0.4, 0.47] & -2.59 & Large  \\ 
        Content vs. FANBOYS (unique) & 0.28 & $<0.05$ & [0.24, 0.33] & -0.16 & Small  \\ 
        Content vs. Transition Phrases (total) & 0.3 & $<0.05$ & [0.26, 0.34] & -1.39 & Large  \\ 
        Content vs. Transition Phrases (unique) & 0.32 & $<0.05$ & [0.28, 0.36] & -0.38 & Medium  \\
\hline
    \end{tabular}
\end{table}

\begin{table}[!ht]
    \centering
    \caption{P value, Confidence Interval and Effect sizes of Table \ref{grades-readability-corr-p1p7} \Tone (*no means $ p \ge 0.05$)}
    \begin{tabular}{lccccc}
        \textbf{Comparison} & \rotB{\textbf{Pearson's r}} & \rotB{\textbf{p-value}} & \rotB{\textbf{95\% CI}} & 
        \rotB{\textbf{Effect Size (Cohen's d)}} & \rotB{\textbf{Cohen's d Interpretation}} \\ \\ \hline
        \rone vs. FleschGL & 0.02 & no & [-0.02, 0.07] & -1.49 & Large effect size \\ 
        \rone vs. FleschRE & -0.17 & $<0.05$ & [-0.21, -0.12] & -10.91 & Large effect size \\ 
        \rone vs. SmogInd & 0.16 & $<0.05$ & [0.12, 0.21] & -5.38 & Large effect size \\ 
        \rone vs. ColeLia & 0.3 & $<0.05$ & [0.26, 0.34] & -3.21 & Large effect size \\ 
        \rone vs. GunningFog & -0.19 & $<0.05$ & [-0.24, -0.15] & -1.66 & Large effect size \\ 
        \rone vs. AutoRea & 0.01 & no & [-0.03, 0.06] & -1.7 & Large effect size \\
\hline
    \end{tabular}
\end{table}

\begin{table}[!ht]
    \centering
    \caption{P value, Confidence Interval and Effect sizes of Table \ref{grades-readability-corr-p1p7} \Tseven (*no means $ p \ge 0.05$)}
    \begin{tabular}{lccccc}
        \textbf{Comparison} & \rotB{\textbf{Pearson's r}} & \rotB{\textbf{p-value}} & \rotB{\textbf{95\% CI}} & 
        \rotB{\textbf{Effect Size (Cohen's d)}} & \rotB{\textbf{Cohen's d Interpretation}} \\ \\ \hline
        \rone vs. FleschGL & -0.19 & $<0.05$ & [-0.24, -0.14] & 1.62 & Large effect size \\ 
        \rone vs. FleschRE & 0.13 & $<0.05$ & [0.08, 0.18] & -11.02 & Large effect size \\ 
        \rone vs. SmogInd & 0.16 & $<0.05$ & [0.11, 0.21] & 0.92 & Large effect size \\ 
        \rone vs. ColeLia & 0.11 & $<0.05$ & [0.06, 0.16] & 1.85 & Large effect size \\ 
        \rone vs. GunningFog & -0.3 & $<0.05$ & [-0.34, -0.25] & 0.57 & Large effect size \\ 
        \rone vs. AutoRea & -0.14 & $<0.05$ & [-0.18, -0.09] & 1.39 & Large effect size \\
\hline
    \end{tabular}
\end{table}

\begin{table}[!ht]
    \centering
    \caption{P value, Confidence Interval and Effect sizes of Table \ref{llama3-grades-readability-corr-p1p7} \Tone for \lamatri  (*no means $ p \ge 0.05$)}
    \begin{tabular}{lccccc}
        \textbf{Comparison} & \rotB{\textbf{Pearson's r}} & \rotB{\textbf{p-value}} & \rotB{\textbf{95\% CI}} & 
        \rotB{\textbf{Effect Size (Cohen's d)}} & \rotB{\textbf{Cohen's d Interpretation}} \\ \\ \hline
        \lamatri vs. FleschGL & 0.15 & $<0.05$ & [0.1, 0.2] & -2.45 & Large effect size \\ 
        \lamatri vs. FleschRE & -0.39 & $<0.05$ & [-0.43, -0.35] & -11.15 & Large effect size \\ 
        \lamatri vs. SmogInd & 0.29 & $<0.05$ & [0.25, 0.34] & -6.54 & Large effect size \\ 
        \lamatri vs. ColeLia & 0.37 & $<0.05$ & [0.33, 0.41] & -4.41 & Large effect size \\ 
        \lamatri vs. GunningFog & -0.17 & $<0.05$ & [-0.21, -0.12] & -2.62 & Large effect size \\ 
        \lamatri vs. AutoRea & 0.04 & no & [-0.01, 0.08] & -2.5 & Large effect size \\ 
        \lamatri vs. LinseaW & -0.11 & $<0.05$ & [-0.16, -0.07] & -2.61 & Large effect size \\ 
        \lamatri vs. DaleChall & 0.1 & $<0.05$ & [0.06, 0.15] & -5.31 & Large effect size \\ \hline
    \end{tabular}
\end{table}

\begin{table}[!ht]
    \centering
    \caption{P value, Confidence Interval and Effect sizes of Table \ref{llama3-grades-readability-corr-p1p7} \Tseven for \lamatri (*no means $ p \ge 0.05$)}
    \begin{tabular}{lccccc}
        \textbf{Comparison} & \rotB{\textbf{Pearson's r}} & \rotB{\textbf{p-value}} & \rotB{\textbf{95\% CI}} & 
        \rotB{\textbf{Effect Size (Cohen's d)}} & \rotB{\textbf{Cohen's d Interpretation}} \\ \\ \hline
        \lamatri vs. FleschGL & 0.05 & no & [-0.0, 0.1] & 0.86 & Large effect size \\ 
        \lamatri vs. FleschRE & -0.16 & $<0.05$ & [-0.21, -0.11] & -11.19 & Large effect size \\ 
        \lamatri vs. SmogInd & 0.26 & $<0.05$ & [0.21, 0.3] & -0.04 & Small effect size \\ 
        \lamatri vs. ColeLia & 0.19 & $<0.05$ & [0.14, 0.23] & 0.85 & Large effect size \\ 
        \lamatri vs. GunningFog & -0.12 & $<0.05$ & [-0.17, -0.08] & -0.15 & Small effect size \\ 
        \lamatri vs. AutoRea & 0 & no & [-0.04, 0.05] & 0.73 & Large effect size \\ 
        \lamatri vs. LinseaW & -0.08 & $<0.05$ & [-0.13, -0.03] & -0.38 & Medium effect size \\ 
        \lamatri vs. DaleChall & -0.19 & $<0.05$ & [-0.24, -0.14] & -0.25 & Medium effect size \\ \hline
    \end{tabular}
\end{table}

\begin{table}[!ht]
    \centering
    \caption{P value, Confidence Interval and Effect sizes of Table \ref{neg-corr-llm-mistake-p1p7} and \ref{llama3-neg-corr-llm-mistake-p1p7} \Tone (*no means $ p \ge 0.05$)}
    \begin{tabular}{lccccc}
        \textbf{Comparison} & \rotB{\textbf{Pearson's r}} & \rotB{\textbf{p-value}} & \rotB{\textbf{95\% CI}} & 
        \rotB{\textbf{Effect Size (Cohen's d)}} & \rotB{\textbf{Cohen's d Interpretation}} \\ \\ \hline
        
        \rone vs. Aspell misspellings  & 0.14 & $<0.05$ & [0.09, 0.18] & -0.94 & Large effect size \\ 
        \rone vs. LanguageTool grammar & 0.09 & $<0.05$ & [0.05, 0.14] & 0.65 & Large effect size \\ 
        \rone vs. LanguageTool spelling & 0.11 & $<0.05$ & [0.06, 0.16] & -1.1 & Large effect size \\ 
        \rone vs. LanguageTool style & 0.09 & $<0.05$ & [0.04, 0.13] & 3.46 & Large effect size \\ 
        \rone vs. LanguageTool punctuation & 0.02 & no & [-0.03, 0.06] & 1.65 & Large effect size \\ 
        \rone vs. LanguageTool capitalization & 0.11 & $<0.05$ & [0.06, 0.15] & 3.96 & Large effect size \\
\hline
    \end{tabular}
\end{table}

\begin{table}[!ht]
    \centering
    \caption{P value, Confidence Interval and Effect sizes of Table \ref{neg-corr-llm-mistake-p1p7} and \ref{llama3-neg-corr-llm-mistake-p1p7} \Tseven (*no means $ p \ge 0.05$)}
    \begin{tabular}{lccccc}
        \textbf{Comparison} & \rotB{\textbf{Pearson's r}} & \rotB{\textbf{p-value}} & \rotB{\textbf{95\% CI}} & 
        \rotB{\textbf{Effect Size (Cohen's d)}} & \rotB{\textbf{Cohen's d Interpretation}} \\ \\ \hline
        
        \rone vs. Aspell misspellings & 0.11 & $<0.05$ & [0.06, 0.15] & 1.13 & Large effect size \\ 
        \rone vs. LanguageTool grammar & 0.05 & $<0.05$ & [0.01, 0.1] & 3.31 & Large effect size \\ 
        \rone vs. LanguageTool spelling & 0.2 & $<0.05$ & [0.15, 0.25] & 0.43 & Medium effect size \\ 
        \rone vs. LanguageTool style & 0.11 & $<0.05$ & [0.06, 0.16] & 1.39 & Large effect size \\ 
        \rone vs. LanguageTool punctuation & 0.11 & $<0.05$ & [0.06, 0.16] & 2.91 & Large effect size \\ 
        \rone vs. LanguageTool capitalization & 0.05 & $<0.05$ & [0.0, 0.1] & 3.84 & Large effect size \\
\hline
    \end{tabular}
\end{table}

\begin{table}[!ht]
    \centering
    \caption{P value, Confidence Interval and Effect sizes of Table \ref{asap-plus-corr-ltool}  (*no means $ p \ge 0.05$)}
    \begin{tabular}{lccccc}
        \textbf{Comparison} & \rotB{\textbf{Pearson's r}} & \rotB{\textbf{p-value}} & \rotB{\textbf{95\% CI}} & 
        \rotB{\textbf{Effect Size (Cohen's d)}} & \rotB{\textbf{Cohen's d Interpretation}} \\ \\ \hline

        Content vs. Aspell misspellings & 0.04 & no & [-0.01, 0.08] & -1.02 & Large  \\ 
        Content vs. LanguageTool grammar & -0.01 & no & [-0.06, 0.03] & 0.41 & Medium  \\ 
        Content vs. LanguageTool spelling & -0.01 & no & [-0.05, 0.04] & -1.17 & Large  \\ 
        Content vs. LanguageTool style & 0.06 & $<0.05$ & [0.01, 0.1] & 2.87 & Large  \\ 
        Content vs. LanguageTool punctuation & -0.03 & no & [-0.07, 0.02] & 1.31 & Large  \\ 
        Content vs. LanguageTool capitalization & 0.05 & $<0.05$ & [0.01, 0.1] & 3.26 & Large  \\
\hline
    \end{tabular}
\end{table}

\begin{table}[!ht]
    \centering
    \caption{P value, Confidence Interval and Effect sizes of Table \ref{chatgpt-feedback-corr-p1p7} and \ref{llama3-chatgpt-feedback-corr-p1p7} \Tone (*no means $ p \ge 0.05$)}
    \begin{tabular}{lccccc}
        \multicolumn{1}{p{2.8cm}}{\textbf{Comparison with \CGPT Explanation (Exp.)} }
        & \rotB{\textbf{Pearson's r}} & \rotB{\textbf{p-value}} & \rotB{\textbf{95\% CI}} & 
        \rotB{\textbf{Effect Size (Cohen's d)}} & \rotB{\textbf{Cohen's d Interpretation}} \\ \\ \hline
        
        \rone vs. Exp. Length (sentences) & 0.12 & $<0.05$ & [0.07, 0.16] & -1.18 & Large effect size \\ 
        \rone vs. Exp. Length (tokens) & 0.17 & $<0.05$ & [0.12, 0.21] & -5.01 & Large effect size \\ 
        \rone vs. Exp. Sentiment (average) & 0.06 & $<0.05$ & [0.01, 0.11] & 6.37 & Large effect size \\ 
        \rone vs. Exp. Sentiment (max) & 0.04 & no & [-0.01, 0.08] & 5.93 & Large effect size \\ 
        \rtwo vs. Exp. Length (sentences) & 0.11 & $<0.05$ & [0.06, 0.16] & -1.18 & Large effect size \\ 
        \rtwo vs. Exp. Length (tokens) & 0.15 & $<0.05$ & [0.11, 0.2] & -5.01 & Large effect size \\
\hline
    \end{tabular}
\end{table}

\begin{table}[!ht]
    \centering
    \caption{P value, Confidence Interval and Effect sizes of Table \ref{chatgpt-feedback-corr-p1p7} and \ref{llama3-chatgpt-feedback-corr-p1p7} \Tseven (*no means $ p \ge 0.05$)}
    \begin{tabular}{lccccc}
        \multicolumn{1}{p{2.8cm}}{\textbf{Comparison with \CGPT explanation} }
        & \rotB{\textbf{Pearson's r}} & \rotB{\textbf{p-value}} & \rotB{\textbf{95\% CI}} & 
        \rotB{\textbf{Effect Size (Cohen's d)}} & \rotB{\textbf{Cohen's d Interpretation}} \\ \\ \hline
        \rone vs. Exp. Length (sentences) & 0.05 & $<0.05$ & [0.0, 0.1] & 0.91 & Large effect size \\ 
        \rone vs. Exp. Length (tokens) & 0.04 & no & [-0.01, 0.09] & -3.63 & Large effect size \\ 
        \rone vs. Exp. Sentiment (average) & 0.05 & no & [-0.0, 0.1] & 4.96 & Large effect size \\ 
        \rone vs. Exp. Sentiment (max) & 0.02 & no & [-0.03, 0.07] & 4.73 & Large effect size \\ 
        \rtwo vs. Exp. Length (sentences) & 0.06 & $<0.05$ & [0.01, 0.11] & 0.9 & Large effect size \\ 
        \rtwo vs. Exp. Length (tokens) & 0.04 & no & [-0.01, 0.09] & -3.63 & Large effect size \\
\hline
    \end{tabular}
\end{table}

\begin{table}[!ht]
    \centering
\caption{P value, Confidence Interval and Effect sizes of Table \ref{lama-feedback-corr-p1p7} and \ref{llama3-lama-feedback-corr-p1p7} \Tone }
    \begin{tabular}{lccccc}
        \multicolumn{1}{p{2.8cm}}{\textbf{Comparison with \lama-2 explanation} }
        & \rotB{\textbf{Pearson's r}} & \rotB{\textbf{p-value}} & \rotB{\textbf{95\% CI}} & 
        \rotB{\textbf{Effect Size (Cohen's d)}} & \rotB{\textbf{Cohen's d Interpretation}} \\ \\ \hline
        
        \rone vs. Exp. Length (sentences) & 0.26 & $<0.05$ & [0.22, 0.31] & -3.24 & Large effect size \\ 
        \rone vs. Exp. Length (tokens) & 0.26 & $<0.05$ & [0.22, 0.31] & -5.98 & Large effect size \\ 
        \rone vs. Exp. Sentiment (average) & 0.3 & $<0.05$ & [0.25, 0.34] & 5.13 & Large effect size \\ 
        \rone vs. Exp. Sentiment (max) & 0.24 & $<0.05$ & [0.19, 0.28] & 5.19 & Large effect size \\ 
        \rtwo vs. Exp. Length (sentences) & 0.25 & $<0.05$ & [0.21, 0.29] & -3.24 & Large effect size \\ 
        \rtwo vs. Exp. Length (tokens) & 0.27 & $<0.05$ & [0.23, 0.31] & -5.98 & Large effect size \\
\hline
    \end{tabular}
\end{table}

\begin{table}[!ht]
    \centering
    
    \caption{P value, Confidence Interval and Effect sizes of Table \ref{lama-feedback-corr-p1p7} and \ref{llama3-lama-feedback-corr-p1p7} \Tseven }
    \begin{tabular}{lccccc}
        \multicolumn{1}{p{2.8cm}}{\textbf{Comparison with \lama-2 explanation} }
        & \rotB{\textbf{Pearson's r}} & \rotB{\textbf{p-value}} & \rotB{\textbf{95\% CI}} & 
        \rotB{\textbf{Effect Size (Cohen's d)}} & \rotB{\textbf{Cohen's d Interpretation}} \\ \\ \hline
        \rone vs. Exp. Length (sentences) & 0.16 & $<0.05$ & [0.12, 0.21] & 0.63 & Large effect size \\ 
        \rone vs. Exp. Length (tokens) & 0.21 & $<0.05$ & [0.16, 0.26] & -2.91 & Large effect size \\ 
        \rone vs. Exp. Sentiment (average) & 0.26 & $<0.05$ & [0.21, 0.3] & 4.75 & Large effect size \\ 
        \rone vs. Exp. Sentiment (max) & 0.22 & $<0.05$ & [0.18, 0.27] & 4.68 & Large effect size \\ 
        \rtwo vs. Exp. Length (sentences) & 0.13 & $<0.05$ & [0.08, 0.17] & 0.63 & Large effect size \\ 
        \rtwo vs. Exp. Length (tokens) & 0.17 & $<0.05$ & [0.12, 0.22] & -2.91 & Large effect size \\
\hline
    \end{tabular}
\end{table}

\begin{table}[!ht]
    \centering
    \caption{P value, Confidence Interval and Effect sizes of Table \ref{lama3-feedback-corr-p1p7} \Tone }
    \begin{tabular}{lccccc}
        \multicolumn{1}{p{2.8cm}}{\textbf{Comparison with \lamatri explanation} }
        & \rotB{\textbf{Pearson's r}} & \rotB{\textbf{p-value}} & \rotB{\textbf{95\% CI}} & 
        \rotB{\textbf{Effect Size (Cohen's d)}} & \rotB{\textbf{Cohen's d Interpretation}} \\ \\ \hline
        
        \rone vs. Exp. Length (sentences) & 0.35 & $<0.05$ & [0.3, 0.39] & -3.78 & Large effect size \\ 
        \rone vs. Exp. Length (tokens) & 0.4 & $<0.05$ & [0.36, 0.44] & -7.17 & Large effect size \\ 
        \rone vs. Exp. Sentiment (average) & 0.23 & $<0.05$ & [0.19, 0.28] & 4.91 & Large effect size \\ 
        \rone vs. Exp. Sentiment (max) & 0.21 & $<0.05$ & [0.17, 0.26] & 5.09 & Large effect size \\ 
        \rtwo vs. Exp. Length (sentences) & 0.37 & $<0.05$ & [0.32, 0.41] & -3.8 & Large effect size \\ 
        \rtwo vs. Exp. Length (tokens) & 0.41 & $<0.05$ & [0.37, 0.45] & -7.17 & Large effect size \\
\hline
    \end{tabular}
\end{table}

\begin{table}[!ht]
    \centering
    \caption{P value, Confidence Interval and Effect sizes of Table \ref{lama3-feedback-corr-p1p7} \Tseven }
    \begin{tabular}{lccccc}
        \multicolumn{1}{p{2.8cm}}{\textbf{Comparison with \lamatri explanation} }
        & \rotB{\textbf{Pearson's r}} & \rotB{\textbf{p-value}} & \rotB{\textbf{95\% CI}} & 
        \rotB{\textbf{Effect Size (Cohen's d)}} & \rotB{\textbf{Cohen's d Interpretation}} \\ \\ \hline
        \rone vs. Exp. Length (sentences) & 0.25 & $<0.05$ & [0.21, 0.3] & 1.33 & Large effect size \\ 
        \rone vs. Exp. Length (tokens) & 0.34 & $<0.05$ & [0.3, 0.38] & -7.32 & Large effect size \\ 
        \rone vs. Exp. Sentiment (average) & 0.28 & $<0.05$ & [0.23, 0.32] & 4.76 & Large effect size \\ 
        \rone vs. Exp. Sentiment (max) & 0.2 & $<0.05$ & [0.15, 0.25] & 4.5 & Large effect size \\ 
        \rtwo vs. Exp. Length (sentences) & 0.29 & $<0.05$ & [0.24, 0.33] & 1.3 & Large effect size \\ 
        \rtwo vs. Exp. Length (tokens) & 0.36 & $<0.05$ & [0.32, 0.4] & -7.32 & Large effect size \\
\hline
    \end{tabular}
\end{table}

To identify the specific group differences, we performed post-hoc analyses using the Tukey Honestly Significant Difference (HSD) test for the results detailed in Sections \ref{section-RQ4} and \ref{llama3-section-RQ4}. Post-hoc analysis helps in identifying specific differences between pairs of groups after finding a significant overall difference. In our reporting, we included the following elements:

\begin{itemize}
    \item Group1 and Group2 columns are the groups being compared
    \item Meandiff is the difference between the group means (mean of Group1 is subtracted from mean of Group2)
    \item P-adj is the corrected p-value which takes into account the multiple comparisons being conducted
    \item CI lower is the lower band of the confidence interval. In the current example the confidence interval at the 95\% level since $\alpha = 0.05$.
    \item CI upper is the upper band of the confidence interval. In the current example the confidence interval at the 95\% level since $\alpha = 0.05.$
    \item Reject is the decision rule based on the corrected p-value 
\end{itemize}

\begin{table}[!ht]
    \centering
    \caption{Tukey HSD test on Table \ref{misspellcategory-p1} (\CGPT)} 
\begin{tabular}{llccccc}
    \textbf{Group1} & \textbf{Group2} & \textbf{Meandiff} & \textbf{P-adj} & \textbf{CI lower} & \textbf{CI upper} & \textbf{Reject} \\ \hline \hline
    
\multicolumn{7}{c} {\CGPT Misspelling Category} \\ \hline  \\  
    \multicolumn{7}{c} {Aspell misspelling} \\ \hline
            No mention & Unqualified mention & 3.978 & 0.0 & 2.0122 & 5.9439 & True \\ 
            No mention & Qualified mention & 5.3687 & 0.0 & 4.1207 & 6.6167 & True \\ 
            Unqualified mention & Qualified mention & 1.3906 & 0.3191 & -0.87 & 3.6512 & False \\
            \hline 
            
    \multicolumn{7}{c} {LanguageTool spelling} \\
\hline 
    \end{tabular}
\end{table}

\begin{table}[!ht]
    \centering
    \caption{Tukey HSD test on Table \ref{misspellcategory-p1} (\lama)}
    \begin{tabular}{llccccc}
    \textbf{Group1} & \textbf{Group2} & \textbf{Meandiff} & \textbf{P-adj} & \textbf{CI lower} & \textbf{CI upper} & \textbf{Reject} \\ \hline \hline

\multicolumn{7}{c} {\lama Misspelling Category} \\ \hline \\
    \multicolumn{7}{c} {Aspell Misspelling Category} \\ \hline
    No mention & Unqualified mention & 1.2196 & 0.0904 & -0.1437 & 2.5829 & False \\ 
    No mention & Qualified mention & 6.6317 & 0.0 & 3.9932 & 9.2703 & True \\ 
    Unqualified mention & Qualified mention & 5.4122 & 0.0 & 2.4958 & 8.3285 & True \\ 
    \hline
    
    \multicolumn{7}{c} {LanguageTool Spelling} \\ \hline
    No mention & Unqualified mention & 1.2424 & 0.1129 & -0.2159 & 2.7008 & False \\ 
    No mention & Qualified mention & 7.2897 & 0.0 & 4.4673 & 10.1121 & True \\ 
    Unqualified mention & Qualified mention & 6.0473 & 0.0 & 2.9277 & 9.1669 & True \\ 
    \hline
    
    \multicolumn{7}{c} {\lama Score} \\ \hline
    No mention & Unqualified mention & 0.1439 & 0.0253 & 0.0142 & 0.2735 & True \\ 
    No mention & Qualified mention & -1.0048 & 0.0 & -1.2558 & -0.7538 & True \\ 
    Unqualified mention & Qualified mention & -1.1486 & 0.0 & -1.426 & -0.8713 & True \\ 
    \hline 
    
    \multicolumn{7}{c} {\lama Sentiment (average)} \\ \hline
    No mention & Unqualified mention & 0.0445 & 0.4861 & -0.0466 & 0.1356 & False \\ 
    No mention & Qualified mention & -0.5642 & 0.0 & -0.7405 & -0.3879 & True \\ 
    Unqualified mention & Qualified mention & -0.6087 & 0.0 & -0.8036 & -0.4138 & True \\ 
    \hline 
    
    \multicolumn{7}{c} {\lama Sentiment (max)} \\ \hline
    No mention & Unqualified mention & -0.0142 & 0.9393 & -0.113 & 0.0846 & False \\ 
    No mention & Qualified mention & -0.5163 & 0.0 & -0.7075 & -0.3252 & True \\ 
    Unqualified mention & Qualified mention & -0.5022 & 0.0 & -0.7134 & -0.2909 & True \\ 
    \hline 
    \end{tabular}
\end{table}

\begin{table}[!ht]
    \centering
    \caption{Tukey HSD test on Table \ref{llama3-misspellcategory-p1} with \lamatri}
    \begin{tabular}{llccccc}
    \textbf{Group1} & \textbf{Group2} & \textbf{Meandiff} & \textbf{P-adj} & \textbf{CI lower} & \textbf{CI upper} & \textbf{Reject} \\ \hline \hline

    \multicolumn{7}{c} {\lamatri Misspelling Category} \\ \hline \\
\multicolumn{7}{c} {Aspell misspelling} \\ \hline
No mention & Unqualified mention & 5.4726 & 0.0 & 4.0327 & 6.9125 & True \\ 
No mention & Qualified mention & 4.9629 & 0.0 & 3.6628 & 6.263 & True \\ 
Unqualified mention & Qualified mention & -0.5097 & 0.7959 & -2.3665 & 1.3471 & False \\ 
\hline

\multicolumn{7}{c} {LanguageTool spelling} \\ \hline
No mention & Unqualified mention & 5.4189 & 0.0 & 3.8708 & 6.967 & True \\ 
No mention & Qualified mention & 5.0866 & 0.0 & 3.6889 & 6.4844 & True \\ 
Unqualified mention & Qualified mention & -0.3323 & 0.9194 & -2.3286 & 1.6641 & False \\ 
\hline

\multicolumn{7}{c} {\lamatri Score} \\ \hline
No mention & Unqualified mention & -0.0071 & 0.9968 & -0.2255 & 0.2113 & False \\ 
No mention & Qualified mention & -0.4458 & 0.0 & -0.643 & -0.2486 & True \\ 
Unqualified mention & Qualified mention & -0.4387 & 0.0008 & -0.7203 & -0.1571 & True \\ 
\hline 

\multicolumn{7}{c} {\lamatri Sentiment (average)} \\ \hline
No mention & Unqualified mention & 0.001 & 0.9999 & -0.1686 & 0.1707 & False \\ 
No mention & Qualified mention & -0.153 & 0.0504 & -0.3062 & 0.0002 & False \\ 
Unqualified mention & Qualified mention & -0.154 & 0.2246 & -0.3728 & 0.0648 & False \\ 
\hline 

\multicolumn{7}{c} {\lamatri Sentiment (average)} \\ \hline
No mention & Unqualified mention & -0.0327 & 0.8741 & -0.188 & 0.1226 & False \\ 
No mention & Qualified mention & -0.1657 & 0.0155 & -0.306 & -0.0255 & True \\ 
Unqualified mention & Qualified mention & -0.133 & 0.2644 & -0.3333 & 0.0673 & False \\ 
\hline 

    \end{tabular}
\end{table}

\begin{table}[!ht]
    \centering
    \caption{Tukey HSD test on Table \ref{misspellcategory-p7} (\CGPT)}
    \begin{tabular}{llccccc}
    \textbf{Group1} & \textbf{Group2} & \textbf{Meandiff} & \textbf{P-adj} & \textbf{CI lower} & \textbf{CI upper} & \textbf{Reject} \\ \hline \hline
    
\multicolumn{7}{c} {\CGPT Misspelling Category} \\ \hline  \\    
    \multicolumn{7}{c} {Aspell Misspelling Category} \\ \hline
    No mention & Unqualified mention & 0.5445 & 0.2205 & -0.2244 & 1.3133 & False \\ 
    No mention & Qualified mention & 1.646 & 0.0 & 1.0751 & 2.217 & True \\ 
    Unqualified mention & Qualified mention & 1.1016 & 0.0059 & 0.2633 & 1.9399 & True \\ 
    \hline
    
    \multicolumn{7}{c} {LanguageTool Spelling} \\ \hline
    No mention & Unqualified mention & 0.704 & 0.2062 & -0.2686 & 1.6765 & False \\ 
    No mention & Qualified mention & 1.8044 & 0.0 & 1.0822 & 2.5266 & True \\ 
    Unqualified mention & Qualified mention & 1.1004 & 0.0398 & 0.04 & 2.1608 & True \\ 
    \hline
    
    \multicolumn{7}{c} {\CGPT Score} \\ \hline
    No mention & Unqualified mention & 0.1211 & 0.655 & -0.2029 & 0.445 & False \\ 
    No mention & Qualified mention & -0.9254 & 0.0 & -1.1659 & -0.6848 & True \\ 
    Unqualified mention & Qualified mention & -1.0464 & 0.0 & -1.3997 & -0.6932 & True \\ 
    \hline 
    
    \multicolumn{7}{c} {\CGPT Sentiment (average)} \\ \hline
    No mention & Unqualified mention & 0.0226 & 0.7618 & -0.0527 & 0.0979 & False \\ 
    No mention & Qualified mention & -0.0513 & 0.0803 & -0.1072 & 0.0047 & False \\ 
    Unqualified mention & Qualified mention & -0.0738 & 0.0883 & -0.1559 & 0.0083 & False \\ 
    \hline 
    
    \multicolumn{7}{c} {\CGPT Sentiment (max)} \\ \hline
    No mention & Unqualified mention & 0.0292 & 0.7636 & -0.0687 & 0.1271 & False \\ 
    No mention & Qualified mention & -0.0324 & 0.5472 & -0.1051 & 0.0402 & False \\ 
    Unqualified mention & Qualified mention & -0.0616 & 0.365 & -0.1683 & 0.0451 & False \\ \\ \hline

    \end{tabular}
\end{table}

\begin{table}[!ht]
    \centering
    \caption{Tukey HSD test on Table \ref{misspellcategory-p7} (\lama)}
    \begin{tabular}{llccccc}
    \textbf{Group1} & \textbf{Group2} & \textbf{Meandiff} & \textbf{P-adj} & \textbf{CI lower} & \textbf{CI upper} & \textbf{Reject} \\ \hline \hline
    
\multicolumn{7}{c} {\lama Misspelling Category} \\ \hline \\
    \multicolumn{7}{c} {Aspell Misspelling Category} \\ \hline
    No mention & Unqualified mention & 0.5445 & 0.2205 & -0.2244 & 1.3133 & False \\ 
    No mention & Qualified mention & 1.646 & 0.0 & 1.0751 & 2.217 & True \\ 
    Unqualified mention & Qualified mention & 1.1016 & 0.0059 & 0.2633 & 1.9399 & True \\ 
    \hline
    
    \multicolumn{7}{c} {LanguageTool Spelling} \\ \hline
    No mention & Unqualified mention & 0.704 & 0.2062 & -0.2686 & 1.6765 & False \\ 
    No mention & Qualified mention & 1.8044 & 0.0 & 1.0822 & 2.5266 & True \\ 
    Unqualified mention & Qualified mention & 1.1004 & 0.0398 & 0.04 & 2.1608 & True \\ 
    \hline
    
    \multicolumn{7}{c} {\lama Score} \\ \hline
    No mention & Unqualified mention & 0.1211 & 0.655 & -0.2029 & 0.445 & False \\ 
    No mention & Qualified mention & -0.9254 & 0.0 & -1.1659 & -0.6848 & True \\ 
    Unqualified mention & Qualified mention & -1.0464 & 0.0 & -1.3997 & -0.6932 & True \\ 
    \hline 
    
    \multicolumn{7}{c} {\lama Sentiment (average)} \\ \hline
    No mention & Unqualified mention & 0.0226 & 0.7618 & -0.0527 & 0.0979 & False \\ 
    No mention & Qualified mention & -0.0513 & 0.0803 & -0.1072 & 0.0047 & False \\ 
    Unqualified mention & Qualified mention & -0.0738 & 0.0883 & -0.1559 & 0.0083 & False \\ 
    \hline 
    
    \multicolumn{7}{c} {\lama Sentiment (max)} \\ \hline
    No mention & Unqualified mention & 0.0292 & 0.7636 & -0.0687 & 0.1271 & False \\ 
    No mention & Qualified mention & -0.0324 & 0.5472 & -0.1051 & 0.0402 & False \\ 
    Unqualified mention & Qualified mention & -0.0616 & 0.365 & -0.1683 & 0.0451 & False \\ 
    \hline 
    \end{tabular}
\end{table}

\begin{table}[!ht]
    \centering
    \caption{Tukey HSD test on Table \ref{llama3-misspellcategory-p7} (\lamatri)}
    \begin{tabular}{llccccc}
    \textbf{Group1} & \textbf{Group2} & \textbf{Meandiff} & \textbf{P-adj} & \textbf{CI lower} & \textbf{CI upper} & \textbf{Reject} \\ \hline \hline
    
\multicolumn{7}{c} {\lama Misspelling Category} \\ \hline
    No mention & Unqualified mention & 1.6317 & 0.0 & 1.0424 & 2.221 & True \\ 
    No mention & Qualified mention & 3.8705 & 0.0 & 3.3251 & 4.4159 & True \\ 
    Unqualified mention & Qualified mention & 2.2388 & 0.0 & 1.619 & 2.8585 & True \\ 
    \hline
    
    \multicolumn{7}{c} {LanguageTool Spelling} \\ \hline
    No mention & Unqualified mention & 1.646 & 0.0 & 0.8711 & 2.4209 & True \\ 
    No mention & Qualified mention & 3.4538 & 0.0 & 2.7366 & 4.171 & True \\ 
    Unqualified mention & Qualified mention & 1.8079 & 0.0 & 0.993 & 2.6228 & True \\ 
    \hline
    
    \multicolumn{7}{c} {\lama Score} \\ \hline
    No mention & Unqualified mention & -0.3063 & 0.1483 & -0.6908 & 0.0783 & False \\ 
    No mention & Qualified mention & -2.0522 & 0.0 & -2.4081 & -1.6963 & True \\ 
    Unqualified mention & Qualified mention & -1.7459 & 0.0 & -2.1503 & -1.3415 & True \\ 
    \hline 
    
    \multicolumn{7}{c} {\lama Sentiment (average)} \\ \hline
    No mention & Unqualified mention & -0.0388 & 0.5642 & -0.1279 & 0.0504 & False \\ 
    No mention & Qualified mention & -0.3181 & 0.0 & -0.4006 & -0.2356 & True \\ 
    Unqualified mention & Qualified mention & -0.2793 & 0.0 & -0.373 & -0.1856 & True \\ 
    \hline 
    
    \multicolumn{7}{c} {\lama Sentiment (max)} \\ \hline
    No mention & Unqualified mention & 0.0889 & 0.0976 & -0.0121 & 0.1898 & False \\ 
    No mention & Qualified mention & -0.2181 & 0.0 & -0.3115 & -0.1246 & True \\ 
    Unqualified mention & Qualified mention & -0.3069 & 0.0 & -0.4131 & -0.2008 & True \\ 
    \hline 
    \end{tabular}
\end{table}


\begin{table}[!ht]
    \centering
    \caption{Tukey HSD test on Table \ref{grammaticalcategory-p1} (\CGPT)}
    \begin{tabular}{llccccc}
    \textbf{Group1} & \textbf{Group2} & \textbf{Meandiff} & \textbf{P-adj} & \textbf{CI lower} & \textbf{CI upper} & \textbf{Reject} \\ \hline \hline

    \multicolumn{7}{c} {\CGPT Grammar Category} \\ \hline \\
    
    \multicolumn{7}{c} {LanguageTool Grammar} \\ \hline
    No mention & Unqualified mention & 0.131 & 0.7383 & -0.283 & 0.545 & False \\ 
    No mention & Qualified mention & 0.578 & 0.0002 & 0.2404 & 0.9155 & True \\ 
    Unqualified mention & Qualified mention & 0.4469 & 0.073 & -0.0316 & 0.9255 & False \\ 
    \hline
    
    \multicolumn{7}{c} {\CGPT Score} \\ \hline
    No mention & Unqualified mention & 0.1008 & 0.0346 & 0.0057 & 0.1959 & True \\ 
    No mention & Qualified mention & -0.0363 & 0.5144 & -0.1138 & 0.0412 & False \\ 
    Unqualified mention & Qualified mention & -0.1371 & 0.0097 & -0.247 & -0.0273 & True \\ 
    \hline
    
    \multicolumn{7}{c} {\CGPT Sentiment (average)} \\ \hline
    No mention & Unqualified mention & -0.186 & 0.0 & -0.2757 & -0.0963 & True \\ 
    No mention & Qualified mention & -0.2579 & 0.0 & -0.331 & -0.1847 & True \\ 
    Unqualified mention & Qualified mention & -0.0719 & 0.235 & -0.1756 & 0.0318 & False \\ 
    \hline 
    
    \multicolumn{7}{c} {\CGPT Sentiment (max)} \\ \hline
    No mention & Unqualified mention & -0.1476 & 0.0013 & -0.2458 & -0.0494 & True \\ 
    No mention & Qualified mention & -0.2469 & 0.0 & -0.3269 & -0.1668 & True \\ 
    Unqualified mention & Qualified mention & -0.0992 & 0.1006 & -0.2127 & 0.0143 & False \\ 
    \hline 
    \end{tabular}
\end{table}

\begin{table}[!ht]
    \centering
    \caption{Tukey HSD test on Table \ref{grammaticalcategory-p1} (\lama)}
    \begin{tabular}{llccccc}
    \textbf{Group1} & \textbf{Group2} & \textbf{Meandiff} & \textbf{P-adj} & \textbf{CI lower} & \textbf{CI upper} & \textbf{Reject} \\ \hline \hline

    \multicolumn{7}{c} {\lama Grammar Category} \\ \hline \\
    
    \multicolumn{7}{c} {LanguageTool Grammar} \\ \hline
    No mention & Unqualified mention & 0.1308 & 0.6799 & -0.2356 & 0.4972 & False \\ 
    No mention & Qualified mention & 0.9196 & 0.0101 & 0.1792 & 1.66 & True \\ 
    Unqualified mention & Qualified mention & 0.7889 & 0.0533 & -0.0086 & 1.5863 & False \\ 
    \hline
    
    \multicolumn{7}{c} {\lama Score} \\ \hline
    No mention & Unqualified mention & 0.085 & 0.0904 & -0.01 & 0.1801 & False \\ 
    No mention & Qualified mention & -0.938 & 0.0 & -1.1301 & -0.7459 & True \\ 
    Unqualified mention & Qualified mention & -1.0231 & 0.0 & -1.23 & -0.8162 & True \\ 
    \hline
    
    \multicolumn{7}{c} {\lama Sentiment (average)} \\ \hline
    No mention & Unqualified mention & 0.0199 & 0.7692 & -0.0476 & 0.0873 & False \\ 
    No mention & Qualified mention & -0.4554 & 0.0 & -0.5917 & -0.319 & True \\ 
    Unqualified mention & Qualified mention & -0.4752 & 0.0 & -0.6221 & -0.3284 & True \\ 
    \hline 
    
    \multicolumn{7}{c} {\lama Sentiment (max)} \\ \hline
    No mention & Unqualified mention & -0.0244 & 0.7152 & -0.0976 & 0.0489 & False \\ 
    No mention & Qualified mention & -0.3992 & 0.0 & -0.5472 & -0.2512 & True \\ 
    Unqualified mention & Qualified mention & -0.3748 & 0.0 & -0.5342 & -0.2154 & True \\ 
    \hline 
    \end{tabular}
\end{table}

\begin{table}[!ht]
    \centering
    \caption{Tukey HSD test on Table \ref{llama3-grammaticalcategory-p1} (\lamatri)}
    \begin{tabular}{llccccc}
    \textbf{Group1} & \textbf{Group2} & \textbf{Meandiff} & \textbf{P-adj} & \textbf{CI lower} & \textbf{CI upper} & \textbf{Reject} \\ \hline \hline

    \multicolumn{7}{c} {\lamatri Grammar Category} \\ \hline \\
    
    \multicolumn{7}{c} {LanguageTool Grammar} \\ \hline
    No mention & Unqualified mention & 0.6469 & 0.0 & 0.3358 & 0.9579 & True \\ 
    No mention & Qualified mention & 0.8334 & 0.0 & 0.4871 & 1.1797 & True \\ 
    Unqualified mention & Qualified mention & 0.1865 & 0.4457 & -0.1742 & 0.5472 & False \\ 
    \hline
    
    \multicolumn{7}{c} {\lamatri Score} \\ \hline
    No mention & Unqualified mention & -0.5362 & 0.0 & -0.6577 & -0.4146 & True \\ 
    No mention & Qualified mention & -0.777 & 0.0 & -0.9123 & -0.6416 & True \\ 
    Unqualified mention & Qualified mention & -0.2408 & 0.0002 & -0.3817 & -0.0998 & True \\ 
    \hline
    
    \multicolumn{7}{c} {\lamatri Sentiment (average)} \\ \hline
    No mention & Unqualified mention & -0.2826 & 0.0 & -0.3795 & -0.1857 & True \\ 
    No mention & Qualified mention & -0.395 & 0.0 & -0.5029 & -0.2871 & True \\ 
    Unqualified mention & Qualified mention & -0.1124 & 0.0498 & -0.2248 & -0.0001 & True \\ 
    \hline 
    
    \multicolumn{7}{c} {\lamatri Sentiment (max)} \\ \hline
    No mention & Unqualified mention & -0.3142 & 0.0 & -0.4027 & -0.2257 & True \\ 
    No mention & Qualified mention & -0.3385 & 0.0 & -0.4371 & -0.2399 & True \\ 
    Unqualified mention & Qualified mention & -0.0243 & 0.844 & -0.127 & 0.0784 & False \\ 
    \hline 
    \end{tabular}
\end{table}


\begin{table}[!ht]
    \centering
    \caption{Tukey HSD test on Table \ref{grammaticalcategory-p7} (\CGPT)}
    \begin{tabular}{llccccc}
    \textbf{Group1} & \textbf{Group2} & \textbf{Meandiff} & \textbf{P-adj} & \textbf{CI lower} & \textbf{CI upper} & \textbf{Reject} \\ \hline \hline
\multicolumn{7}{c} {\CGPT Grammar Category} \\ \hline  \\ 
    \multicolumn{7}{c} {LanguageTool Grammar} \\ \hline
    No mention & Unqualified mention & 0.0223 & 0.9793 & -0.2461 & 0.2907 & False \\ 
    No mention & Qualified mention & 0.3864 & 0.0003 & 0.1543 & 0.6184 & True \\ 
    Unqualified mention & Qualified mention & 0.3641 & 0.0003 & 0.144 & 0.5841 & True \\ 
    \hline
    
    \multicolumn{7}{c} {\CGPT Score} \\ \hline
    No mention & Unqualified mention & 0.4071 & 0.006 & 0.0967 & 0.7175 & True \\ 
    No mention & Qualified mention & -0.7031 & 0.0 & -0.9714 & -0.4348 & True \\ 
    Unqualified mention & Qualified mention & -1.1102 & 0.0 & -1.3646 & -0.8557 & True \\ 
    \hline
    
    \multicolumn{7}{c} {\CGPT Sentiment (average)} \\ \hline
    No mention & Unqualified mention & 0.016 & 0.8618 & -0.0562 & 0.0882 & False \\ 
    No mention & Qualified mention & -0.0975 & 0.0007 & -0.16 & -0.0351 & True \\ 
    Unqualified mention & Qualified mention & -0.1135 & 0.0 & -0.1727 & -0.0543 & True \\ 
    \hline 
    
    \multicolumn{7}{c} {\CGPT Sentiment (max)} \\ \hline
    No mention & Unqualified mention & 0.0479 & 0.4577 & -0.0463 & 0.1421 & False \\ 
    No mention & Qualified mention & -0.0559 & 0.2419 & -0.1373 & 0.0256 & False \\ 
    Unqualified mention & Qualified mention & -0.1037 & 0.0047 & -0.181 & -0.0265 & True \\ 
    \hline 
    \end{tabular}
\end{table}

\begin{table}[!ht]
    \centering
    \caption{Tukey HSD test on Table \ref{grammaticalcategory-p7} (\lama)}
    \begin{tabular}{llccccc}
    \textbf{Group1} & \textbf{Group2} & \textbf{Meandiff} & \textbf{P-adj} & \textbf{CI lower} & \textbf{CI upper} & \textbf{Reject} \\ \hline \hline

    \multicolumn{7}{c} {\lama Grammar Category} \\ \hline \\
    
    \multicolumn{7}{c} {LanguageTool Grammar} \\ \hline
    No mention & Unqualified mention & 0.1588 & 0.2748 & -0.0839 & 0.4014 & False \\ 
    No mention & Qualified mention & 0.3082 & 0.0017 & 0.0983 & 0.5181 & True \\ 
    Unqualified mention & Qualified mention & 0.1494 & 0.3848 & -0.1164 & 0.4152 & False \\ 
    \hline
    
    \multicolumn{7}{c} {\lama Score} \\ \hline
    No mention & Unqualified mention & -0.2552 & 0.3014 & -0.6599 & 0.1496 & False \\ 
    No mention & Qualified mention & -1.8381 & 0.0 & -2.1882 & -1.488 & True \\ 
    Unqualified mention & Qualified mention & -1.5829 & 0.0 & -2.0263 & -1.1395 & True \\ 
    \hline
    
    \multicolumn{7}{c} {\lama Sentiment (average)} \\ \hline
    No mention & Unqualified mention & -0.0639 & 0.29 & -0.1637 & 0.0359 & False \\ 
    No mention & Qualified mention & -0.2761 & 0.0 & -0.3624 & -0.1898 & True \\ 
    Unqualified mention & Qualified mention & -0.2122 & 0.0 & -0.3215 & -0.1029 & True \\ 
    \hline 
    
    \multicolumn{7}{c} {\lama Sentiment (max)} \\ \hline
    No mention & Unqualified mention & -0.0229 & 0.8234 & -0.1134 & 0.0676 & False \\ 
    No mention & Qualified mention & -0.2137 & 0.0 & -0.2919 & -0.1354 & True \\ 
    Unqualified mention & Qualified mention & -0.1908 & 0.0 & -0.2899 & -0.0916 & True \\ 
    \hline 
    \end{tabular}
\end{table}

\begin{table}[!ht]
    \centering
    \caption{Tukey HSD test on Table \ref{llama3-grammaticalcategory-p7} (\lamatri)}
    \begin{tabular}{llccccc}
    \textbf{Group1} & \textbf{Group2} & \textbf{Meandiff} & \textbf{P-adj} & \textbf{CI lower} & \textbf{CI upper} & \textbf{Reject} \\ \hline \hline

    \multicolumn{7}{c} {\lamatri Grammar Category} \\ \hline \\
    
    \multicolumn{7}{c} {LanguageTool Grammar} \\ \hline
    No mention & Unqualified mention & 0.0096 & 0.9955 & -0.2389 & 0.2581 & False \\ 
    No mention & Qualified mention & 0.7094 & 0.0 & 0.4622 & 0.9567 & True \\ 
    Unqualified mention & Qualified mention & 0.6999 & 0.0 & 0.5026 & 0.8971 & True \\ 
    \hline
    
    \multicolumn{7}{c} {\lamatri Score} \\ \hline
    No mention & Unqualified mention & 0.1781 & 0.5891 & -0.2479 & 0.604 & False \\ 
    No mention & Qualified mention & -1.7737 & 0.0 & -2.1975 & -1.3498 & True \\ 
    Unqualified mention & Qualified mention & -1.9518 & 0.0 & -2.2899 & -1.6137 & True \\ 
    \hline
    
    \multicolumn{7}{c} {\lamatri Sentiment (average)} \\ \hline
    No mention & Unqualified mention & 0.041 & 0.5905 & -0.0573 & 0.1394 & False \\ 
    No mention & Qualified mention & -0.2928 & 0.0 & -0.3906 & -0.195 & True \\ 
    Unqualified mention & Qualified mention & -0.3338 & 0.0 & -0.4119 & -0.2558 & True \\ 
    \hline 
    
    \multicolumn{7}{c} {\lamatri Sentiment (max)} \\ \hline
    No mention & Unqualified mention & 0.0828 & 0.1865 & -0.0281 & 0.1937 & False \\ 
    No mention & Qualified mention & -0.2578 & 0.0 & -0.3682 & -0.1474 & True \\ 
    Unqualified mention & Qualified mention & -0.3406 & 0.0 & -0.4286 & -0.2526 & True \\ 
    \hline 
    \end{tabular}
\end{table}


\begin{table}[!ht]
    \centering
    \caption{Tukey HSD test on Table \ref{llama3-humancategory-p1} or \ref{humancategory-p1} }
    \begin{tabular}{llccccc}
    \textbf{Group1} & \textbf{Group2} & \textbf{Meandiff} & \textbf{P-adj} & \textbf{CI lower} & \textbf{CI upper} & \textbf{Reject} \\ \hline \hline
    \multicolumn{7}{c} {\CGPT Score} \\ \hline
    Lower & Middle & 0.6076 & 0.0 & 0.3607 & 0.8546 & True \\
    Lower & Upper & 0.7919 & 0.0 & 0.5443 & 1.0395 & True \\
    Middle & Upper & 0.1843 & 0.0 & 0.1226 & 0.246 & True \\\hline
    \multicolumn{7}{c} {\lama Score } \\ \hline
    Lower & Middle & 1.7155 & 0.0 & 1.4686 & 1.9625 & True \\
    Lower & Upper & 2.2693 & 0.0 & 2.0218 & 2.5169 & True \\
    Middle & Upper & 0.5538 & 0.0 & 0.4921 & 0.6155 & True \\ \hline
    \multicolumn{7}{c} {\lamatri Score } \\ \hline
    Lower & Middle & 1.0583 & 0.0 & 0.66 & 1.4566 & True \\
    Lower & Upper & 1.9425 & 0.0 & 1.5432 & 2.3418 & True \\
    Middle & Upper & 0.8842 & 0.0 & 0.7847 & 0.9837 & True \\
    \hline 
    \end{tabular}
\end{table}

\begin{table}[!ht]
    \centering
    \caption{Tukey HSD test on Table \ref{llama3-humancategory-p7} or \ref{humancategory-p7} }
    \begin{tabular}{llccccc}
    \textbf{Group1} & \textbf{Group2} & \textbf{Meandiff} & \textbf{P-adj} & \textbf{CI lower} & \textbf{CI upper} & \textbf{Reject} \\ \hline \hline

    \multicolumn{7}{c} {\CGPT Score} \\ \hline
    Lower & Middle & 1.0849 & 0.0 & 0.652 & 1.5177 & True \\
    Lower & Upper & 1.5391 & 0.0 & 1.1073 & 1.971 & True \\
    Middle & Upper & 0.4542 & 0.0 & 0.2348 & 0.6737 & True \\ \hline

    \multicolumn{7}{c} {\lama Score } \\ \hline
    Lower & Middle & 1.0414 & 0.0002 & 0.433 & 1.6497 & True \\
    Lower & Upper & 2.7294 & 0.0 & 2.1225 & 3.3363 & True \\
    Middle & Upper & 1.688 & 0.0 & 1.3796 & 1.9965 & True \\ \hline

    \multicolumn{7}{c} {\lamatri Score } \\ \hline
    Lower & Middle & 2.6748 & 0.0 & 2.0978 & 3.2518 & True \\
    Lower & Upper & 4.6779 & 0.0 & 4.1023 & 5.2535 & True \\
    Middle & Upper & 2.0031 & 0.0 & 1.7105 & 2.2956 & True \\ \hline
    \end{tabular}
\end{table}

\section{Additional Information}
First, we compared two language-checking tools, Aspell and LanguageTool, noting that Aspell strongly correlates with LanguageTool in terms of spelling accuracy. Next, we discuss our experiments with machine learning models using ChatGPT embeddings, where the Support Vector Regression model demonstrated superior performance. Lastly, we included the definition of Cohen's d as a measurement of effect size. Details can be found below.

\subsection{Aspell strongly correlates with \ltool spelling} \label{aspell-ltool-section}

Remarkably, the strong correlation coefficient of 0.95 between Aspell and LanguageTool, as depicted in Table \ref{aspell-ltool-corr}, underscores the robustness and consistency of errors identified by both tools. This high level of agreement between the two spell/grammar checkers reinforces our confidence in the accuracy of the mistakes detected, providing a solid foundation for our subsequent analyses.

\begin{table}[!h]
    \centering
    \captionsetup{width=0.75\textwidth}
    \caption{Aspell strongly correlates with \ltool spelling mistake count}
    \begin{tabular}{lccccc}
        \diagbox[width=3.8cm, height=2.2cm]{Aspell \\ Misspelling }{ \ltool \\ Error Count } &
        \rotB{\textbf{Grammar}} & 
        \rotB{\textbf{Spelling}} & 
        \rotB{\textbf{Style}} & 
        \rotB{\textbf{Punctuation}} & 
        \rotB{\textbf{Capitalization}} \\ \\ \hline
        \textbf{\Tone } & 0.28 & \textbf{0.95} & 0.05 & 0.15 & 0.14 \\ 
        \textbf{\Tseven} & 0.32 & \textbf{0.75} & 0.29 & 0.17 & 0.16 \\ 
    \end{tabular}
    \label{aspell-ltool-corr}
\end{table}

\subsection{Support Vector Regression model performs best}

We conducted an investigation into the performance of various machine learning models, aiming to achieve the ability to approximate human scores. Our dataset comprises a total of 3352 sample essays from \Tone and \Tseven combined. To standardize the human rater scores of different ranges for the two tasks, we normalized the average score given by two raters. By Utilizing OpenAI's embedding model \textit{text-embedding-ada-002} with 1536 dimensions we generate essay embeddings to train the regression models. Our evaluation metric is the Pearson correlation coefficient denoted as $r$ used to assess performance. Table \ref{ml-result} presents the results of this analysis.
\begin{table}[!h]
    \centering
    \captionsetup{width=0.7\textwidth}
    \caption{Machine Learning models with \CGPT embedding in predicting human scores }
    \begin{tabular}{lcc}
        \textbf{Regression Models} & \textbf{Train $r$} & \textbf{Test $r$} \\ \hline
        \textbf{Multiple Linear} & 0.95 & 0.73 \\ 
        \textbf{Random Forest} & 0.96 & 0.66 \\ 
        \textbf{Decision Tree} & 1.00 & 0.43 \\ 
        \textbf{Support Vector} & 0.9 & \textbf{0.82} \\ 
        \textbf{Xgboost} & 1.00 & 0.72 \\
    \end{tabular}
    \label{ml-result}
\end{table}

Utilizing \CGPT we previously obtained a correlation of $r=0.21-0.23$ with the overall scores given by human raters shown in \ref{human-llm-donot-correlate} and $r=0.33-0.36$ with the trait scores given by human raters shown in \ref{traits-llm-donot-correlate}. All the regression models improved upon these results. Among the evaluated models, the Support Vector Machine (SVM) emerged as the top performer, achieving the strongest correlation score of $r=0.82$ on the test dataset. This indicates a strong correlation between the predicted and actual human scores when using SVM. Conversely, models such as decision tree regression displayed weaker correlations, showing minimal learning with $r=0.43$ on the test dataset. While multiple linear and XGBoost regression models showed almost similar performance with $r=0.73$ and $r=0.72$ respectively, but they fell short of the predictive power demonstrated by SVM. These findings suggest that SVM with \CGPT embedding exhibits promising potential for accurately predicting human grades on our dataset, outperforming other regression models considered in our analysis.First, we compared two language-checking tools, Aspell and LanguageTool, noting that Aspell strongly correlates with LanguageTool in terms of spelling accuracy. Next, we discuss our experiments with machine learning models using ChatGPT embeddings, where the Support Vector Regression model demonstrated superior performance. Lastly, we included the definition of Cohen's d as a measurement of effect size. Details can be found below.

\subsection{Effect Size: Cohen's d} \label{cohenD}
Cohen's d typically ranges from negative to positive values. A negative Cohen's d indicates that the mean of the first group is smaller than the mean of the second group, while a positive Cohen's d indicates that the mean of the first group is larger than the mean of the second group. The absolute value of Cohen's d reflects the magnitude of the effect size, with larger absolute values indicating larger effect sizes. Generally, the following conventions are used to interpret the effect size:

Small effect size: $d = 0.2$.
Medium effect size: $d = 0.5$.
Large effect size: $d >= 0.8$. 

\end{appendices}

\end{document}